\documentclass[Afour,sageh,times]{sagej}

\usepackage{moreverb,url}
\usepackage{enumitem}
\usepackage{booktabs}
\usepackage{rotating}
\usepackage{pifont}
\newcommand{\cmark}{\ding{51}}
\newcommand{\xmark}{\ding{55}}
\usepackage{makecell}

\usepackage{xcolor}
\usepackage{tikz}
\usetikzlibrary{arrows.meta}

\newcommand{\timelinecite}[1]{%
  \shortstack[c]{\citeauthor{#1}\\\citeyearpar{#1}}%
}

\usepackage{booktabs}
\usepackage{tabularx}
\usepackage{array}
\usepackage{multirow}

\usepackage[colorlinks,bookmarksopen,bookmarksnumbered,citecolor=red,urlcolor=red]{hyperref}

\usepackage[table]{xcolor}

\newcommand\BibTeX{{\rmfamily B\kern-.05em \textsc{i\kern-.025em b}\kern-.08em
T\kern-.1667em\lower.7ex\hbox{E}\kern-.125emX}}

\def\volumeyear{2026}

\begin{document}

\runninghead{Ma et al.}

\title{Robot Learning from Human Videos: A Survey}

\author{Junyi Ma\affilnum{1,*}, Erhang Zhang\affilnum{1,*}, Haoran Yang\affilnum{1}, Ditao Li\affilnum{1}, Chenyang Xu\affilnum{1}, Guangming Wang\affilnum{2}, Hesheng Wang\affilnum{1}}

\affiliation{\affilnum{1}Shanghai Jiao Tong University, China\\
\affilnum{2}University of Cambridge, UK\\ \affilnum{*}Authors are with equal contributions}

\corrauth{Hesheng Wang, is with the Department of Automation, Shanghai Jiao Tong University, Shanghai 200240, China and the Key Laboratory of System Control and Information Processing, Ministry of Education of China.}

\email{wanghesheng@sjtu.edu.cn}

\begin{abstract}
A critical bottleneck hindering further advancement in embodied AI and robotics is the challenge of scaling robot data. To address this, the field of learning robot manipulation skills from human video data has attracted rapidly growing attention in recent years, driven by the abundance of human activity videos and advances in computer vision. This line of research promises to enable robots to acquire skills passively from the vast and readily available resource of human demonstrations, substantially favoring scalable learning for generalist robotic systems. Therefore, we present this survey to provide a comprehensive and up-to-date review of human-video-based learning techniques in robotics, focusing on both human-robot skill transfer and data foundations. We first review the policy learning foundations in robotics, and then describe the fundamental interfaces to incorporate human videos. Subsequently, we introduce a hierarchical taxonomy of transferring human videos to robot skills, covering task-, observation-, and action-oriented pathways, along with a cross-family analysis of their couplings with different data configurations and learning paradigms. In addition, we investigate the data foundations including widely-used human video datasets and video generation schemes, and provide large-scale statistical trends in dataset development and utilization. Ultimately, we emphasize the challenges and limitations intrinsic to this field, and delineate potential avenues for future research. The paper list of our survey is available at \url{https://github.com/IRMVLab/awesome-robot-learning-from-human-videos}.
\end{abstract}


\maketitle

\section{Introduction}

With the sharp increase in demand for automation in modern industry and society, the past few years have witnessed rapid progress in embodied AI and robotics. A key engine of this growth is the remarkable advancement of robot learning technologies such as imitation learning and reinforcement learning. They provide essential and extensive solutions for robotic manipulation and whole-body control. 

\begin{figure}[t]
  \centering
  \captionsetup{belowskip=1pt}
  \includegraphics[width=1\linewidth]{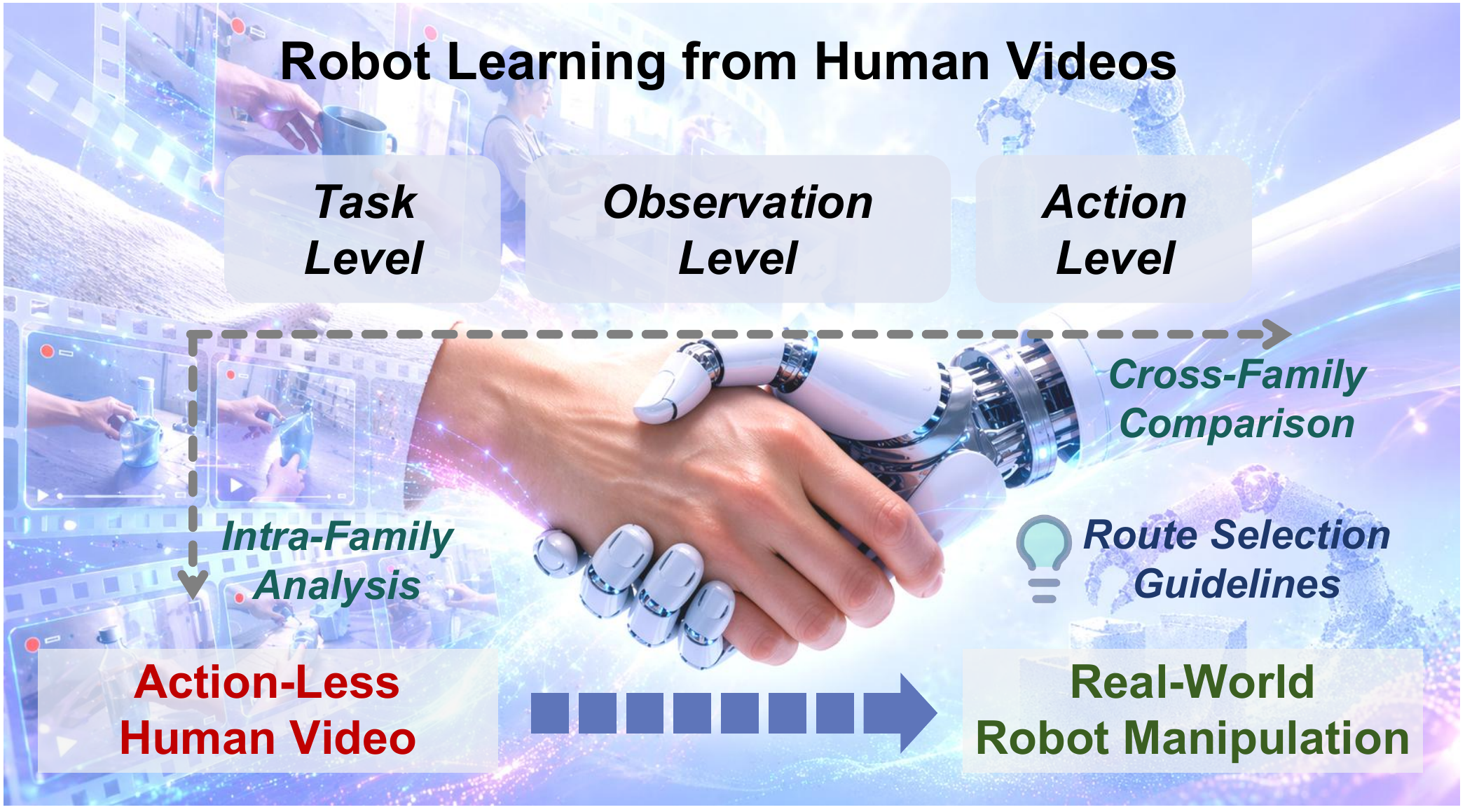}
  \caption{Illustration of bridging human videos and robot execution reviewed in this survey. We organize related works into three categories: task-oriented, observation-oriented, and action-oriented transfer pathways. We conduct intra-family analysis and cross-family comparison to highlight their design principles and trade-offs, and further provide practical guidelines for selecting suitable LfHV routes.}
  \label{fig:teaser}
  \vspace{-0.49cm}
\end{figure}

\begin{figure*}[t]
  \centering
  \captionsetup{aboveskip=2pt, belowskip=0pt}
  \includegraphics[width=1\linewidth]{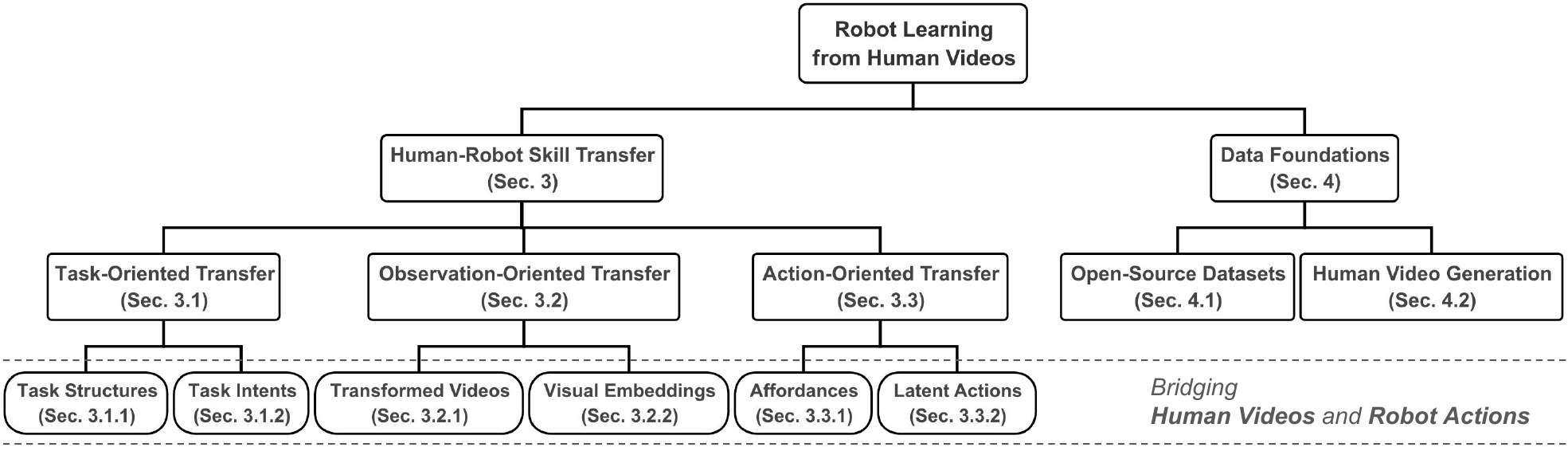}
  \caption{Taxonomy of robot learning from human videos.}
  \label{fig:taxonomy}
  \vspace{-0.3cm}
\end{figure*}

However, the increasing complexity of task scenarios in manufacturing, logistics, and daily service settings poses growing challenges to the versatility and generalization ability of robotic systems. Conventional imitation learning paradigms~\citep{brohan2022rt,zhao2023learning,fu2024mobile,ze20243d,chi2025diffusion} rely on a labor-intensive and repetitive process for collecting kinesthetic or teleoperated demonstrations. This substantially impedes the deployment efficiency of robot policies in real-world applications. Moreover, the limited diversity and paucity of demonstration data further constrain generalization to new operation environments. While integrating reinforcement learning algorithms~\citep{schulman2017proximal,fujimoto2018addressing,haarnoja2018soft,hafner2019dream,janner2019trust} helps improve robot policy adaptability through interaction with the environments, they basically depend on carefully engineered reward functions and large amounts of trial-and-error experience. This not only incurs high training costs in real robotic systems, but also raises concerns regarding sample efficiency, training stability, and exploration safety. Therefore, both imitation learning and reinforcement learning still face fundamental challenges in scaling robot skill acquisition to diverse open-world scenarios. In recent years, the advancements in large language models and vision-language models have been largely driven by the dramatic increase in training data scale~\citep{chen2024expanding,grattafiori2024llama,yang2025qwen3}. Inspired by this, improving the generalization ability of robot policies requires identifying data sources that are substantially easier to collect and scale up than conventional robot data.

In this context, human activity videos have emerged as a promising source of supervision for robot learning. Benefiting from continued progress in computer vision, human videos can be analyzed more accurately and efficiently. Such data format inherently contains dense task semantics and rich interaction patterns with diverse objects, closely aligning with the requirements of robot policy learning. In addition, compared to teleoperated demonstration data, human videos are much easier to collect, and a vast amount of readily available video data already exists on the Internet. This makes large-scale data acquisition for developing generalist robots more feasible. Although existing surveys~\citep{mccarthy2025towards,eze2025learning,feng2026human,zheng2026video} provide valuable perspectives on video-based robot learning, several aspects still warrant a more focused review and summary for LfHV. In particular, the field still lacks a taxonomy centered on how information flows from human videos to robot execution. Besides, a systematic cross-comparison of transfer pathways in terms of viewpoint choices, reliance on real robot data, and learning paradigms remains underexplored. The organization of human video sources is also fragmented, especially with respect to dataset prevalence, usage patterns, and the emerging role of video generation schemes. Moreover, given the rapid pace of progress in the literature, some very recent methods and datasets that are naturally beyond their scope should be further discussed.

\begin{table*}[t]
\centering
\scriptsize
\setlength{\tabcolsep}{4.2pt}
\caption{Quantitative evidence showing that human videos are more data-efficient than conventional robot teleoperation.}
\label{tab:human_video_efficiency}
\begin{tabularx}{\linewidth}{p{2.3cm}p{1.8cm}p{3.4cm} X}
\toprule
Reference & Setting & Compared setting & Quantitative efficiency claim \\
\midrule
\cite{jang2022bc} & Single-arm & Human videos vs. teleoperated robot demonstrations & Human videos can be collected \textbf{5x--7x faster} than teleoperated robot demonstrations. \\
\cite{kareer2025egomimic} & Bimanual & One hour of human data vs. one hour of robot data & One hour of human data yields about \textbf{1400 demonstrations}, compared with only \textbf{135 demonstrations} from one hour of robot data (\textbf{$\sim$10.4x} higher data yield). \\
\cite{dan2025x} & Single-arm & Human videos vs. teleoperated robot demonstrations & Human videos take about \textbf{20s} per clip, compared with \textbf{60s} per robot demonstration. \\
\cite{zhou2025you} & Bimanual & Human-video-driven auto-rollout demonstrations vs. teleoperation & Auto-rollout is \textbf{much faster than teleoperation}, collecting about \textbf{300 demonstrations in 8 hours}. It further supports \textbf{100x} data expansion to \textbf{5K--24K} trajectories per task. \\
\cite{freeman2026warped} & Single-arm & Egocentric human demonstrations vs. teleoperation & The human-video-based pipeline requires \textbf{5x--8x less data collection time}, i.e., is approximately \textbf{5x--8x faster} than teleoperation. \\
\bottomrule
\end{tabularx}
\end{table*}

To this end, we conduct this survey to provide a comprehensive summary and detailed classification of the methodological landscape, data foundations, and emerging research trends in robot learning from human videos. Our goal is to offer a clearer foundation for future research in embodied AI and robotics. As illustrated in Fig.~\ref{fig:teaser}, we conceptualize this problem as bridging human videos and robot manipulation through multiple levels of transfer. The general structure of this survey is illustrated in Fig.~\ref{fig:taxonomy}. Our main contributions can be summarized as follows:
\begin{itemize}[leftmargin=1em]
    \setlength{\parskip}{0pt}
    \item \textbf{Hierarchical bridging mechanisms of skill transfer:} We present a hierarchical taxonomy of the pathways linking human videos to robot execution. It highlights how human-robot skill transfer can be established at the task, observation, and action levels. For each level, we identify the key video-derived intermediates that serve as transfer bridges, and provide practical guidelines for selecting suitable LfHV routes.
    \item \textbf{Cross-family analysis across data configurations and learning paradigms:} Beyond taxonomy construction, we provide a comparative analysis of how different transfer families relate to viewpoint choice, dependence on real-world robot data, and learning paradigms. This reveals the characteristic design couplings and methodological tradeoffs underlying existing LfHV methods.
    \item \textbf{Systematic summary of human-object interaction analysis:} Beyond categorizing skill transfer pathways, we systematically summarize the commonly used methods for parsing human-object interactions in videos. In particular, we review representative tools for hand and object detection, tracking, reconstruction, and pose estimation in 2D, 3D, and 4D spaces. This helps researchers identify which off-the-shelf interaction analysis techniques are currently practical and popular in the LfHV literature.
    \item \textbf{Evolution and preferences of human video data:} We present the largest statistical analysis of human video sources to date (by far) in the LfHV context, systematically summarizing the development trends of human video datasets and relevant generative techniques. Based on the comprehensive landscape, we further analyze the data preferences exhibited by different categories of LfHV methods.
    \item \textbf{Future directions across models, data, and benchmarks:} Building upon the trends observed in both bridging mechanisms and data foundations, we identify several promising directions for future research. In particular, we highlight opportunities in new modeling paradigms, richer data modalities, more standardized benchmarks, and more collaborative ecosystems.
\end{itemize}

To maintain a clearer and more targeted scope, this survey attends to LfHV works that leverage human video data to facilitate robotic manipulation policies, rather than those on whole-body control and locomotion without complex manipulation~\citep{mao2024learning,allshire2025visual,li2025robomirror,yang2026zerowbc}. The remainder of the paper is organized as follows: Sec.~\ref{sec:background} (Background) reviews foundational robot learning paradigms along with their interfaces for incorporating human videos, and introduces a unified formulation of the LfHV problem. Sec.~\ref{sec:hr_skill_transfer} (Human-Robot Skill Transfer) categorizes the bridging mechanism between human videos and robot execution, and investigates related works in a hierarchical manner. Sec.~\ref{sec:data_foundations} (Data Foundations) introduces open-source datasets for existing human videos, and generative techniques for imagined counterparts. Sec.~\ref{sec:discussion} (Discussion) delves into the key challenges in this emerging field and outlines promising future directions. Finally, Sec.~\ref{sec:conclusion} (Conclusion) provides a summary of this survey.

\begin{figure*}[t]
  \centering
  \captionsetup{aboveskip=2pt, belowskip=0pt}
  \includegraphics[width=1\linewidth]{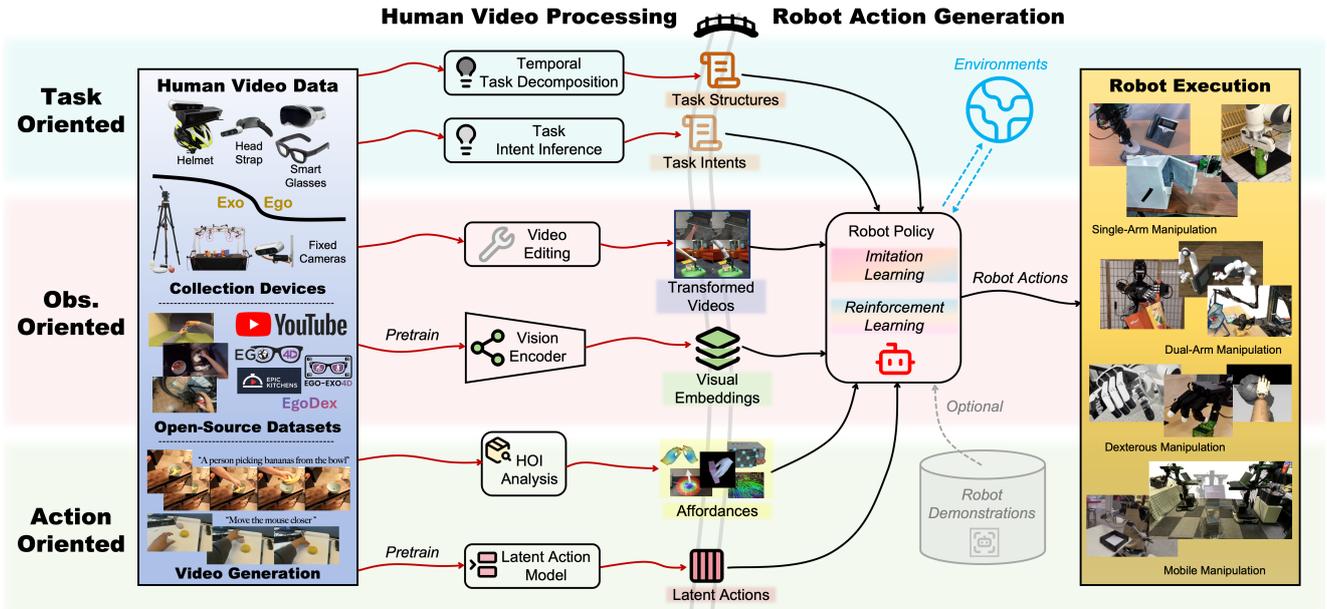}
  \caption{Illustration of bridging human videos and robot execution in task, observation, and action levels.}
  \label{fig:bridging_overview}
\end{figure*}

\section{Background} \label{sec:background}

Robot learning from human videos typically builds upon two fundamental paradigms in robotics: \textit{imitation learning} (IL) and \textit{reinforcement learning} (RL). They provide the core frameworks for learning visuomotor policies, and the existing LfHV approaches can be regarded as extending or reinterpreting them. Therefore, before reviewing human-video-based robot learning, we briefly revisit the mathematical foundations of IL and RL, and discuss the interfaces through which human videos can be incorporated into these frameworks. A unified formulation that characterizes LfHV as minimizing cross-embodiment discrepancies is further provided in this section.

\subsection{Imitation Learning}

Imitation Learning aims to learn a policy by mimicking expert demonstrations. In robotics, a trajectory is typically represented as $\tau = \{(o_t, s_t, a_t)\}_{t=1}^T$, where $o_t$ denotes visual observations, $s_t$ represents proprioceptive states, and $a_t$ corresponds to actions. In modern embodied AI systems, task instructions $l$ are often incorporated to specify task goals. The standard formulation of imitation learning is maximum likelihood estimation over a dataset of expert demonstrations $\mathcal{D}_{\text{expert}}$:
\begin{equation}
\max_{\theta} \sum_{\tau \in \mathcal{D}_{\text{expert}}} \sum_{t} 
\log \pi_{\theta}(a_t \mid o_{\leq t}, s_{\leq t}, l),
\label{eq:il_eq}
\end{equation}
where $\pi_{\theta}$ denotes the IL policy parameterized by $\theta$.

Incorporating human videos into robot imitation learning aims to reduce the substantial cost of collecting $\mathcal{D}_{\text{expert}}$ with teleoperation (see Tab.~\ref{tab:human_video_efficiency}). It therefore attempts to reconstruct $l$, $o_t$, and $a_t$ in Eq.~(\ref{eq:il_eq}) with information extracted from human videos. $s_t$ can then be determined according to the forms of $o_t$ and $a_t$. For task instructions $l$, human videos inherently provide the procedural structure of how humans accomplish target manipulation tasks. Thus, task instructions can be readily parsed from human videos in text or image form for robot task planning. For visual observations $o_t$, the most intuitive strategy is to transform human video observations into embodiment-agnostic or robot-centric observations. Alternatively, learning a vision encoder from human videos to compress $o_t$ into visual representations directly usable by robot policies is also practical. For actions $a_t$, explicit affordances like hand-object poses and latent action representations extracted from human videos can be used as guidance for robot action generation.

\subsection{Reinforcement Learning}

Reinforcement Learning focuses on learning a policy through interaction with an environment, aiming to maximize cumulative rewards. Formally, the RL objective is:
\begin{equation}
\max_{\theta} \; \mathbb{E}_{\tau \sim \pi_{\theta}} \left[
\sum_{t=1}^{T} \gamma^{t-1} R_{\phi}(s_t, a_t)
\right], \,\, \text{with } \theta \leftarrow \theta_{0},
\end{equation}
where $R_{\phi}(s_t, a_t)$ denotes the reward function parameterized by $\phi$, $\gamma$ is a discount factor, $\theta$ represents the learnable policy parameters, and $\theta_{0}$ denotes the initialization of the policy. Compared with imitation learning, RL does not necessarily require expert action labels, but instead improves policies through trial-and-error exploration guided by rewards. However, designing reasonable reward functions and achieving efficient exploration are notoriously major challenges for real-world robotic applications.

Therefore, human videos can be accordingly incorporated into the RL framework through two important interfaces, policy initialization for $\theta_{0}$ and reward learning for $R_{\phi}$. For policy initialization, human videos can provide explicit affordances like hand trajectories as an initial policy prior $\theta_{0}$. Such initialization can reduce the burden of exploration and improve sample efficiency in downstream robot learning. For reward shaping, instead of manually engineered $R_{\phi}$, informative signals like visual embeddings and affordances can be inferred from human videos to construct the reward functions. In this way, human videos guide RL either by providing a better starting point for policy optimization or by supplying a more informative and consistent learning objective.

As can be seen, the role of human videos in robot learning can be interpreted through the lenses of imitation learning and reinforcement learning. From the perspective of imitation learning, human videos serve as a scalable substitute for expert demonstrations by providing task instructions, visual observations, and action cues. From the perspective of reinforcement learning, they offer informative priors for policy initialization and reward construction, thus reducing exploration difficulty and manual reward engineering. This perspective also suggests that most existing LfHV methods can be viewed as operating on specific interfaces of IL and RL. Different bridging mechanisms can be further abstracted from these interfaces of IL and RL paradigms, facilitating the following methodological taxonomy.

\subsection{A Unified Formulation of Learning from Human Videos} \label{sec:unified_formulation}

Despite differences in policy learning paradigms, a robot typically receives task instructions, perceives its environment, and then generates actions to complete manipulation tasks. Thus, the problem of learning from human videos can be further abstracted as minimizing multiple sources of cross-embodiment discrepancies. Specifically, we can characterize the learning objective as:
\begin{equation}
\Phi^* = \arg\min_{\Phi} 
\mathcal{L}_{\text{obs}}(\Phi) + 
\mathcal{L}_{\text{act}}(\Phi) + 
\mathcal{L}_{\text{obj}}(\Phi),
\end{equation}
where $\Phi$ denotes the bridging function that maps human videos to a shared representation space. Here, $\mathcal{L}_{\text{obs}}$ measures the observation gap between human and robot perception, $\mathcal{L}_{\text{act}}$ captures the action gap between human behaviors and robot control spaces, and $\mathcal{L}_{\text{obj}}$ represents the objective gap due to the absence of task annotations or explicit reward in human videos. Notably, this conceptual formulation naturally encompasses a broader range of transfer paradigms, including those that fall outside conventional IL or RL, such as direct retargeting from human demonstrations.

Based on this formulation, we can naturally divide the literature into different categories. In the following section, we review representative methods under each category and analyze their design choices to narrow the gap between human videos and robot control.

\section{Human-Robot Skill Transfer} \label{sec:hr_skill_transfer}

According to the formulation in Sec.~\ref{sec:unified_formulation}, the bridging mechanisms from human videos to robot execution can be organized into three categories: \textit{task-oriented transfer, observation-oriented transfer, and action-oriented transfer}. As shown in Fig.~\ref{fig:bridging_overview}, these categories correspond to six forms of information flow, including \textit{task structures, task intents, transformed videos, visual embeddings, affordances, and latent actions}. Next, we will review representative methods under each category and discuss how they enable skill transfer from human videos to robotic systems. We assign a method to the family corresponding to the primary information bridge that dominates human-robot skill transfer, even if auxiliary components from other families are used.

\subsection{Task-Oriented Transfer}

Task-oriented transfer aims to bridge human videos and robot execution at the level of task instructions. Although in-the-wild human videos cannot directly provide action labels for robot policy learning, the procedural organization and objectives of a task can be transferred across embodiments as instructions. Accordingly, this category highlights extracting high-level task knowledge from human videos to guide downstream robot decision-making. Existing methods in this bridging mechanism can be divided into two groups: (1) \textit{task structures as a bridge}, which explicitly decompose a demonstrated task in a human video into temporal instructions, and (2) \textit{task intents as a bridge}, which infer global task objectives or task-phase transition signals without constructing a complete instruction sequence. That is, task structures are explicit, temporally organized representations of task procedures, whereas task intents denote compact, implicit representations of task objectives.

\begin{figure}[t]
  \centering
  \includegraphics[width=1\linewidth]{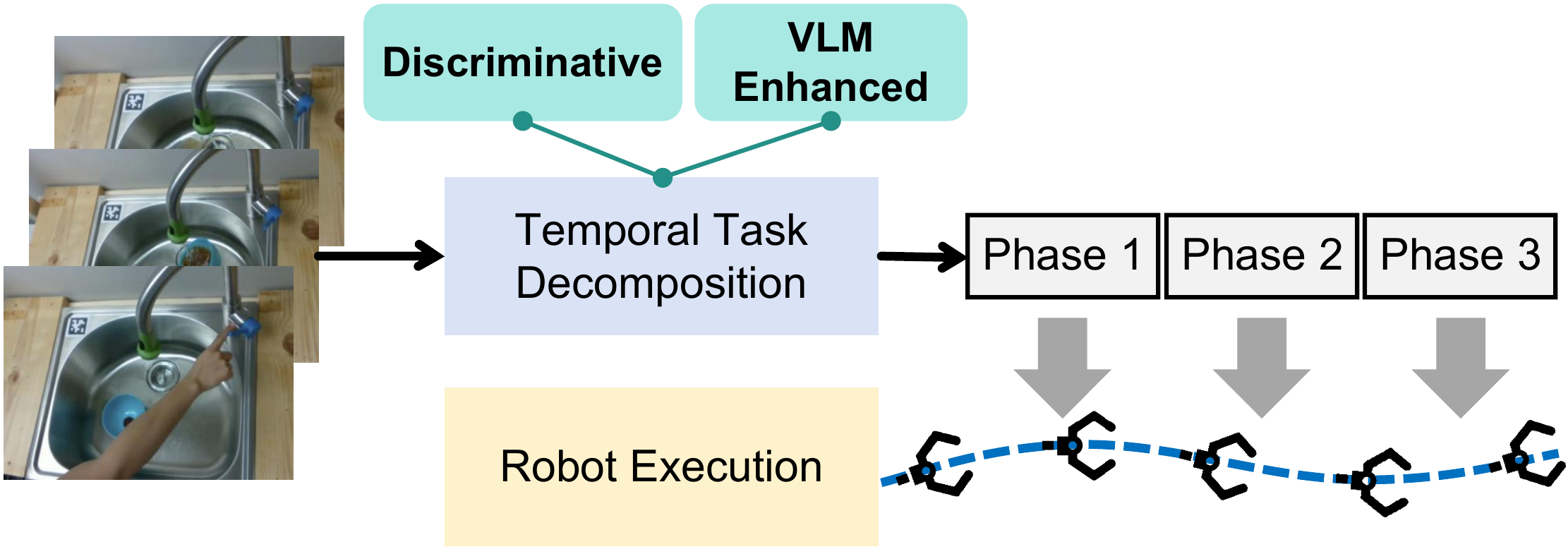}
  \caption{High-level diagram of \textit{task structures as a bridge}.}
  \label{fig:task_structure_transfer}
\end{figure}

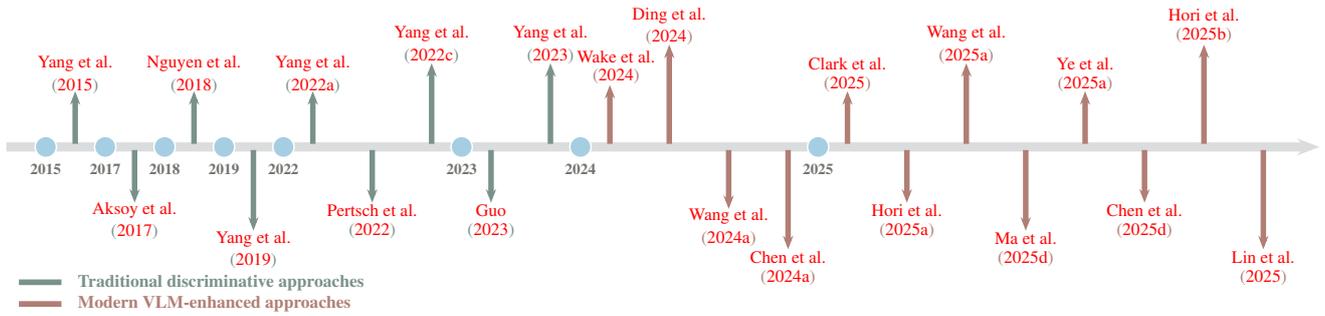
\begin{figure*}[t]
  \centering
  \resizebox{\linewidth}{!}{%
  \begin{tikzpicture}[x=1cm,y=1cm,>=Stealth]
    \definecolor{timelinegreen}{RGB}{122,145,139}
    \definecolor{timelineorange}{RGB}{176,126,116}
    \definecolor{timelinegray}{RGB}{218,220,221}
    \definecolor{timelineyear}{RGB}{118,116,112}
    \definecolor{timelinebubble}{RGB}{166,206,227}

    \draw[timelinegray,line width=4.2pt,-{Stealth[length=3.6mm]}] (-0.10,0) -- (20.90,0);

    \foreach \x/\year in {0.53/2015,1.48/2017,2.43/2018,3.38/2019,4.33/2022,7.18/2023,9.08/2024,12.88/2025} {
      \fill[timelinebubble,draw=white,line width=0.7pt] (\x,0) circle (0.17);
      \node[font=\bfseries\scriptsize,text=timelineyear] at (\x,-0.36) {\year};
    }

    \draw[timelinegreen,line width=2.5pt,-{Stealth[length=2.2mm]}] (1.00,0.05) -- (1.00,0.90);
    \node[align=center,font=\fontsize{8}{6.2}\selectfont,text=timelinegreen] at (1.00,1.17) {\timelinecite{yang2015robot}};

    \draw[timelinegreen,line width=2.5pt,-{Stealth[length=2.2mm]}] (1.95,-0.05) -- (1.95,-0.90);
    \node[align=center,font=\fontsize{8}{6.2}\selectfont,text=timelinegreen] at (1.95,-1.19) {\timelinecite{aksoy2017unsupervised}};

    \draw[timelinegreen,line width=2.5pt,-{Stealth[length=2.2mm]}] (2.90,0.05) -- (2.90,0.90);
    \node[align=center,font=\fontsize{8}{6.2}\selectfont,text=timelinegreen] at (2.90,1.17) {\timelinecite{nguyen2018translating}};

    \draw[timelinegreen,line width=2.5pt,-{Stealth[length=2.2mm]}] (3.85,-0.05) -- (3.85,-1.35);
    \node[align=center,font=\fontsize{8}{6.2}\selectfont,text=timelinegreen] at (3.85,-1.65) {\timelinecite{yang2019learning}};

    \draw[timelinegreen,line width=2.5pt,-{Stealth[length=2.2mm]}] (4.80,0.05) -- (4.80,0.90);
    \node[align=center,font=\fontsize{8}{6.2}\selectfont,text=timelinegreen] at (4.80,1.17) {\timelinecite{yang2022learning}};

    \draw[timelinegreen,line width=2.5pt,-{Stealth[length=2.2mm]}] (5.75,-0.05) -- (5.75,-0.90);
    \node[align=center,font=\fontsize{8}{6.2}\selectfont,text=timelinegreen] at (5.75,-1.19) {\timelinecite{pertsch2022cross}};

    \draw[timelinegreen,line width=2.5pt,-{Stealth[length=2.2mm]}] (6.70,0.05) -- (6.70,1.35);
    \node[align=center,font=\fontsize{8}{6.2}\selectfont,text=timelinegreen] at (6.70,1.65) {\timelinecite{yang2022explicit}};

    \draw[timelinegreen,line width=2.5pt,-{Stealth[length=2.2mm]}] (7.65,-0.05) -- (7.65,-0.90);
    \node[align=center,font=\fontsize{8}{6.2}\selectfont,text=timelinegreen] at (7.65,-1.19) {\timelinecite{guo2023learning}};

    \draw[timelinegreen,line width=2.5pt,-{Stealth[length=2.2mm]}] (8.60,0.05) -- (8.60,1.35);
    \node[align=center,font=\fontsize{8}{6.2}\selectfont,text=timelinegreen] at (8.60,1.65) {\timelinecite{yang2023watch}};

    \draw[timelineorange,line width=2.5pt,-{Stealth[length=2.2mm]}] (9.55,0.05) -- (9.55,1.00);
    \node[align=center,font=\fontsize{8}{6.2}\selectfont,text=timelineorange] at (9.65,1.29) {\timelinecite{wake2024gpt}};

    \draw[timelineorange,line width=2.5pt,-{Stealth[length=2.2mm]}] (10.50,0.05) -- (10.50,1.65);
    \node[align=center,font=\fontsize{8}{6.2}\selectfont,text=timelineorange] at (10.50,1.95) {\timelinecite{ding2024knowledge}};

    \draw[timelineorange,line width=2.5pt,-{Stealth[length=2.2mm]}] (11.45,-0.05) -- (11.45,-1.00);
    \node[align=center,font=\fontsize{8}{6.2}\selectfont,text=timelineorange] at (11.45,-1.29) {\timelinecite{wang2024vlm}};

    \draw[timelineorange,line width=2.5pt,-{Stealth[length=2.2mm]}] (12.40,-0.05) -- (12.40,-1.65);
    \node[align=center,font=\fontsize{8}{6.2}\selectfont,text=timelineorange] at (12.40,-1.95) {\timelinecite{chen2024vlmimic}};

    \draw[timelineorange,line width=2.5pt,-{Stealth[length=2.2mm]}] (13.35,0.05) -- (13.35,0.90);
    \node[align=center,font=\fontsize{8}{6.2}\selectfont,text=timelineorange] at (13.35,1.17) {\timelinecite{clark2025action}};

    \draw[timelineorange,line width=2.5pt,-{Stealth[length=2.2mm]}] (14.30,-0.05) -- (14.30,-0.90);
    \node[align=center,font=\fontsize{8}{6.2}\selectfont,text=timelineorange] at (14.30,-1.19) {\timelinecite{hori2025interactive}};

    \draw[timelineorange,line width=2.5pt,-{Stealth[length=2.2mm]}] (15.25,0.05) -- (15.25,1.35);
    \node[align=center,font=\fontsize{8}{6.2}\selectfont,text=timelineorange] at (15.25,1.65) {\timelinecite{wang2025chain}};

    \draw[timelineorange,line width=2.5pt,-{Stealth[length=2.2mm]}] (16.20,-0.05) -- (16.20,-1.35);
    \node[align=center,font=\fontsize{8}{6.2}\selectfont,text=timelineorange] at (16.20,-1.65) {\timelinecite{ma2025egoloc}};

    \draw[timelineorange,line width=2.5pt,-{Stealth[length=2.2mm]}] (17.15,0.05) -- (17.15,0.90);
    \node[align=center,font=\fontsize{8}{6.2}\selectfont,text=timelineorange] at (17.15,1.17) {\timelinecite{ye2025watch}};

    \draw[timelineorange,line width=2.5pt,-{Stealth[length=2.2mm]}] (18.10,-0.05) -- (18.10,-0.90);
    \node[align=center,font=\fontsize{8}{6.2}\selectfont,text=timelineorange] at (18.10,-1.19) {\timelinecite{chen2025fmimic}};

    \draw[timelineorange,line width=2.5pt,-{Stealth[length=2.2mm]}] (19.05,0.05) -- (19.05,1.65);
    \node[align=center,font=\fontsize{8}{6.2}\selectfont,text=timelineorange] at (19.05,1.95) {\timelinecite{hori2025robot}};

    \draw[timelineorange,line width=2.5pt,-{Stealth[length=2.2mm]}] (20.00,-0.05) -- (20.00,-1.65);
    \node[align=center,font=\fontsize{8}{6.2}\selectfont,text=timelineorange] at (20.00,-1.95) {\timelinecite{lin2025physbrain}};

    \draw[timelinegreen,line width=2.6pt,rounded corners=1pt] (0.10,-2.18) -- (0.78,-2.18);
    \node[anchor=west,font=\bfseries\footnotesize,text=timelinegreen] at (0.92,-2.18) {Traditional discriminative approaches};
    
    \draw[timelineorange,line width=2.6pt,rounded corners=1pt] (0.10,-2.53) -- (0.78,-2.53);
    \node[anchor=west,font=\bfseries\footnotesize,text=timelineorange] at (0.92,-2.53) {Modern VLM-enhanced approaches};

  \end{tikzpicture}%
  }
  \caption{Chronological overview of methods under \textit{task structures as a bridge} in Sec.~\ref{sec:task_structures}.}
  \label{fig:task_structure_timeline}
\end{figure*}

\subsubsection{Task Structures} \label{sec:task_structures}
As shown in Fig.~\ref{fig:task_structure_transfer}, \textit{task structures as a bridge} aim to convert human videos into explicit and temporally organized intermediate phases before robot execution (e.g., fine-grained form: position above the handle $\rightarrow$ move down $\rightarrow$ close the gripper $\rightarrow$ move up, or coarse-grained form: pick knife from counter $\rightarrow$ cut cabbage with knife $\rightarrow$ place knife on counter). It guides the robot to follow targeted task plans at different stages. For clarity, we group related methods into \textit{traditional discriminative approaches} and \textit{modern VLM-enhanced approaches}, and summarize their development in the timeline shown in Fig.~\ref{fig:task_structure_timeline}.

\textit{(a) Traditional discriminative approaches:} Early representative works under \textit{task structures as a bridge} decompose tasks of human videos with discriminative models of limited scale or rule-based mechanisms. For example, \cite{yang2015robot} first recognize grasp types and manipulated objects from video frames, and then integrate these perceptual cues into a so-called visual sentence. Based on a probabilistic manipulation action grammar, the visual sentence is further parsed into a hierarchical syntax tree, which is finally translated into an executable sequence of atomic task instructions. To further simplify the task structure parsing pipeline, inspired by~\cite{aksoy2017unsupervised}, \cite{nguyen2018translating} directly translate human videos into verb-noun command sequences using the combination of CNN and RNN in an end-to-end manner. Following this video-to-command line, \cite{yang2019learning}, \cite{yang2022explicit}, and \cite{yang2023watch} further improve LSTM-based command generation for robotic manipulation by jointly exploiting global scene features and grasp-aware local object features. Through integrating a more scalable action recognition model, \cite{pertsch2022cross} directly predict skill distributions on the human demonstration videos, obtaining discrete high-level semantic labels of task structures. To further understand the task structure of periodic tasks, \cite{yang2022learning} introduce the RepNet architecture to estimate the number of action cycles from a single human video, which enables the automatic decomposition of complex long-term tasks into standardized periodic components. To obtain a finer-grained task decomposition in the interaction level, \cite{guo2023learning} identifies hand-object and object-object relationships to segment human videos into three generalizable primitives of contact events. However, due to limitations in model capacity and reliance on predefined rules, these methods exhibit limited generalization in diverse complex task decomposition.

\textit{(b) Modern VLM-enhanced approaches:} Fortunately, benefiting from the rapid development of generative foundation models in recent years, an increasing number of works have adopted VLMs for temporal task decomposition. These works have achieved stronger generalizability on Internet-scale human videos. For example, \cite{wake2024gpt} pioneeringly use GPT-4V~\citep{achiam2023gpt} to recognize human actions in the video and transcribe them into sequential text instructions.
By introducing chain-of-thought techniques, \cite{clark2025action} improve the VLM reasoning capability of automatically annotating human videos with detailed task plan descriptions. SeeDo~\citep{wang2024vlm} and Super-Mimic~\citep{ye2025watch} further exploit hand velocity information for keyframe extraction and video segmentation, and then use VLMs to generate language descriptions for the resulting subtasks. 
To further enrich the hierarchy of task planning, PhysBrain~\citep{lin2025physbrain} adopts three scenario-dependent video segmentation strategies, encompassing fixed-interval segmentation, event-driven segmentation, and kinematic-aware segmentation. Then, it generates task descriptions through a seven-dimensional templated VQA scheme. As can be noted, existing VLM-based methods like SeeDo~\citep{wang2024vlm}, Super-Mimic~\citep{ye2025watch}, and PhysBrain~\citep{lin2025physbrain} basically understand human behaviors and generate corresponding task descriptions only from a single segmented clip or keyframe. However, the micro-steps in long-horizon tasks naturally depend on the context of the entire video. As a result, clip-level and keyframe-level instruction generation loses long-range dependencies and leads to suboptimal task understanding. To address this issue, \cite{hori2025robot} propose a long-context Q-Former that further incorporates temporal context from neighboring clips beyond the current video segment. This architecture improves the accuracy of robot confirmation generation and fine-grained task planning. To alleviate the open-loop nature of VLM reasoning, \cite{hori2025interactive} introduce an additional human error-correction mechanism, enabling interactive replanning of explicit subtask sequences when the initial plan is incorrect. \cite{ma2025egoloc} instead develop an end-to-end temporal interaction localization algorithm, enabling closed-loop feedback without human intervention. It integrates an effective VLM checker to automatically verify the task decomposition results from hand-object interaction events.

Although the above VLM-based methods achieve robust temporal task decomposition across Internet-scale human videos, the resulting task structures remain loosely coupled with robot action planning. They often require an additional stage that uses the decomposed instructions to guide a separate robot policy. To narrow this gap, \cite{ding2024knowledge} propose a knowledge-based Programming by Demonstration framework. It decomposes human videos into structured action plans and semantically maps them to executable robot tasks with product-aware parameterization. VLMimic~\citep{chen2024vlmimic} and FMimic~\citep{chen2025fmimic} further use VLMs to summarize both high-level semantic constraints on control actions and low-level geometric code parameters from human videos. This enables robots to more directly replay the demonstrated behaviors. Furthermore, CoM~\citep{wang2025chain} integrates multimodal cues to represent human videos as structured manipulation programs. This scheme decomposes the demonstrated task into ordered subtasks with associated control-relevant parameters, and then translates them into executable robot code. It further tightens the connection between task structures and robot action generation. 

\textit{(c) Conclusion for task structures:} \textit{Task structures as a bridge} provide robots with explicit and temporally organized procedural knowledge before action generation. Their evolution shows a clear progression from rule-based and discriminative decomposition toward VLM-enhanced task parsing with stronger scalability and richer semantics. Meanwhile, recent efforts further narrow the gap between task decomposition and downstream execution by making the extracted task structures more directly actionable. Nevertheless, these methods still operate primarily at the planning level, and therefore often require additional grounding modules to translate task structures into embodiment-specific robot actions.

\begin{figure}[t]
  \centering
  \includegraphics[width=1\linewidth]{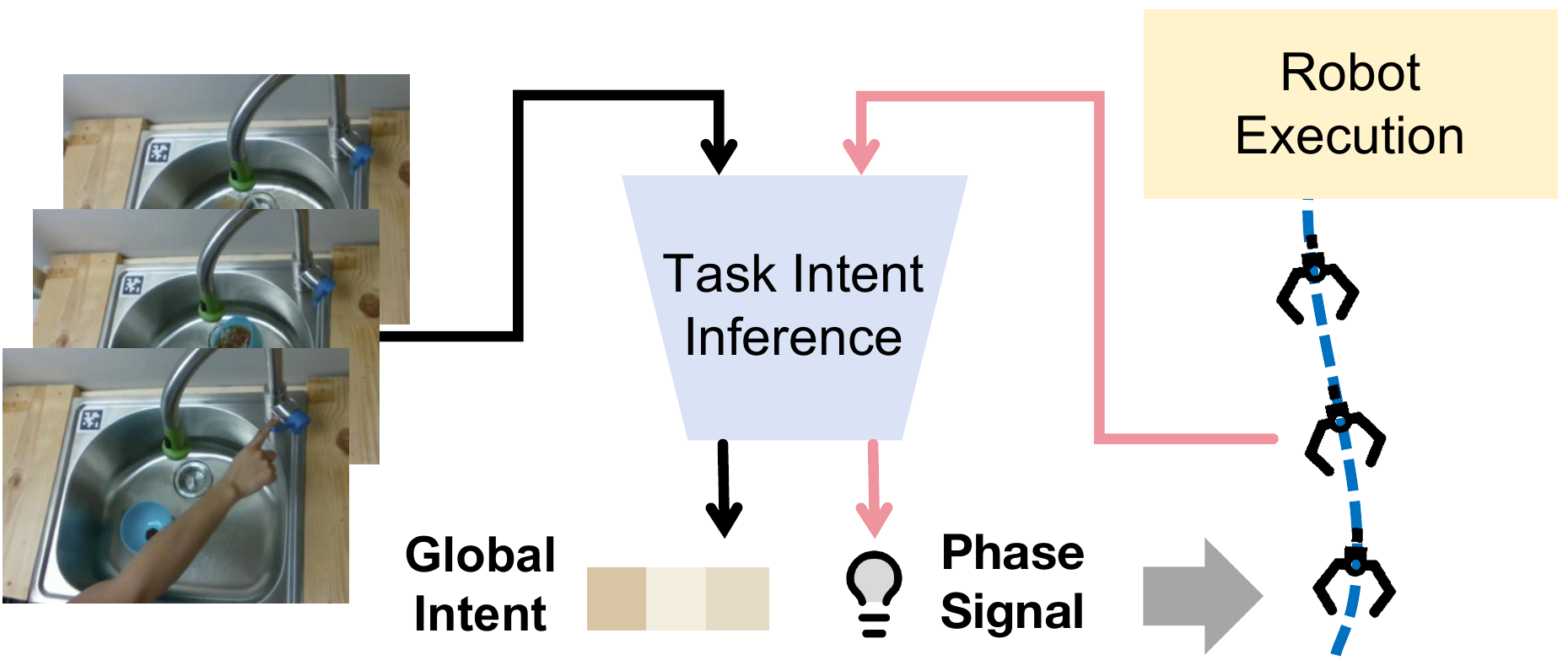}
  \caption{High-level diagram of \textit{task intents as a bridge}.}
  \label{fig:task_intent_transfer}
\end{figure}

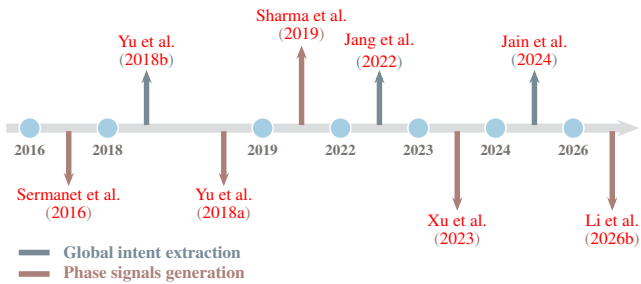
\begin{figure}[t]
  \centering
  \resizebox{\linewidth}{!}{%
  \begin{tikzpicture}[x=1cm,y=1cm,>=Stealth]
    \definecolor{timelineblue}{RGB}{120,138,150}
    \definecolor{timelinebrick}{RGB}{171,128,120}
    \definecolor{timelinegray}{RGB}{218,220,221}
    \definecolor{timelineyear}{RGB}{118,116,112}
    \definecolor{timelinebubble}{RGB}{166,206,227}

    \draw[timelinegray,line width=4.2pt,-{Stealth[length=3.6mm]}] (-0.07,0) -- (10.18,0);

    \foreach \x/\year in {0.31/2016,1.57/2018,4.08/2019,5.34/2022,6.60/2023,7.85/2024,9.11/2026} {
      \fill[timelinebubble,draw=white,line width=0.7pt] (\x,0) circle (0.17);
      \node[font=\bfseries\scriptsize,text=timelineyear] at (\x,-0.36) {\year};
    }

    \draw[timelineblue,line width=2.5pt,-{Stealth[length=2.2mm]}] (2.20,0.05) -- (2.20,0.95);
    \node[align=center,font=\fontsize{8}{8.6}\selectfont,text=timelineblue] at (2.20,1.25) {\timelinecite{yu2018one}};

    \draw[timelineblue,line width=2.5pt,-{Stealth[length=2.2mm]}] (5.97,0.05) -- (5.97,0.95);
    \node[align=center,font=\fontsize{8}{8.6}\selectfont,text=timelineblue] at (5.97,1.25) {\timelinecite{jang2022bc}};

    \draw[timelineblue,line width=2.5pt,-{Stealth[length=2.2mm]}] (8.48,0.05) -- (8.48,0.95);
    \node[align=center,font=\fontsize{8}{8.6}\selectfont,text=timelineblue] at (8.48,1.25) {\timelinecite{jain2024vid2robot}};

    \draw[timelinebrick,line width=2.5pt,-{Stealth[length=2.2mm]}] (0.94,-0.05) -- (0.94,-0.95);
    \node[align=center,font=\fontsize{8}{8.6}\selectfont,text=timelinebrick] at (0.94,-1.25) {\timelinecite{sermanet2016unsupervised}};

    \draw[timelinebrick,line width=2.5pt,-{Stealth[length=2.2mm]}] (3.45,-0.05) -- (3.45,-0.95);
    \node[align=center,font=\fontsize{8}{8.6}\selectfont,text=timelinebrick] at (3.45,-1.25) {\timelinecite{yu2018one_hil}};

    \draw[timelinebrick,line width=2.5pt,-{Stealth[length=2.2mm]}] (4.71,0.05) -- (4.71,1.35);
    \node[align=center,font=\fontsize{8}{8.6}\selectfont,text=timelinebrick] at (4.71,1.67) {\timelinecite{sharma2019third}};

    \draw[timelinebrick,line width=2.5pt,-{Stealth[length=2.2mm]}] (7.23,-0.05) -- (7.23,-1.35);
    \node[align=center,font=\fontsize{8}{8.6}\selectfont,text=timelinebrick] at (7.23,-1.67) {\timelinecite{xu2023xskill}};

    \draw[timelinebrick,line width=2.5pt,-{Stealth[length=2.2mm]}] (9.74,-0.05) -- (9.74,-1.35);
    \node[align=center,font=\fontsize{8}{8.6}\selectfont,text=timelinebrick] at (9.74,-1.67) {\timelinecite{li2026act}};

    \draw[timelineblue,line width=2.6pt,rounded corners=1pt] (0.13,-2.03) -- (0.63,-2.03);
    \node[anchor=west,font=\bfseries\footnotesize,text=timelineblue] at (0.73,-2.03) {Global intent extraction};

    \draw[timelinebrick,line width=2.6pt,rounded corners=1pt] (0.13,-2.38) -- (0.63,-2.38);
    \node[anchor=west,font=\bfseries\footnotesize,text=timelinebrick] at (0.73,-2.38) {Phase signals generation};
  \end{tikzpicture}%
  }
  \caption{Chronological overview of methods under \textit{task intents as a bridge} in Sec.~\ref{sec:task_intents}.}
  \label{fig:task_intent_timeline}
\end{figure}

\subsubsection{Task Intents} \label{sec:task_intents}

Compared with explicit \textit{task structures as a bridge}, \textit{task intents as a bridge} do not require converting a human video into a complete sequence of symbolic instructions. Instead, as shown in Fig.~\ref{fig:task_intent_transfer}, they distill higher-level guidance signals from human videos, such as global task objectives and task-phase transition signals. They can subsequently condition robot control as task intents. Therefore, this bridging mechanism attends to \textit{what should be achieved next} and \textit{how task progress should evolve next} across embodiments. We categorize the related works into \textit{global intent extraction} and \textit{phase signals generation}, as illustrated in Fig.~\ref{fig:task_intent_timeline}.

\textit{(a) Global intent extraction:} Early works in this category represent the entire human video as a global task intent for policy adaptation. For instance, \cite{yu2018one} use domain-adaptive meta-learning to extract task-relevant intent from a single human demonstration video and transfer it to robot policy adaptation for a new task. In this formulation, the human video serves as a compact representation of the task objective to be accomplished next. BC-Z~\citep{jang2022bc} scales this idea to a much broader multi-task setting by conditioning a shared robot policy on task embeddings derived from language or human videos. In this way, human videos serve as compact descriptors of overall task objectives, enabling zero-shot generalization without requiring robot demonstrations for every new task. Similarly, Vid2Robot~\citep{jain2024vid2robot} also conditions robot policy on vision tokens of a human prompt video as the global intent. However, it uses more powerful cross-attention Transformer~\citep{vaswani2017attention} to incorporate extracted global intents compared to the FiLM layer~\citep{perez2018film} of BC-Z~\citep{jang2022bc}.

\textit{(b) Phase signals generation:} When robot manipulation tasks become more complex and long-horizon, global task intents alone are often insufficient. Richer intent signals that reflect task progress become inevitable. To this end, \cite{sermanet2016unsupervised} discover implicit intermediate phases from human videos and convert them into dense perceptual rewards. Although this work is not designed for symbolic task parsing, it shows that human videos can provide task-phase signals without constructing explicit command sequences. Building upon this idea, \cite{yu2018one_hil} use phase predictors to infer task progression from an untrimmed human video, such that the robot can autonomously decide which human primitive to invoke next. Similarly, XSkill~\citep{xu2023xskill} extracts cross-embodiment skill prototypes from unlabeled videos, and specifies how these latent skills should be composed for an unseen robot task. To intuitively represent task objectives across different phases, \cite{sharma2019third} progressively generate the robot's first-person visual subgoals from a third-person human video. This enables a low-level controller to generate actions for achieving these subgoals in the robot's own environment. More recently, \cite{li2026act} extend the scope of task intents from manipulation goals to non-Markovian active perception. By introducing a cognitive auxiliary head, this method uses human videos to infer when the robot should switch behaviors between information-seeking and task-execution phases.

\textit{(c) Conclusion for task intents:} \textit{Task intents as a bridge} relax the requirement of explicitly parsing complete instruction sequences. It instead uses human videos to generate compact high-level guidance for robot control. Existing methods mainly instantiate such guidance through extracting the overall objective of a task and exploring how task progress should evolve over time. Compared with task structures, task intents are usually more flexible and easier to transfer across embodiments, since they avoid committing to a fixed symbolic decomposition. Nonetheless, elaborated robot learning policies are still required to ground the inferred intents into concrete action planning.

\subsubsection{Summary for Task-Oriented Transfer}

Task-oriented transfer provides the most embodiment-agnostic pathway for bridging human videos and robot execution. Instead of requiring a direct correspondence between observation and action, it extracts task-relevant guidance at the instruction level, where cross-embodiment transfer is more naturally feasible. From this perspective, \textit{task structures as a bridge} and \textit{task intents as a bridge} can be viewed as two complementary ways of organizing high-level knowledge from human videos. The former emphasizes explicit procedural decomposition, making the transferred knowledge more interpretable for downstream planning. The latter relaxes the need for full symbolic parsing and instead provides compact guidance signals that are often more flexible for cross-domain and long-horizon policy adaptation. Their difference essentially reflects a broader tradeoff between \textit{explicitness} and \textit{flexibility}. 

A key remaining bottleneck lies in grounding task guidance from human videos into executable robot behaviors with sufficient precision and robustness. Future progress may depend on more tightly coupling task-oriented transfer with observation-oriented and action-oriented transfer. Accordingly, high-level task semantics from human videos can be connected with embodiment-aware perception and low-level control. Such integration may ultimately determine whether task-oriented transfer remains only a planning aid or becomes a more central interface for scalable robot learning from human videos.

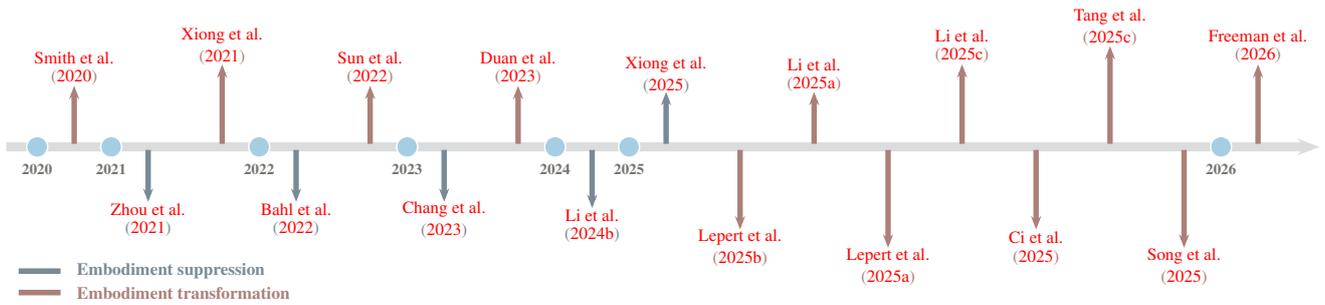
\begin{figure*}[t]
  \centering
  \resizebox{\linewidth}{!}{%
  \begin{tikzpicture}[x=1cm,y=1cm,>=Stealth]
    \definecolor{timelineblue}{RGB}{120,138,150}
    \definecolor{timelinebrick}{RGB}{171,128,120}
    \definecolor{timelinegray}{RGB}{218,220,221}
    \definecolor{timelineyear}{RGB}{118,116,112}
    \definecolor{timelinebubble}{RGB}{166,206,227}

    \draw[timelinegray,line width=4.2pt,-{Stealth[length=3.6mm]}] (-0.10,0) -- (21.20,0);

    \foreach \x/\year in {0.40/2020,1.60/2021,4.00/2022,6.40/2023,8.80/2024,10.00/2025,19.60/2026} {
      \fill[timelinebubble,draw=white,line width=0.7pt] (\x,0) circle (0.17);
      \node[font=\bfseries\scriptsize,text=timelineyear] at (\x,-0.36) {\year};
    }

    \draw[timelineblue,line width=2.5pt,-{Stealth[length=2.2mm]}] (2.20,-0.05) -- (2.20,-0.90);
    \node[align=center,font=\fontsize{8}{8.6}\selectfont,text=timelineblue] at (2.20,-1.19) {\timelinecite{zhou2021manipulator}};

    \draw[timelineblue,line width=2.5pt,-{Stealth[length=2.2mm]}] (4.60,-0.05) -- (4.60,-0.90);
    \node[align=center,font=\fontsize{8}{8.6}\selectfont,text=timelineblue] at (4.60,-1.19) {\timelinecite{bahl2022human}};

    \draw[timelineblue,line width=2.5pt,-{Stealth[length=2.2mm]}] (7.00,-0.05) -- (7.00,-0.90);
    \node[align=center,font=\fontsize{8}{8.6}\selectfont,text=timelineblue] at (7.00,-1.19) {\timelinecite{chang2023look}};

    \draw[timelineblue,line width=2.5pt,-{Stealth[length=2.2mm]}] (9.40,-0.05) -- (9.40,-1.00);
    \node[align=center,font=\fontsize{8}{8.6}\selectfont,text=timelineblue] at (9.40,-1.29) {\timelinecite{li2024ag2manip}};

    \draw[timelineblue,line width=2.5pt,-{Stealth[length=2.2mm]}] (10.60,0.05) -- (10.60,0.90);
    \node[align=center,font=\fontsize{8}{8.6}\selectfont,text=timelineblue] at (10.60,1.17) {\timelinecite{xiong2025ag2x2}};

    \draw[timelinebrick,line width=2.5pt,-{Stealth[length=2.2mm]}] (1.00,0.05) -- (1.00,1.00);
    \node[align=center,font=\fontsize{8}{8.6}\selectfont,text=timelinebrick] at (1.00,1.29) {\timelinecite{smith2020avid}};

    \draw[timelinebrick,line width=2.5pt,-{Stealth[length=2.2mm]}] (3.40,0.05) -- (3.40,1.35);
    \node[align=center,font=\fontsize{8}{8.6}\selectfont,text=timelinebrick] at (3.40,1.65) {\timelinecite{xiong2021learning}};

    \draw[timelinebrick,line width=2.5pt,-{Stealth[length=2.2mm]}] (5.80,0.05) -- (5.80,1.00);
    \node[align=center,font=\fontsize{8}{8.6}\selectfont,text=timelinebrick] at (5.80,1.29) {\timelinecite{sun2022learning}};

    \draw[timelinebrick,line width=2.5pt,-{Stealth[length=2.2mm]}] (8.20,0.05) -- (8.20,1.00);
    \node[align=center,font=\fontsize{8}{8.6}\selectfont,text=timelinebrick] at (8.20,1.29) {\timelinecite{duan2023ar2}};

    \draw[timelinebrick,line width=2.5pt,-{Stealth[length=2.2mm]}] (11.80,-0.05) -- (11.80,-1.35);
    \node[align=center,font=\fontsize{8}{8.6}\selectfont,text=timelinebrick] at (11.80,-1.65) {\timelinecite{lepert2025phantom}};

    \draw[timelinebrick,line width=2.5pt,-{Stealth[length=2.2mm]}] (13.00,0.05) -- (13.00,0.90);
    \node[align=center,font=\fontsize{8}{8.6}\selectfont,text=timelinebrick] at (13.00,1.17) {\timelinecite{li2025h2r}};

    \draw[timelinebrick,line width=2.5pt,-{Stealth[length=2.2mm]}] (14.20,-0.05) -- (14.20,-1.65);
    \node[align=center,font=\fontsize{8}{8.6}\selectfont,text=timelinebrick] at (14.20,-1.95) {\timelinecite{lepert2025masquerade}};

    \draw[timelinebrick,line width=2.5pt,-{Stealth[length=2.2mm]}] (15.40,0.05) -- (15.40,1.35);
    \node[align=center,font=\fontsize{8}{8.6}\selectfont,text=timelinebrick] at (15.40,1.65) {\timelinecite{li2025mimicdreamer}};

    \draw[timelinebrick,line width=2.5pt,-{Stealth[length=2.2mm]}] (16.60,-0.05) -- (16.60,-1.35);
    \node[align=center,font=\fontsize{8}{8.6}\selectfont,text=timelinebrick] at (16.60,-1.65) {\timelinecite{ci2025h2r}};

    \draw[timelinebrick,line width=2.5pt,-{Stealth[length=2.2mm]}] (17.80,0.05) -- (17.80,1.65);
    \node[align=center,font=\fontsize{8}{8.6}\selectfont,text=timelinebrick] at (17.80,1.95) {\timelinecite{tang2025trajectory}};

    \draw[timelinebrick,line width=2.5pt,-{Stealth[length=2.2mm]}] (19.00,-0.05) -- (19.00,-1.65);
    \node[align=center,font=\fontsize{8}{8.6}\selectfont,text=timelinebrick] at (19.00,-1.95) {\timelinecite{song2025mitty}};

    \draw[timelinebrick,line width=2.5pt,-{Stealth[length=2.2mm]}] (20.20,0.05) -- (20.20,1.35);
    \node[align=center,font=\fontsize{8}{8.6}\selectfont,text=timelinebrick] at (20.20,1.65) {\timelinecite{freeman2026warped}};

    \draw[timelineblue,line width=2.6pt,rounded corners=1pt] (0.10,-2.03) -- (0.78,-2.03);
    \node[anchor=west,font=\bfseries\footnotesize,text=timelineblue] at (0.92,-2.03) {Embodiment suppression};

    \draw[timelinebrick,line width=2.6pt,rounded corners=1pt] (0.10,-2.38) -- (0.78,-2.38);
    \node[anchor=west,font=\bfseries\footnotesize,text=timelinebrick] at (0.92,-2.38) {Embodiment transformation};

  \end{tikzpicture}%
  }
  \caption{Chronological overview of methods under \textit{transformed videos as a bridge} in Sec.~\ref{sec:transformed_videos}.}
  \label{fig:transformed_videos_timeline}
\end{figure*}

\subsection{Observation-Oriented Transfer}

Compared to task-oriented transfer, observation-oriented transfer focuses on bridging human videos and robot execution at the level of visual perception. Instead of extracting symbolic instructions or high-level intents, it aims to transform raw human videos into observation representations that are directly compatible with the robot's perception and control pipeline. Human videos and robot observations often differ significantly in viewpoint, embodiment appearance, and environmental conditions. Therefore, this category emphasizes generating transferable visual representations that align the perceptual spaces across embodiments. Specifically, we categorize existing works for observation-oriented transfer based on two criteria: (1) \textit{transformed videos as a bridge}, which directly translate human videos into embodiment-agnostic or robot-like visual formats, and (2) \textit{visual embeddings as a bridge}, which instead aim to learn embodiment-invariant latent visual representations that align human and robot observations within a shared feature space.

\begin{figure}[t]
  \centering
  \includegraphics[width=1\linewidth]{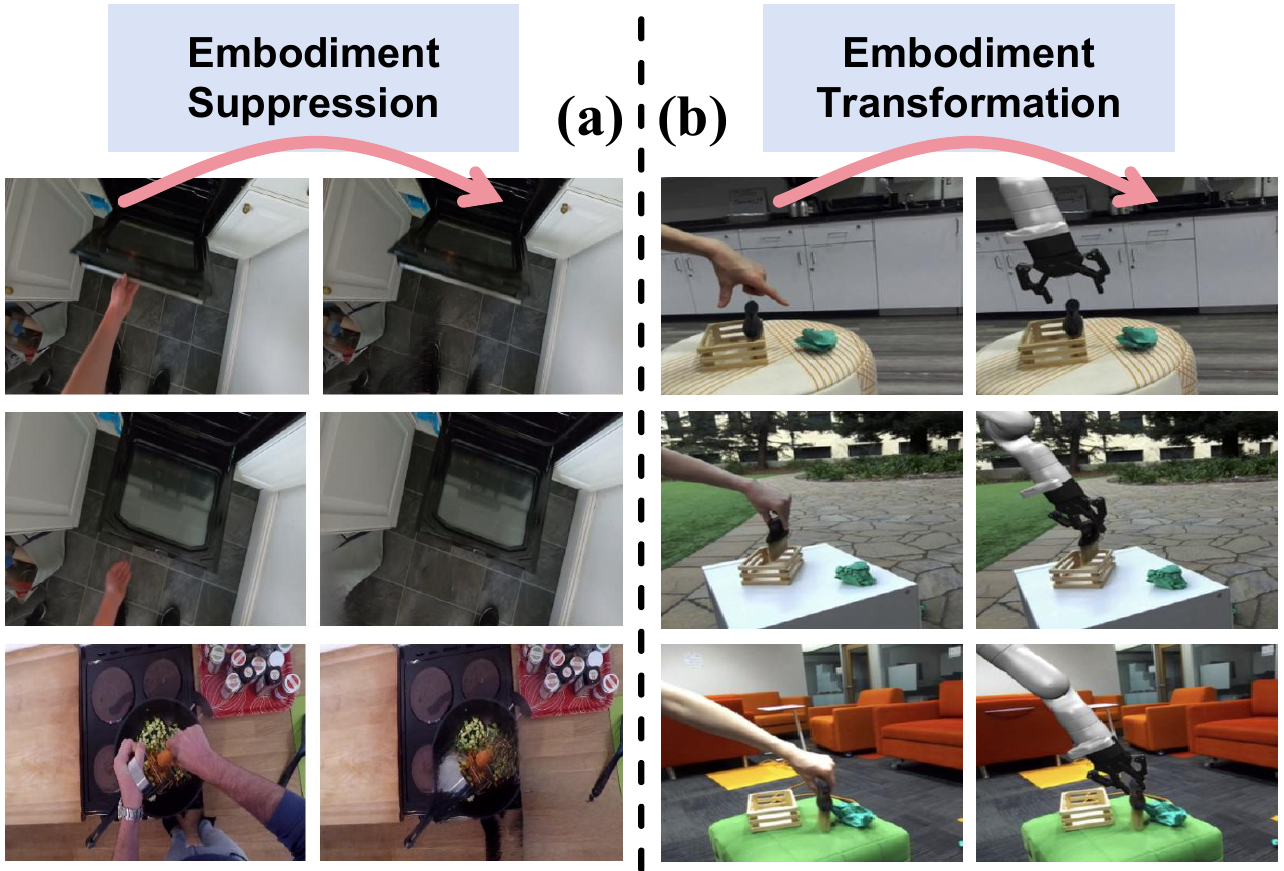}
  \caption{Illustration of \textit{transformed videos as a bridge}. Some elements are adapted from \cite{li2024ag2manip,lepert2025phantom}.}
  \label{fig:transformed_videos_transfer}
\end{figure}

\subsubsection{Transformed Videos} \label{sec:transformed_videos}
The most straightforward approach to bridge human videos and robot perception is to transform human videos into visual observations that suppress human embodiment cues (i.e., observed body appearances). The transformed observations can be further rendered with specific robot embodiments optionally. Accordingly, we divide the related methods into \textit{embodiment suppression} and \textit{embodiment transformation}, and present them using a timeline illustrated in Fig.~\ref{fig:transformed_videos_timeline}.

\textit{(a) Embodiment suppression:} Since robot observations do not contain human appearances, several works suppress embodiment-related visual cues from videos to narrow the morphological gap between human demonstrations and robot observations, as illustrated in Fig.~\ref{fig:transformed_videos_transfer}(a). This bridging mechanism is inspired by MIR~\citep{zhou2021manipulator}, which builds invisible-arm environments to focus on object state changes and learn manipulator-independent representations. For example, \cite{bahl2022human} remove the agents from both human and robot videos using an off-the-shelf video inpainting method~\citep{lee2019copy}, while maintaining object states and scene dynamics. Based on these inpainted videos, they further design an agent-agnostic alignment objective to compare human and robot executions, improving robot policy through real-world interactions. In contrast, \cite{chang2023look} go beyond simple agent removal and explicitly factorize egocentric human videos into agent and environment components, based on a single-frame inpainting model~\citep{rombach2022high}. Compared to the aforementioned work \citep{bahl2022human} that directly discards the hand signals, this agent-environment factorization provides a more structured way to suppress embodiment cues. It preserves the end-effector motion that indicates where interactions occur in egocentric views.

Interestingly, a similar evolutionary pattern can also be observed from Ag2Manip~\citep{li2024ag2manip} to Ag2x2~\citep{xiong2025ag2x2}. Ag2Manip~\citep{li2024ag2manip} segments out the human hands from videos and then combines E$^2$FGVI~\citep{li2022towards} inpainting with temporal contrastive learning to obtain agent-agnostic visual representations. In contrast, Ag2x2~\citep{xiong2025ag2x2} recognizes that completely removing human factors from videos may lose important object progression and bimanual coordination information. It therefore retains the coordinates of hand-object interaction points in the human videos and uses them jointly with visual observations for policy learning.

\textit{(b) Embodiment transformation:} Although embodiment suppression removes the visual appearance of human hands and arms from videos, robot observations during execution still contain the robot's own embodiment. To further narrow the human-robot observation gap, as shown in Fig.~\ref{fig:transformed_videos_transfer}(b), several works move beyond suppressing human embodiment cues. They additionally render robot arms into the videos after removing human appearances, thereby achieving observation-level embodiment transformation. As a pioneering work, AVID~\citep{smith2020avid} trains a CycleGAN~\citep{zhu2017unpaired} to translate human manipulation videos into robot videos, and manually selects representative goal images for each task stage to indicate completion states. However, before the emergence of diffusion models~\citep{ho2020denoising,rombach2022high}, the quality of translating human embodiment into robot embodiment was not sufficiently plausible, leading to suboptimal robot policy learning. Thus, \cite{xiong2021learning} observe that directly extracting state representations from translated images as in AVID~\citep{smith2020avid} is susceptible to visual artifacts. To address this issue, they further add a keypoint detection stage after video translation based on MUNIT~\citep{huang2018multimodal}, and use salient regions rather than the entire image to construct rewards for reinforcement learning. \cite{sun2022learning} instead integrate embodiment transformation into the imitation learning paradigm. They use the predicted keypoint trajectories together with the transformed robot images as expert demonstration data for policy training. 

With the development of generative modeling, the underlying video translation paradigm has evolved from the aforementioned GAN-based stylization to diffusion-based embodiment transformation with inpainting techniques. For example, Phantom~\citep{lepert2025phantom} removes human arms through E$^2$FGVI~\citep{li2022towards} like Ag2Manip~\citep{li2024ag2manip}, and further overlays rendered robots in their place to produce robot-consistent observations. Despite obtaining high-quality rendering results, it needs the fixed third-person camera setups to synthesize the curated exocentric robot viewpoint. This assumption limits robustness when dealing with the large amount of egocentric human activity videos with diverse viewpoints available on the Internet. 
To solve this problem, Masquerade~\citep{lepert2025masquerade} places a virtual bimanual robot in a robot camera coordinate system with known intrinsics and extrinsics, and drives the robot end-effectors to track the recovered human hand motion. It then renders robot observations from the original viewpoint, reducing the viewpoint discrepancy between in-the-wild egocentric human videos and robot execution.
Besides, considering in-the-wild human videos are usually not equipped with corresponding metric depth, \cite{lepert2025masquerade} further use the 2D keypoint locations as supervisory labels for an auxiliary loss in their vision model. This circumvents the requirement of using depth observations to refine HaMeR like Phantom~\citep{lepert2025phantom}. They also demonstrate that their co-training pipeline enables robust robot policies on complex long-horizon bimanual tasks, compared to their concurrent work H2R~\citep{li2025h2r} only using viewpoint-aligned videos for vision encoder pretraining. MimicDreamer~\citep{li2025mimicdreamer} proposes a more systematic human-to-robot video alignment framework that jointly addresses viewpoint alignment, action alignment, and visual alignment. The resulting transformed videos and aligned action trajectories are directly used to train a $\pi_0$ VLA model~\citep{black2024pi_0}. More recently, WARPED~\citep{freeman2026warped} further pushes rendering-based embodiment transformation toward wrist-aligned observations. It combines hand-object tracking and trajectory retargeting with Gaussian Splatting to synthesize photorealistic robot wrist-view observations and aligned actions. Different from these rendering-based approaches, H2R-Grounder~\citep{ci2025h2r} formulates embodiment transformation as a paired-data-free video generation problem. It learns to decompose both human and robot videos into a shared representation consisting of manipulator pose cues and cleaned background videos. Then, a video diffusion model is fine-tuned given a minimal pose indicator and background, to synthesize temporally aligned robot videos from human demonstrations. This scheme avoids the floating and misalignment artifacts common in direct robot rendering~\citep{lepert2025masquerade,li2025h2r}.
To facilitate easier downstream applications, \cite{duan2023ar2} develop an iOS application that lets users record robot-consistent manipulation demonstrations in augmented reality. The recorded demonstrations can be directly used for the following behavior cloning~\citep{shridhar2023perceiver}, significantly improving the deployment efficiency.

In contrast to the above methods that use transformed videos directly for robot policy learning, Mitty~\citep{song2025mitty} additionally exploits them to train an end-to-end human-to-robot video transformation model. This paradigm yields better generalization to unseen tasks and environments. Similar to Mitty~\citep{song2025mitty}, \cite{tang2025trajectory} also focus on end-to-end transformation modeling. Furthermore, they introduce sparse optical flow trajectories as an additional generation condition, leading to plausible visual quality and trajectory controllability. 

\textit{(c) Conclusion for transformed videos:} \textit{Transformed videos as a bridge} provide the most direct way to reduce the visual discrepancy between human videos and robot observations. Existing methods have mainly developed along two parallel directions: embodiment suppression, which removes human-specific appearance cues while preserving interaction-relevant scene information, and embodiment transformation, which further synthesizes robot-consistent visual observations from human videos. The rapid progress of video generation models has strongly facilitated the development of this bridging mechanism. Nonetheless, these methods still rely heavily on the quality of video inpainting and rendering. Consequently, generation artifacts in geometric inconsistency and viewpoint mismatch may still propagate into downstream robot policy learning, especially in unconstrained in-the-wild settings.

\begin{figure*}[t]
  \centering
  \resizebox{\linewidth}{!}{%
  \begin{tikzpicture}[x=1cm,y=1cm,>=Stealth]
    \definecolor{timelineblue}{RGB}{120,138,150}
    \definecolor{timelinebrick}{RGB}{171,128,120}
    \definecolor{timelinegray}{RGB}{218,220,221}
    \definecolor{timelineyear}{RGB}{118,116,112}
    \definecolor{timelinebubble}{RGB}{166,206,227}

    \draw[timelinegray,line width=4.2pt,-{Stealth[length=3.6mm]}] (-0.10,0) -- (25.50,0);

    \fill[timelinebubble,draw=white,line width=0.7pt] (0.64,0) circle (0.17);
    \node[font=\bfseries\scriptsize,text=timelineyear] at (0.64,-0.36) {2017};
    
    \fill[timelinebubble,draw=white,line width=0.7pt] (1.36,0) circle (0.17);
    \node[font=\bfseries\scriptsize,text=timelineyear] at (1.36,-0.36) {2018};
    
    \fill[timelinebubble,draw=white,line width=0.7pt] (2.55,0) circle (0.17);
    \node[font=\bfseries\scriptsize,text=timelineyear] at (2.55,-0.36) {2020};
    
    \fill[timelinebubble,draw=white,line width=0.7pt] (3.70,0) circle (0.17);
    \node[font=\bfseries\scriptsize,text=timelineyear] at (3.70,0.36) {2021};
    
    \fill[timelinebubble,draw=white,line width=0.7pt] (4.30,0) circle (0.17);
    \node[font=\bfseries\scriptsize,text=timelineyear] at (4.33,-0.36) {2022};
    
    \fill[timelinebubble,draw=white,line width=0.7pt] (7.30,0) circle (0.17);
    \node[font=\bfseries\scriptsize,text=timelineyear] at (7.33,-0.36) {2023};
    
    \fill[timelinebubble,draw=white,line width=0.7pt] (13.90,0) circle (0.17);
    \node[font=\bfseries\scriptsize,text=timelineyear] at (13.80,0.36) {2024};
    
    \fill[timelinebubble,draw=white,line width=0.7pt] (17.50,0) circle (0.17);
    \node[font=\bfseries\scriptsize,text=timelineyear] at (17.50,-0.36) {2025};
    
    \fill[timelinebubble,draw=white,line width=0.7pt] (24.70,0) circle (0.17);
    \node[font=\bfseries\scriptsize,text=timelineyear] at (24.70,-0.36) {2026};

    \draw[timelineblue,line width=2.5pt,-{Stealth[length=2.2mm]}] (1.00,0.05) -- (1.00,0.85);
    \node[align=center,font=\fontsize{6.0}{7.2}\selectfont,text=timelineblue] at (0.9,1.15) {\timelinecite{sermanet2017time}};

    \draw[timelineblue,line width=2.5pt,-{Stealth[length=2.2mm]}] (1.60,0.05) -- (1.60,1.30);
    \node[align=center,font=\fontsize{6.0}{7.2}\selectfont,text=timelineblue] at (1.60,1.62) {\timelinecite{rothfuss2018deep}};

    \draw[timelinebrick,line width=2.5pt,-{Stealth[length=2.2mm]}] (2.20,-0.05) -- (2.20,-0.85);
    \node[align=center,font=\fontsize{6.0}{7.2}\selectfont,text=timelinebrick] at (2.20,-1.15) {\timelinecite{liu2018imitation}};

    \draw[timelineblue,line width=2.5pt,-{Stealth[length=2.2mm]}] (2.80,0.05) -- (2.80,1.75);
    \node[align=center,font=\fontsize{6.0}{7.2}\selectfont,text=timelineblue] at (2.80,2.09) {\timelinecite{schmeckpeper2020reinforcement}};

    \draw[timelinebrick,line width=2.5pt,-{Stealth[length=2.2mm]}] (3.40,-0.05) -- (3.40,-1.30);
    \node[align=center,font=\fontsize{6.0}{7.2}\selectfont,text=timelinebrick] at (3.40,-1.62) {\timelinecite{smith2020avid}};

    \draw[timelinebrick,line width=2.5pt,-{Stealth[length=2.2mm]}] (4.00,-0.05) -- (4.00,-0.85);
    \node[align=center,font=\fontsize{6.0}{7.2}\selectfont,text=timelinebrick] at (4.00,-1.15) {\timelinecite{chen2021learning}};

    \draw[timelineblue,line width=2.5pt,-{Stealth[length=2.2mm]}] (4.60,0.05) -- (4.60,0.85);
    \node[align=center,font=\fontsize{6.0}{7.2}\selectfont,text=timelineblue] at (4.60,1.15) {\timelinecite{nair2022r3m}};

    \draw[timelineblue,line width=2.5pt,-{Stealth[length=2.2mm]}] (5.20,0.05) -- (5.20,1.30);
    \node[align=center,font=\fontsize{6.0}{7.2}\selectfont,text=timelineblue] at (5.20,1.62) {\timelinecite{xiao2022masked}};

    \draw[timelineblue,line width=2.5pt,-{Stealth[length=2.2mm]}] (5.80,0.05) -- (5.80,1.75);
    \node[align=center,font=\fontsize{6.0}{7.2}\selectfont,text=timelineblue] at (5.80,2.09) {\timelinecite{ma2022vip}};

    \draw[timelinebrick,line width=2.5pt,-{Stealth[length=2.2mm]}] (6.40,-0.05) -- (6.40,-1.30);
    \node[align=center,font=\fontsize{6.0}{7.2}\selectfont,text=timelinebrick] at (6.40,-1.62) {\timelinecite{zakka2022xirl}};

    \draw[timelinebrick,line width=2.5pt,-{Stealth[length=2.2mm]}] (7.00,-0.05) -- (7.00,-1.75);
    \node[align=center,font=\fontsize{6.0}{7.2}\selectfont,text=timelinebrick] at (7.00,-2.09) {\timelinecite{xiong2022robotube}};

    \draw[timelineblue,line width=2.5pt,-{Stealth[length=2.2mm]}] (7.60,0.05) -- (7.60,0.85);
    \node[align=center,font=\fontsize{6.0}{7.2}\selectfont,text=timelineblue] at (7.60,1.15) {\timelinecite{ma2023liv}};

    \draw[timelineblue,line width=2.5pt,-{Stealth[length=2.2mm]}] (8.20,0.05) -- (8.20,1.30);
    \node[align=center,font=\fontsize{6.0}{7.2}\selectfont,text=timelineblue] at (8.20,1.62) {\timelinecite{bhateja2023robotic}};

    \draw[timelineblue,line width=2.5pt,-{Stealth[length=2.2mm]}] (8.80,0.05) -- (8.80,1.75);
    \node[align=center,font=\fontsize{6.0}{7.2}\selectfont,text=timelineblue] at (8.5,2.09) {\timelinecite{radosavovic2023real}};

    \draw[timelineblue,line width=2.5pt,-{Stealth[length=2.2mm]}] (9.40,0.05) -- (9.40,2.20);
    \node[align=center,font=\fontsize{6.0}{7.2}\selectfont,text=timelineblue] at (9.40,2.56) {\timelinecite{karamcheti2023language}};

    \draw[timelineblue,line width=2.5pt,-{Stealth[length=2.2mm]}] (10.00,0.05) -- (10.00,0.85);
    \node[align=center,font=\fontsize{6.0}{7.2}\selectfont,text=timelineblue] at (10.00,1.15) {\timelinecite{dasari2023unbiased}};

    \draw[timelineblue,line width=2.5pt,-{Stealth[length=2.2mm]}] (10.60,0.05) -- (10.60,1.30);
    \node[align=center,font=\fontsize{6.0}{7.2}\selectfont,text=timelineblue] at (10.30,1.62) {\timelinecite{majumdar2023we}};

    \draw[timelineblue,line width=2.5pt,-{Stealth[length=2.2mm]}] (11.20,0.05) -- (11.20,1.75);
    \node[align=center,font=\fontsize{6.0}{7.2}\selectfont,text=timelineblue] at (11.20,2.09) {\timelinecite{burns2023makes}};

    \draw[timelineblue,line width=2.5pt,-{Stealth[length=2.2mm]}] (11.80,0.05) -- (11.80,2.20);
    \node[align=center,font=\fontsize{6.0}{7.2}\selectfont,text=timelineblue] at (11.80,2.56) {\timelinecite{wu2023unleashing}};

    \draw[timelinebrick,line width=2.5pt,-{Stealth[length=2.2mm]}] (12.0,-0.05) -- (12.0,-1.30);
    \node[align=center,font=\fontsize{6.0}{7.2}\selectfont,text=timelinebrick] at (12.0,-1.62) {\timelinecite{mendonca2023structured}};

    \draw[timelinebrick,line width=2.5pt,-{Stealth[length=2.2mm]}] (12.8,-0.05) -- (12.8,-0.85);
    \node[align=center,font=\fontsize{6.0}{7.2}\selectfont,text=timelinebrick] at (12.8,-1.15) {\timelinecite{chane2023learning}};

    \draw[timelinebrick,line width=2.5pt,-{Stealth[length=2.2mm]}] (13.60,-0.05) -- (13.60,-1.75);
    \node[align=center,font=\fontsize{6.0}{7.2}\selectfont,text=timelinebrick] at (13.60,-2.09) {\timelinecite{huo2023efficient}};

    \draw[timelineblue,line width=2.5pt,-{Stealth[length=2.2mm]}] (14.20,0.05) -- (14.20,0.85);
    \node[align=center,font=\fontsize{6.0}{7.2}\selectfont,text=timelineblue] at (14.20,1.15) {\timelinecite{jain2024vid2robot}};

    \draw[timelinebrick,line width=2.5pt,-{Stealth[length=2.2mm]}] (14.80,-0.05) -- (14.80,-1.30);
    \node[align=center,font=\fontsize{6.0}{7.2}\selectfont,text=timelinebrick] at (14.80,-1.62) {\timelinecite{qian2024contrast}};

    \draw[timelineblue,line width=2.5pt,-{Stealth[length=2.2mm]}] (15.40,0.05) -- (15.40,1.30);
    \node[align=center,font=\fontsize{6.0}{7.2}\selectfont,text=timelineblue] at (15.40,1.62) {\timelinecite{li2024ag2manip}};

    \draw[timelineblue,line width=2.5pt,-{Stealth[length=2.2mm]}] (16.00,0.05) -- (16.00,1.75);
    \node[align=center,font=\fontsize{6.0}{7.2}\selectfont,text=timelineblue] at (16.00,2.09) {\timelinecite{liu2024masked}};

    \draw[timelineblue,line width=2.5pt,-{Stealth[length=2.2mm]}] (16.60,0.05) -- (16.60,0.85);
    \node[align=center,font=\fontsize{6.0}{7.2}\selectfont,text=timelineblue] at (16.60,1.15) {\timelinecite{zeng2024learning}};

    \draw[timelineblue,line width=2.5pt,-{Stealth[length=2.2mm]}] (17.20,0.05) -- (17.20,2.20);
    \node[align=center,font=\fontsize{6.0}{7.2}\selectfont,text=timelineblue] at (17.20,2.56) {\timelinecite{cheang2024gr}};

    \draw[timelineblue,line width=2.5pt,-{Stealth[length=2.2mm]}] (17.80,0.05) -- (17.80,1.30);
    \node[align=center,font=\fontsize{6.0}{7.2}\selectfont,text=timelineblue] at (17.80,1.62) {\timelinecite{sun2025vtao}};

    \draw[timelineblue,line width=2.5pt,-{Stealth[length=2.2mm]}] (18.40,0.05) -- (18.40,1.75);
    \node[align=center,font=\fontsize{6.0}{7.2}\selectfont,text=timelineblue] at (18.40,2.09) {\timelinecite{shah2025mimicdroid}};

    \draw[timelineblue,line width=2.5pt,-{Stealth[length=2.2mm]}] (19.00,0.05) -- (19.00,2.20);
    \node[align=center,font=\fontsize{6.0}{7.2}\selectfont,text=timelineblue] at (19.00,2.56) {\timelinecite{zhang2025generative}};

    \draw[timelineblue,line width=2.5pt,-{Stealth[length=2.2mm]}] (19.60,0.05) -- (19.60,1.30);
    \node[align=center,font=\fontsize{6.0}{7.2}\selectfont,text=timelineblue] at (19.60,1.62) {\timelinecite{jiang2025rynnvla}};

    \draw[timelineblue,line width=2.5pt,-{Stealth[length=2.2mm]}] (20.20,0.05) -- (20.20,1.75);
    \node[align=center,font=\fontsize{6.0}{7.2}\selectfont,text=timelineblue] at (20.20,2.09) {\timelinecite{pai2025mimic}};

    \draw[timelineblue,line width=2.5pt,-{Stealth[length=2.2mm]}] (20.80,0.05) -- (20.80,0.85);
    \node[align=center,font=\fontsize{6.0}{7.2}\selectfont,text=timelineblue] at (20.80,1.15) {\timelinecite{xiong2025ag2x2}};

    \draw[timelineblue,line width=2.5pt,-{Stealth[length=2.2mm]}] (21.40,0.05) -- (21.40,1.30);
    \node[align=center,font=\fontsize{6.0}{7.2}\selectfont,text=timelineblue] at (21.40,1.62) {\timelinecite{zhou2025mitigating}};

    \draw[timelineblue,line width=2.5pt,-{Stealth[length=2.2mm]}] (22.00,0.05) -- (22.00,1.75);
    \node[align=center,font=\fontsize{6.0}{7.2}\selectfont,text=timelineblue] at (22.00,2.09) {\timelinecite{kedia2025one}};

    \draw[timelineblue,line width=2.5pt,-{Stealth[length=2.2mm]}] (22.60,0.05) -- (22.60,2.20);
    \node[align=center,font=\fontsize{6.0}{7.2}\selectfont,text=timelineblue] at (22.60,2.56) {\timelinecite{punamiya2025egobridge}};

    \draw[timelineblue,line width=2.5pt,-{Stealth[length=2.2mm]}] (23.20,0.05) -- (23.20,1.30);
    \node[align=center,font=\fontsize{6.0}{7.2}\selectfont,text=timelineblue] at (23.20,1.62) {\timelinecite{liu2025immimic}};

    \draw[timelineblue,line width=2.5pt,-{Stealth[length=2.2mm]}] (23.80,0.05) -- (23.80,1.75);
    \node[align=center,font=\fontsize{6.0}{7.2}\selectfont,text=timelineblue] at (23.80,2.09) {\timelinecite{zhu2025learning}};

    \draw[timelinebrick,line width=2.5pt,-{Stealth[length=2.2mm]}] (24.40,-0.05) -- (24.40,-1.30);
    \node[align=center,font=\fontsize{6.0}{7.2}\selectfont,text=timelinebrick] at (24.40,-1.62) {\timelinecite{goswami2025world}};

    \draw[timelineblue,line width=2.5pt,-{Stealth[length=2.2mm]}] (25.00,0.05) -- (25.00,2.20);
    \node[align=center,font=\fontsize{6.0}{7.2}\selectfont,text=timelineblue] at (25.00,2.56) {\timelinecite{ye2026visual}};

    \draw[timelineblue,line width=2.6pt,rounded corners=1pt] (0.10,-2.03) -- (0.78,-2.03);
    \node[anchor=west,font=\bfseries\footnotesize,text=timelineblue] at (0.92,-2.03) {Visual pretraining};

    \draw[timelinebrick,line width=2.6pt,rounded corners=1pt] (0.10,-2.38) -- (0.78,-2.38);
    \node[anchor=west,font=\bfseries\footnotesize,text=timelinebrick] at (0.92,-2.38) {Visual guidance};

  \end{tikzpicture}%
  }
  \caption{Chronological overview of methods under \textit{visual embeddings as a bridge} in Sec.~\ref{sec:visual_embeddings}.}
  \label{fig:visual_embeddings_timeline}
\end{figure*}
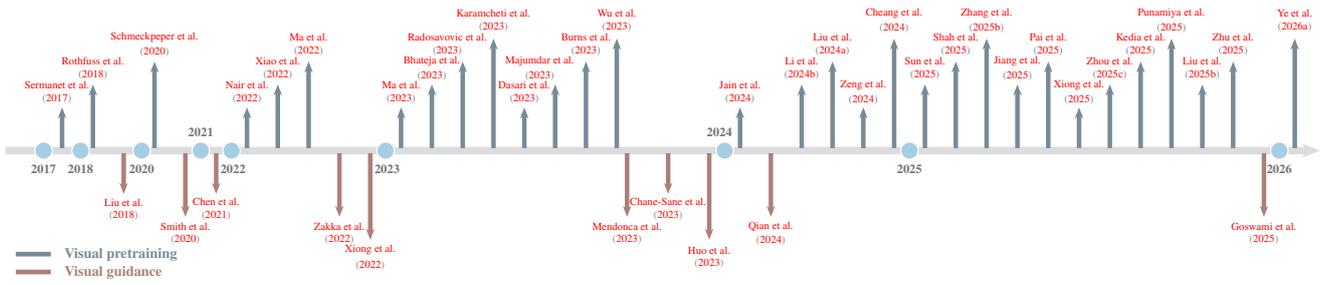

\begin{figure*}[t]
  \centering
  \includegraphics[width=0.95\linewidth]{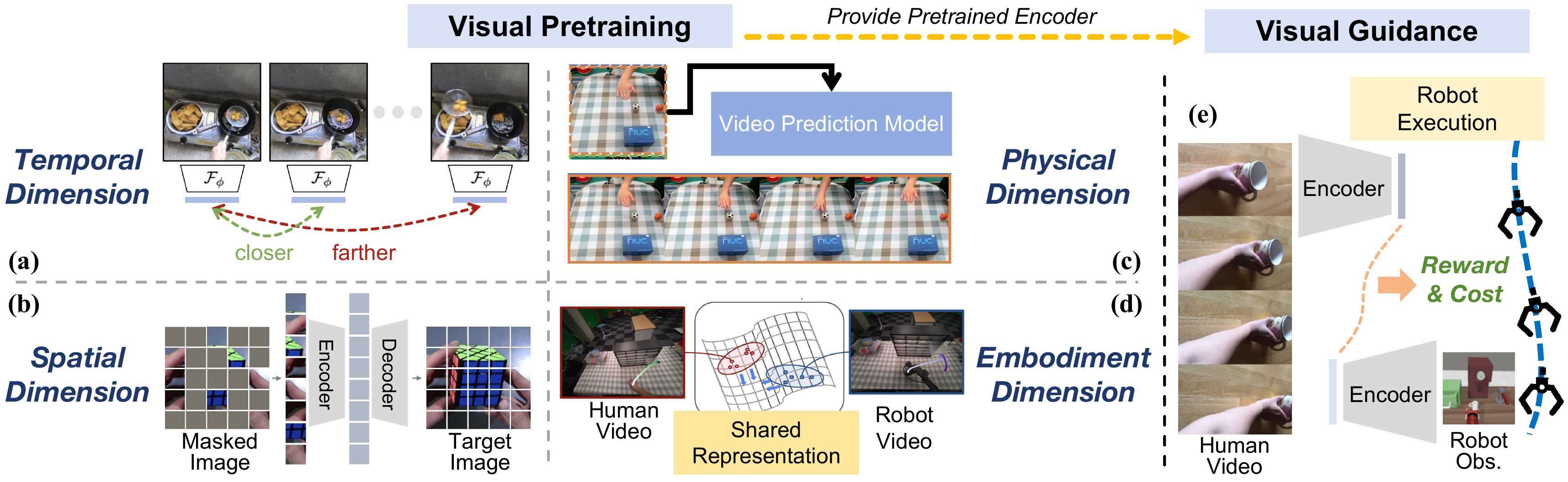}
  \caption{High-level diagram of \textit{visual embeddings as a bridge}. Some elements of the illustrations are adapted from \cite{nair2022r3m,xiao2022masked,zhu2025learning,punamiya2025egobridge,chen2021learning}.}
  \label{fig:visual_embeddings_transfer}
\end{figure*}

\subsubsection{Visual Embeddings} \label{sec:visual_embeddings}
To avoid the explicit generation artifacts introduced by transformed videos, an increasing number of studies focus instead on compressing human videos into visual embeddings. These embeddings can be used to pretrain a perception model for downstream robot policies or to construct visual signals that guide robot exploration. Accordingly, we divide \textit{visual embeddings as a bridge} into two progressive categories: \textit{visual pretraining} and \textit{visual guidance}. The timeline of related work is depicted in Fig.~\ref{fig:visual_embeddings_timeline}.

\textit{(1) Visual pretraining:} Temporal contrastive learning (TCL) is one of the most commonly used strategies for pretraining perception models in the temporal dimension. In the early stage, \cite{sermanet2017time} propose the pioneering framework that employs the TCL pipeline on human videos. They treat visual embeddings of temporally distinct frames from the same sequence as negative samples, leading to viewpoint-invariant visual representations for downstream robot learning. However, this negative-sample construction may fragment the semantic continuity of human-object interactions, making it less suitable for learning single-view robot policies. Instead, the following works~\citep{nair2022r3m,ma2022vip,ma2023liv} pull visual embeddings from adjacent frames closer while pushing apart embeddings from frames that are farther apart along the time axis, as illustrated in Fig.~\ref{fig:visual_embeddings_transfer}(a). Thus, the perception models can learn from human videos how visual states transit through the fine-grained progress of temporal interactions. For example, R3M~\citep{nair2022r3m} achieves this TCL pipeline on the large-scale Ego4D dataset~\citep{grauman2022ego4d} together with video-language alignment and feature sparsity regularization. It produces a reusable frozen visual encoder to generate visual embeddings for robotic manipulation. Owing to its strong generalization across tasks and environments, R3M has inspired multiple foundational perception models in the subsequent robot learning policies~\citep{shaw2023videodex,mendonca2023structured,li2024ag2manip,xiong2025ag2x2}. Compared to R3M which additionally requires video textual descriptions to align its representation, VIP~\citep{ma2022vip} is developed as a fully self-supervised learning paradigm. It further models temporal reachability between visual embeddings through a value-implicit objective. LIV~\citep{ma2023liv} extends the VIP paradigm~\citep{ma2022vip} to multimodal value function learning, simultaneously supporting both language goals and image goals. 
\cite{bhateja2023robotic} further evolve the TCL paradigm of VIP~\citep{ma2022vip} into a temporal-difference learning (TDL) framework, yielding an intent-conditioned value function (ICVF). By modeling the reachability from the current state to a variety of future goal states, the learned value function captures scene dynamics as a transferable visual representation. To capture embodiment-agnostic visual embeddings during pretraining, \cite{li2024ag2manip} and \cite{xiong2025ag2x2} both utilize inpainted human videos to achieve temporal contrastive learning, encouraging the encoder to focus on interaction-relevant scene dynamics rather than embodiment-specific appearance cues.

As can be noted, temporal contrastive learning attends to constraining the visual embeddings in the temporal dimension. In contrast, the masked pretraining (MP) paradigm focuses on the image reconstruction task (see Fig.~\ref{fig:visual_embeddings_transfer}(b)) in the spatial dimension to pretrain the vision encoders. The related works of this paradigm are substantially inspired by the optimization principle of masked autoencoders~\citep{he2022masked}. For instance, \cite{xiao2022masked} propose MVP to build the foundation of the MP paradigm by pretraining visual encoders with masked image reconstruction on large-scale human videos. They freeze these encoders for downstream motor control, demonstrating that spatially masked pretraining alone can yield transferable visual representations for robot learning. \cite{radosavovic2023real} further extend MVP~\citep{xiao2022masked} by scaling both the pretraining human video data and model size, and broaden its downstream application to large-scale imitation learning for real-world robotic manipulation. By carefully reproducing MVP~\citep{xiao2022masked} and R3M~\citep{nair2022r3m}, \cite{karamcheti2023language} demonstrate that introducing language supervision like R3M into the masked pretraining paradigm of MVP can compensate for the semantic limitations while preserving the strong spatial modeling capability for visual embedding generation. In contrast to \cite{karamcheti2023language} using the language modality to assist the vision modality, M$2$VTP by \cite{liu2024masked} incorporates tactile signals in the reconstruction mechanism for complementary interaction cues. It aims to address cases where vision is occluded or fine local geometry is difficult to perceive visually. Based on M$2$VTP~\citep{liu2024masked}, VTAO-BiManip~\citep{sun2025vtao} further introduces hand action and object-centric understanding. It extends the masked pretraining of M$2$VTP from a visual-tactile framework to a visual-tactile-action framework. Recently, \cite{ye2026visual} shift the focus from object understanding to capturing deeper human-like manipulation patterns. They are inspired by inferior parietal lobule (IPL) neurons of the human brain, and propose an IPL token for multisensory integration via a Transformer encoder~\citep{vaswani2017attention} in the masked pretraining process. Given the integrated representation, they train a multitask action policy by RL for expert policies of different tasks and distill these experts into a unified policy by online imitation learning. 
Compared with the above MP paradigms, which optimize perception models through image reconstruction, MimicDroid~\citep{shah2025mimicdroid} directly uses masked images as inputs for hand forecasting pretraining. It implicitly reduces overfitting to human-specific visual cues and encourages the model to focus on the environment and object interactions.

Building on the TCL and MP paradigms, some researchers further delve into what types of existing model architectures and pretraining data can yield more effective visual embeddings for downstream robotic manipulation. For example, \cite{majumdar2023we} turn attention to discovering which visual pretraining paradigm is most effective. They systematically benchmark TCL and MP methods to show that no single model is universally dominant across all robotic tasks. They further train multiple ViT models~\citep{dosovitskiy2020image} with different scales using MAE and finally obtain VC-1, a unified visual encoder that outperforms prior pretraining methods on average. \cite{dasari2023unbiased} further shift the question from which visual pretraining to use toward which data to use. They showcase that the image distribution of the dataset matters more than its scale for downstream visuomotor learning. They also observe that even traditional vision datasets like ImageNet~\citep{deng2009imagenet}, Kinetics~\citep{smaira2020short}, and 100DOH~\citep{shan2020understanding} can be surprisingly competitive for robotic visual pretraining.
In contrast to \cite{dasari2023unbiased} and \cite{majumdar2023we}, \cite{burns2023makes} attend to visual generalization capability specifically and advancing metrics that are predictive of generalization. They identify a recipe for strong generalization: ViT models with high emergent segmentation accuracy tend to generate more generalizable visual embeddings under visual distribution shifts. Besides, emergent segmentation accuracy is a stronger predictor of generalization than many other robustness metrics, without the requirement of an additional training process.

As can be noted, the TCL paradigm captures visual representations by enforcing similarity relations in the temporal dimension, but it does not necessarily require the encoder to model the deeper physical mechanisms underlying human-object interactions. The MP paradigm emphasizes spatial understanding through reconstruction, but most reconstruction targets can be recovered from static texture and appearance cues within a single frame, without reasoning about potential interaction dynamics. These limitations motivate another line of work that learns visual embeddings through human video prediction (HVP), as illustrated in Fig.~\ref{fig:visual_embeddings_transfer}(c). By predicting future observations from past interactions, such methods encourage the vision encoder to capture not only spatiotemporal regularities, but also the causality of physical dynamics that are more directly relevant to downstream robotic manipulation. 
The HVP paradigm is closely related to the notion of \textit{world models} in recent literature, since world models are fundamentally concerned with predicting how future observations evolve from past interactions and action-relevant dynamics~\citep{liao2025genie,ye2026world}. Notably, here we only review those works explicitly using human videos for visual pretraining.
An early work by \cite{rothfuss2018deep} pretrains the visual encoder by predicting future human video frames and uses the resulting visual embeddings to retrieve the most similar human demonstration episode from an episodic memory. The retrieved demonstration is then used to extract object motion for low-level robot actions. However, the predicted frames are often visually blurry, which may limit the quality of the learned visual representations and consequently impair the effectiveness of visual pretraining. In contrast, \cite{zeng2024learning} avoid long-horizon video generation via predicting only the transition and final frames. By emphasizing the key interaction states, their MPI framework encourages the vision encoder to learn the core causal structure and state evolution of manipulation. They empirically validate that MPI outperforms several popular TCL and MP paradigms~\citep{nair2022r3m,xiao2022masked,karamcheti2023language} for visual pretraining. Benefiting from the development of generative techniques and the scaling of model capacity, more recent works have revisited the direct prediction of continuous video clips as a richer supervision signal for visual pretraining. For example, GVF-TAPE~\citep{zhang2025generative} uses human videos as supplementary data to pretrain an end-to-end video prediction model based on rectified flow~\citep{liu2022flow}. 
In contrast to GVF-TAPE~\citep{zhang2025generative} which only performs embodiment-specific video prediction, \cite{zhu2025learning} further introduce cross-prediction video generation for cross-embodiment forecasting. That is, they are dedicated to predicting robot videos from human images and human videos from robot images. As a result, the learned visual embeddings become a more unified representation for capturing task semantics and environmental context across embodiments. Recently, researchers have also found that integrating the human video prediction paradigm into VLA models can facilitate their generation of more relevant and appropriate actions by enabling the anticipation of future events. As exemplified by \cite{wu2023unleashing}, GR-1 integrates large-scale video generative pretraining into a GPT-style VLA that predicts both future images and robot actions in an end-to-end manner, improving action generation performance. GR-2~\citep{cheang2024gr} further scales the amount of human video generative pretraining data by nearly 50 times compared with GR-1~\citep{wu2023unleashing} (from 0.8 million to 38 million), leading to stronger generalization capabilities in diverse unseen scenarios. To bridge the gap between purely human video prediction and action generation, RynnVLA-001~\citep{jiang2025rynnvla} proposes an intermediate stage that incorporates wrist trajectory prediction to learn the association between visual changes and their underlying human motion. Instead of training the video prediction model from scratch, mimic-video~\citep{pai2025mimic} directly uses Cosmos-Predict2~\citep{agarwal2025cosmos,ali2025world} as the video generation backbone and conditions a flow-matching action decoder on its visual representations.

Although the HVP paradigm attends to the physical dimension by pretraining vision encoders with the human video prediction objective, the resulting visual embeddings still encode human interaction behaviors rather than robot interaction states. This inevitably induces a potential morphological gap since only human data are available during visual pretraining~\citep{mower2026robot}. To further mitigate this issue in the embodiment dimension, some works integrate robot data into the visual pretraining process of human videos. As shown in Fig.~\ref{fig:visual_embeddings_transfer}(d), this paradigm implements joint domain adaptation (JDA) to bring the visual embeddings of semantically analogous human and robot observations closer together in a shared representation space. Therefore, the pretrained vision encoders are more suitable for producing visual embeddings in downstream robotic tasks. \cite{schmeckpeper2020reinforcement} propose adversarial domain confusion to learn a domain-invariant encoder that maps human video frames and robot observations into a shared feature space. They further show that a small amount of paired human-robot data helps anchor the visual embedding alignment across embodiments. The subsequent JDA works typically focus on directly decreasing the distance between visual embeddings of human videos and robot observations. For instance, Vid2Robot~\citep{jain2024vid2robot} introduce a video alignment loss based on temporal cycle consistency, encouraging the vision encoder to generate visual embeddings invariant to embodiments and environmental factors. \cite{kedia2025one} also learn a shared vision encoder by pulling matched visual embeddings closer according to the optimal transport distance between paired human and robot snippets. To further mitigate the problem of temporal inconsistency and kinematic variations from different embodiments, \cite{punamiya2025egobridge} introduce dynamic time warping (DTW)~\citep{sakoe1978dynamic} to achieve optimal transport for visual representation alignment. Similar to \cite{punamiya2025egobridge}, \cite{liu2025immimic} also use DTW to pair each human timestep
with its best-matching robot timestep according to visual embeddings. They additionally apply MixUp~\citep{zhang2017mixup} to generate interpolated human data for the following co-training process. \cite{zhou2025mitigating} further let the task description feature interact with visual embeddings of human and robot videos, and implement task-aware contrastive learning across embodiments. 

\textit{(b) Visual guidance:} As shown in Fig.~\ref{fig:visual_embeddings_transfer}(e), another line of research shifts the focus from vision encoder pretraining to how visual embeddings derived from human videos can directly construct reward or cost functions that guide robot interaction with the environment. Encoders obtained through \textit{visual pretraining} can often be directly reused within the \textit{visual guidance} paradigm. \cite{liu2018imitation} formulate the reinforcement learning objective by penalizing the squared Euclidean distance between the robot's current visual embedding and the demonstration embedding translated from human videos. Similarly, XIRL~\citep{zakka2022xirl} defines the reinforcement learning reward as the negative distance to the goal state in the learned visual embedding space. A related one-shot setting is explored by~\cite{huo2023efficient}, who extract a state alignment-based reward from a single observation-only demonstration video and combine it with sampling-based MPC for efficient policy optimization in contact-rich fabric manipulation. RoboTube~\citep{xiong2022robotube} complements the visual guidance paradigm by providing a curated human video dataset together with RT-sim, a suite of simulated twin environments that closely match the appearance and dynamics of the real scenes. It enables visual guidance learned from human videos to be benchmarked reproducibly in simulation and more reliably transferred to real robots. \cite{qian2024contrast} further parse visual embeddings into interaction-aware temporal relations through IAAFormer. The alignment distance reflects the cross-embodiment task process at the sequence level. It provides more structured visual guidance for downstream robot adaptation. \cite{mendonca2023structured} further learn an affordance-grounded latent space from human videos, where distances between latent states can also serve as planning-oriented visual guidance. Rather than relying only on instantaneous visual embedding similarity for reward design, it models how these affordance-centric visual states evolve under actions through a world model, enabling more effective downstream control. \cite{goswami2025world} also follow the world model scheme and propose DexWM, a dexterous world model trained on large-scale human videos. At test time, DexWM performs goal-conditioned MPC using a planning cost defined by the visual state distance and hand-keypoint pixel differences.

In contrast to these methods that use feature distance as visual guidance, \cite{smith2020avid} further convert visual embeddings into a binary classifier that determines whether the input image matches the designated goal image, to construct the reward function with the output log probabilities. Similar to \cite{smith2020avid}, \cite{chen2021learning} develop DVD, which learns a binary classifier on top of visual embeddings to quantify task-wise functional similarity as rewards. They explicitly focus on leveraging in-the-wild human videos along with modest robot demos to improve the generalizability of the reward function, compared to \cite{smith2020avid} only considering in-domain data. \cite{chane2023learning} follow the DVD model~\citep{chen2021learning} to construct the reward function but learn the functional similarity without any annotated robot videos. 

\textit{(c) Conclusion for visual embeddings:} \textit{Visual embeddings as a bridge} avoid the explicit inpainting and rendering errors of transformed videos by mapping human observations into latent representations that are more readily transferable to robot learning. Existing methods have mainly developed along two complementary directions. Visual pretraining learns reusable perception models from human videos through temporal contrastive learning, masked pretraining, human video prediction, and joint domain adaptation. Visual guidance further uses visual embeddings to define reward and cost for downstream robot interaction. Compared with transformed videos, this bridging mechanism is generally more compact and robust to appearance gaps across embodiments. However, whether the learned representations preserve task-relevant interaction dynamics and cross-embodiment consistency sufficiently to support reliable downstream action planning remains largely implicit.

\subsubsection{Summary for Observation-Oriented Transfer}

Observation-oriented transfer addresses one of the most fundamental challenges in LfHV, i.e., how to reduce the perceptual gap between human videos and robot observations. Compared with task-oriented transfer which bridges embodiments through high-level task semantics, this category operates closer to the visuomotor interface. It therefore plays a more direct role in determining whether the robot can interpret human demonstrations in a manner compatible with downstream control. From this perspective, \textit{transformed videos as a bridge} and \textit{visual embeddings as a bridge} represent two complementary solutions to the same problem. The former explicitly modifies human observations into embodiment-agnostic or robot-consistent visual formats, making the transferred information more intuitive and directly usable. The latter instead compresses human observations into shared latent spaces, sacrificing visual explicitness in exchange for greater compactness, robustness, and flexibility. Their difference essentially reflects a tradeoff between \textit{explicit perceptual alignment} and \textit{implicit representation alignment}. Transformed videos provide a more interpretable bridge, whereas visual embeddings offer a more scalable and generalizable interface.

A clear trend in this category is the shift from appearance-level alignment toward deeper modeling of interaction dynamics and cross-embodiment consistency. Early works mainly focused on suppressing embodiment-specific appearance cues or learning generic visual features from human videos. More recent methods increasingly incorporate temporal prediction, joint human-robot alignment, and world model objectives. This evolution suggests that effective observation transfer requires matching visual appearance while also capturing how interactions unfold over time across embodiments. In light of this, the central challenge of observation-oriented transfer lies in preserving action-relevant dynamics while removing embodiment-specific distractors. Therefore, future progress may depend on tighter integration between transformed videos and visual embeddings. It may also require stronger coupling to action-oriented transfer, such that perceptual alignment can better support executable and physically meaningful robot behavior.

\subsection{Action-Oriented Transfer}

In contrast to task-oriented transfer and observation-oriented transfer, the action-oriented transfer bridges human videos and robot execution more directly at the level of action planning. It attends to extracting action-relevant information from human videos like interaction affordances and latent actions, and then transferring them into robot policies as executable guidance. Accordingly, we organize this category into two forms of information flow: \textit{affordances as a bridge}, which expose explicit geometric action cues from human demonstration videos, and \textit{latent actions as a bridge}, which instead compress observed human behaviors into transferable action abstractions for downstream robot policy learning.

\begin{figure*}[t]
  \centering
  \includegraphics[width=1\linewidth]{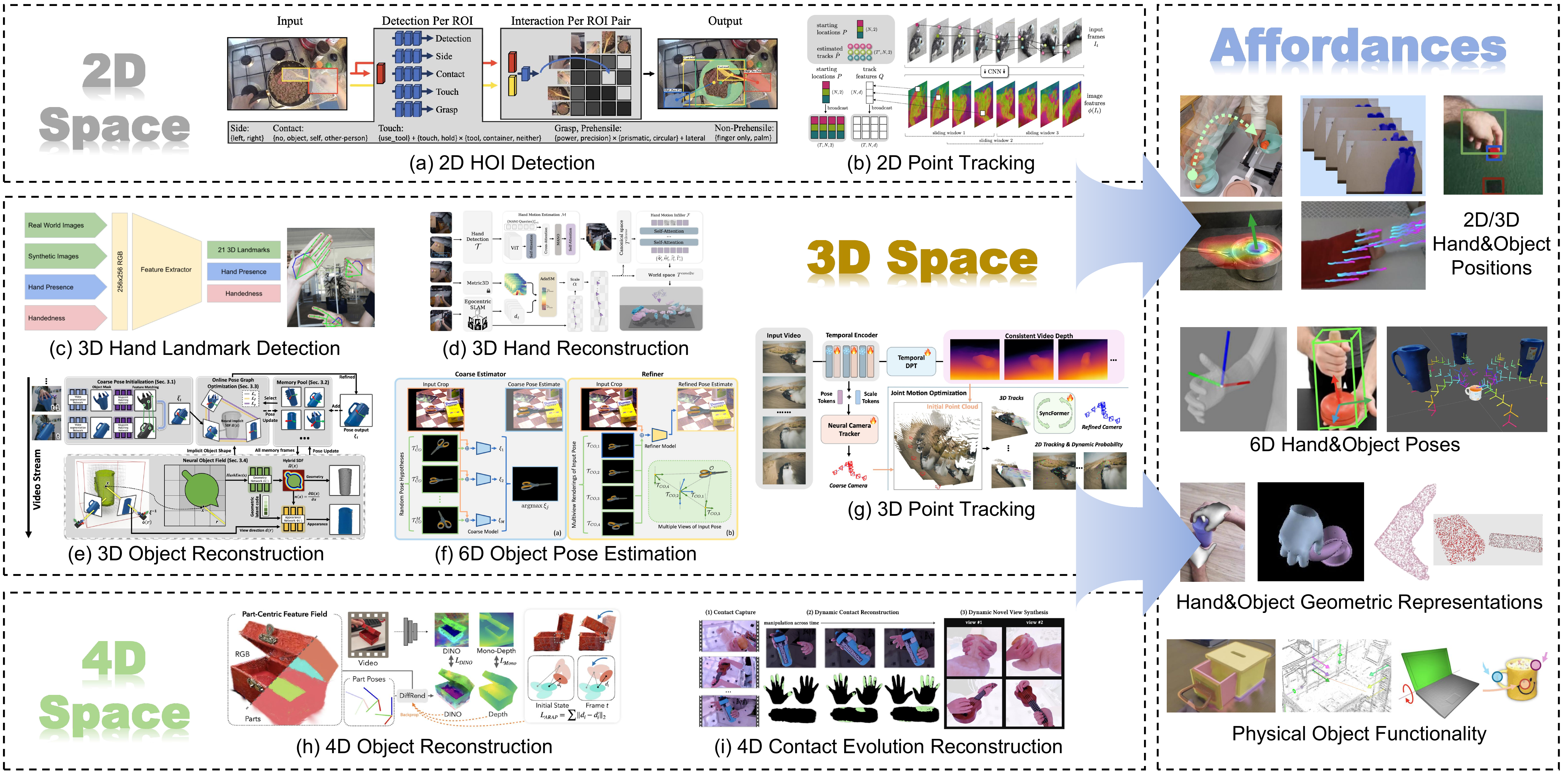}
  \caption{Illustration of HOI analysis techniques and extracted affordances from human videos. Part of images are taken from \cite{zhang2020mediapipe,labbe2022megapose,sivakumar2022robotic,qin2022dexmv,cheng2023towards,wen2023bundlesdf,wang2023mimicplay,kumar2023graph,bahl2023affordances,wen2023any,karaev2024cotracker,bahety2024screwmimic,kerr2024robot,zhang2025hawor,xiao2025spatialtrackerv2,chen2025visa,hsieh2025dexman,hsu2025spot,chen2025web2grasp,zhou2025you,cong2025dytact,wang2026paws}.}
  \label{fig:hoi_affordances}
\end{figure*}

\subsubsection{Affordances} \label{sec:affordances}
\textit{affordances as a bridge} are more directly grounded in action compared with task instructions and visual embeddings. As introduced by \cite{bahl2023affordances}, affordances from human videos explicitly specify \textit{where} and \textit{how} the hand-object interaction (HOI) should occur. That is, when manipulating an object, affordances explicitly indicate where on the object to make contact, and how to make this contact~\citep{kannan2023deft}. Therefore, in this survey, we broadly instantiate \textit{affordances} as all possible interaction-related geometric and functional signals that can be extracted from human videos, including 2D/3D hand\&object positions (e.g., trajectories, motion flow, contact/separation regions), 6D hand\&object poses (e.g., grasping poses, sequential rigid transformations), hand\&object geometric representations (e.g., meshes, point clouds, hand parametric representations), and physical object functionality (e.g., 4D object parts, articulations, functional keypoints). We will detail how these affordances facilitate action-oriented transfer in the following sections.

\textit{(a) HOI analysis:} Hand-object interactions in human videos fundamentally facilitate affordance extraction. Therefore, before delving into specific affordance-based bridging mechanisms, we first review how existing works analyze hand-object interactions to recover action-relevant cues from human videos.

Firstly, we introduce the widely-used HOI detection methods that operate in the 2D space, as illustrated in Fig.~\ref{fig:hoi_affordances}(a)-(b). \cite{shan2020understanding} establish a detection paradigm by jointly inferring 2D hand location, hand side, contact state, and the bounding box of the object in contact from Internet human videos. Their pretrained HOI detector has subsequently served as a practical foundation for extracting hand-contact cues in diverse following LfHV works~\citep{bahl2022human,bahl2023affordances,wang2023mimicplay,kumar2023graph,mendonca2023structured,srirama2024hrp,ju2024robo,jonnavittula2025view,lu2025visual}. Building on such 2D hand-contact analysis, \cite{goyal2022human} further treat human hands as probes for interactive object understanding. By leveraging hand appearance and motion in egocentric human videos, they learn object affordances in the form of regions of interaction and afforded grasps. This shifts the focus from contact detection to richer 2D affordance discovery for downstream robot policy learning. Its augmentation strategies like color jittering and cropping have also inspired subsequent LfHV works such as 2HandedAfforder~\citep{heidinger20252handedafforder}. \cite{cheng2023towards} further extend HOI analysis to a richer contact vocabulary that distinguishes touching, holding, and using. It has also been adopted as an off-the-shelf segmenter in several LfHV works~\citep{heppert2024ditto,shan2025slot,wang2026paws}. The visual foundation models like GLIP~\citep{li2022grounded}, GroundingDINO~\citep{liu2024grounding}, YOLO series~\citep{Ultralytics2023,khanam2024yolov5,sohan2024review,cheng2024yolo}, and SAM series~\cite{kirillov2023segment,zhang2023faster,ren2024grounded,ravi2024sam,carion2025sam} are also widely used by LfHV works to extract hand-object bounding boxes and segmentation masks from human videos in a generalizable manner~\citep{xu2024flow,wang2024vlm,ma2025madiff,yoshida2025developing,ma2025uni,ma2025egoloc,li2025h2r,chen2025visa,hsu2025spot,xiong2025ag2x2}.

In addition to the above 2D HOI region extraction, 2D point tracking methods have also become an important technical component in affordance-based bridging mechanisms. They are basically used to generate object motion flow and further recover the rigid transformation. For example, the CoTracker series~\citep{karaev2024cotracker,karaev2025cotracker3} jointly track large sets of query points across video frames, producing temporally consistent dense trajectories over long video horizons even under occlusion situations. This makes them particularly attractive for LfHV research~\citep{bharadhwaj2024track2act,li2024okami,yuan2024general,werby2025articulated,chen2025visa,hsu2025spot,guzey2025bridging,chen2025graphmimic,haldar2025point,liu2025egozero,ci2025h2r,tang2025functo,guzey2025dexterity,shi2026care}, because egocentric human videos inherently involve frequent mutual occlusion between hands and manipulated objects. TAPIR~\citep{doersch2023tapir} and its enhanced version~\citep{doersch2024bootstap} provide similarly general-purpose alternatives by tracking arbitrary query points through per-frame matching with temporal refinement. Their ability to maintain temporally coherent trajectories also makes them well-suited for estimating hand-object motion flow from human demonstrations in some LfHV works~\citep{xu2024flow,papagiannis2025r,chen2025graphmimic,spiridonov2025generalist}. LocoTrack~\citep{cho2024local} is also a promising 2D point tracking method, which incorporates bidirectional correspondence, matching smoothness, and compact temporal aggregation. It thus holds solid robustness to repetitive textures and homogeneous regions for robot learning from human videos~\citep{hu2025learning}.

As human-object interactions fundamentally take place in 3D real-world environments, 2D HOI analysis alone is often insufficient to fully characterize the underlying interactions. We therefore review popular HOI analysis works for 3D hand-object reconstruction and pose estimation. We begin with widely-used and promising algorithms for 3D human hand detection and reconstruction in the LfHV literature, as illustrated in Fig.~\ref{fig:hoi_affordances}(c)-(d).
As a pioneering work for in-the-wild 3D hand detection, MediaPipe~\citep{zhang2020mediapipe} emphasizes lightweight real-time 3D hand landmark (keypoint) estimation from monocular RGB through palm detection and landmark regression. Its efficiency makes it popular in LfHV for extracting scalable hand-centric affordances from unconstrained human videos~\citep{arunachalam2022dexterous,qin2022one,gao2023k,gu2023rt,zhou2025human,lu2025visual,haldar2025point}. Beyond MediaPipe's hand landmark extraction, FrankMocap~\citep{rong2020frankmocap} estimates 3D hand and body motion from monocular RGB and integrates them into a unified parametric representation to produce the hand mesh structure. It provides a practical tool for recovering 3D hand kinematic cues from human videos as affordances in the LfHV literature~\citep{patel2022learning,mandikal2022dexvip,sivakumar2022robotic,kannan2023deft,srirama2024hrp,bahety2024screwmimic,chen2024vlmimic,chen2025fmimic}. HaMeR~\citep{pavlakos2024reconstructing} further attends to higher-fidelity monocular 3D hand mesh recovery with a fully Transformer-based architecture. It substantially improves robustness and accuracy on in-the-wild hand reconstruction and pose estimation via scaling up the model architecture and data consumption. This makes it more suitable for LfHV settings that require richer geometric information or high-quality hand poses as affordances than sparse landmarks alone~\citep{li2024okami,shi2025zeromimic,papagiannis2025r,lum2025crossing,liu2025egozero,hsieh2025dexman,luo2025being,guzey2025dexterity,chen2026dexterous}. To address the inherent misalignments and incorrect poses of HaMeR's end-to-end regression, WiLoR~\citep{potamias2025wilor} introduces an additional refinement layer that deforms hand pose using mesh-aligned multi-scale features. It supports efficient multi-hand reconstruction and smooth monocular video tracking, which is useful for large-scale affordance extraction from human videos~\citep{zhou2025you,yuan2025hermes,zhu2025learning,shah2025mimicdroid}. While HaMeR and WiLoR mainly recover hand geometry in the camera frame, HaWoR~\citep{zhang2025hawor} further targets world-space hand motion reconstruction from egocentric videos. It decouples camera-space hand recovery from world-space camera trajectory estimation, and further introduces motion infilling for out-of-view frames. As a result, it produces temporally coherent global hand trajectories that are more directly applicable to action-oriented transfer in modern LfHV~\citep{li2025scalable,luo2026being,luo2026joint,feng2025spatial,yang2026aoe}. More recently, EgoHandICL~\citep{xie2026egohandicl} has emerged as a promising hand reconstruction method by introducing in-context learning with VLM-guided exemplar retrieval, which improves semantic alignment and robustness under challenging egocentric HOI conditions. In addition, SAM 3D Body (3DB)~\citep{yang2026sam} extends this line from hand-centric reconstruction to promptable full-body mesh recovery, jointly estimating body, feet, and hands with strong in-the-wild generalization. It thus offers a richer source of whole-body affordance cues for future LfHV works. The extended work of \cite{ma2025uni} attempts to integrate 3DB into the action-oriented transfer paradigm, which has been open-sourced in their repository.

Next, we introduce representative object reconstruction and pose estimation methods in the LfHV literature, as illustrated in Fig.~\ref{fig:hoi_affordances}(e)-(f). In general, obtaining a reliable object reconstruction is a prerequisite for accurate object pose estimation. BundleSDF~\citep{wen2023bundlesdf} is one of the most foundational object reconstruction methods for manipulated objects appearing in human videos. It couples a Neural Object Field with pose graph optimization. This enhances its stability regarding pose changes, occlusions, untextured surfaces, and specular highlights when deployed on LfHV works~\citep{chen2024vlmimic,patel2025robotic,hsu2025spot} using human videos of highly variable quality. Afterwards, more reconstruction methods like InstantMesh~\citep{xu2024instantmesh}, MeshyAI~\citep{meshyai}, and TRELLIS~\citep{xiang2025structured} only require a single image to directly generate object meshes for masked object regions within the target human video frame~\citep{chen2025web2grasp,ye2025video2policy,hsieh2025dexman,soraki2026objectforesight,chen2026dexterous}. They significantly improve the efficiency of task-specific object reconstruction by eliminating the need for continuous scanning.

Object pose estimation can be further performed based on the reconstructed objects. Given a CAD model and a region of interest, MegaPose~\citep{labbe2022megapose} can be directly applied in LfHV works~\citep{liang2024dreamitate,patel2025robotic} to estimate the pose of a novel object in coarse-to-fine refinement without retraining. Over the past two years, FoundationPose~\citep{wen2024foundationpose} has become a more popular object pose estimation method, as it unifies both model-based and model-free novel-object pose estimation and tracking within a single framework. Compared with MegaPose, it further leverages a neural implicit representation for novel-view synthesis together with large-scale synthetic training, enabling stronger generalization and more robust pose tracking in challenging LfHV settings~\citep{chen2024vlmimic,yuan2025hermes,hsieh2025dexman,mao2025robot,lum2025crossing,hsu2025spot,patel2025robotic,soraki2026objectforesight,fan2026robopaint,zou2026activeglasses}. More recently, Any6D~\citep{lee2025any6d} also relaxes the dependence on reconstructed CAD models and introduces a full-to-partial matching strategy. It jointly aligns 2D appearance, 3D geometry, and metric scale through a render-and-compare procedure adapted from FoundationPose~\citep{wen2024foundationpose}. This provides a promising alternative for object scale and pose estimation~\citep{hsieh2025dexman}, especially when high-quality object reconstruction is difficult to obtain.

To achieve more generalizable model-free 6D object pose estimation, some researchers~\citep{bharadhwaj2024track2act,zhu2024vision,haldar2025point,liu2025egozero,tang2025functo} manually lift the aforementioned 2D point tracking into 3D space with depth information or stereo triangulation, and then calculate the rigid transformation from the resulting 3D object flow. Nonetheless, there are off-the-shelf 3D point-tracking methods that can be used to directly generate 3D object flow for pose estimation. As shown in Fig.~\ref{fig:hoi_affordances}(g), SpatialTracker~\citep{xiao2024spatialtracker} and SpatialTrackerV2~\citep{xiao2025spatialtrackerv2} integrate video depth estimation into a unified Transformer-based tracking framework, enabling the direct prediction of temporally consistent 3D point trajectories from human videos~\citep{hsieh2025dexman,yoshida2025generating,yoshida2025developing}.

As shown in Fig.~\ref{fig:hoi_affordances}(h)-(i), attention in HOI analysis has recently begun to shift from static 3D detection and reconstruction toward 4D affordance extraction, where temporal evolution is modeled together with spatial structure. For example, 4D object reconstruction recovers object or part motion over time from human interaction videos, such as part-centric trajectories and articulation states~\citep{kerr2024robot,wang2026paws}. Besides, 4D contact evolution reconstruction estimates how hand-object contact regions temporally change during manipulation~\citep{cong2025dytact}. By incorporating temporal information, these 4D representations provide richer affordance cues than static poses or contact regions, making them a promising direction for extracting physically grounded and execution-feasible affordances from human videos.

As can be noted, a broad ecosystem of off-the-shelf HOI analysis techniques has emerged to support effective affordance extraction from human videos. In particular, 2D methods including HOI detection and point tracking provide scalable and robust tools for identifying interaction regions and estimating object motion flow under challenging conditions such as occlusions. Complementarily, 3D reconstruction and pose estimation methods further lift these interaction cues into spatially grounded representations, enabling the recovery of hand kinematics, object geometry, and rigid transformations. The emerging 4D reconstruction further jointly enriches temporal information and physical constraints in affordances. Together, these plug-and-play 2D, 3D, and 4D HOI analysis approaches form a practical technical foundation for efficiently extracting affordances in action-oriented transfer from human videos to robot policies. Consistent with the scope of this survey, we do not review HOI analysis schemes based on motion-capture devices, since our focus is specifically on transfer from human videos rather than instrumented capture setups.

\begin{figure}[t]
  \centering
  \includegraphics[width=1\linewidth]{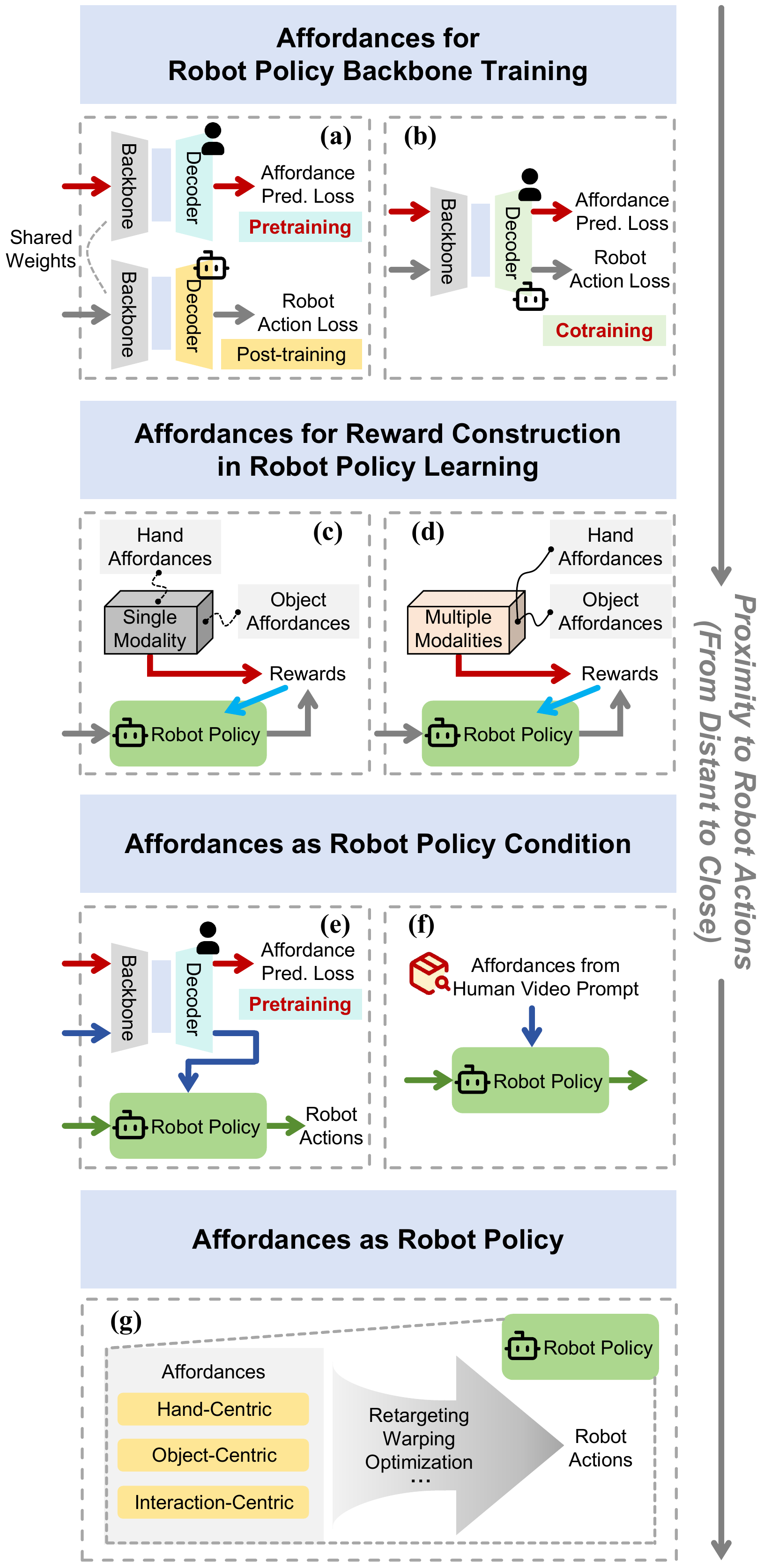}
  \caption{High-level diagram of \textit{affordances as a bridge}.}
  \label{fig:affordances_transfer}
\end{figure}

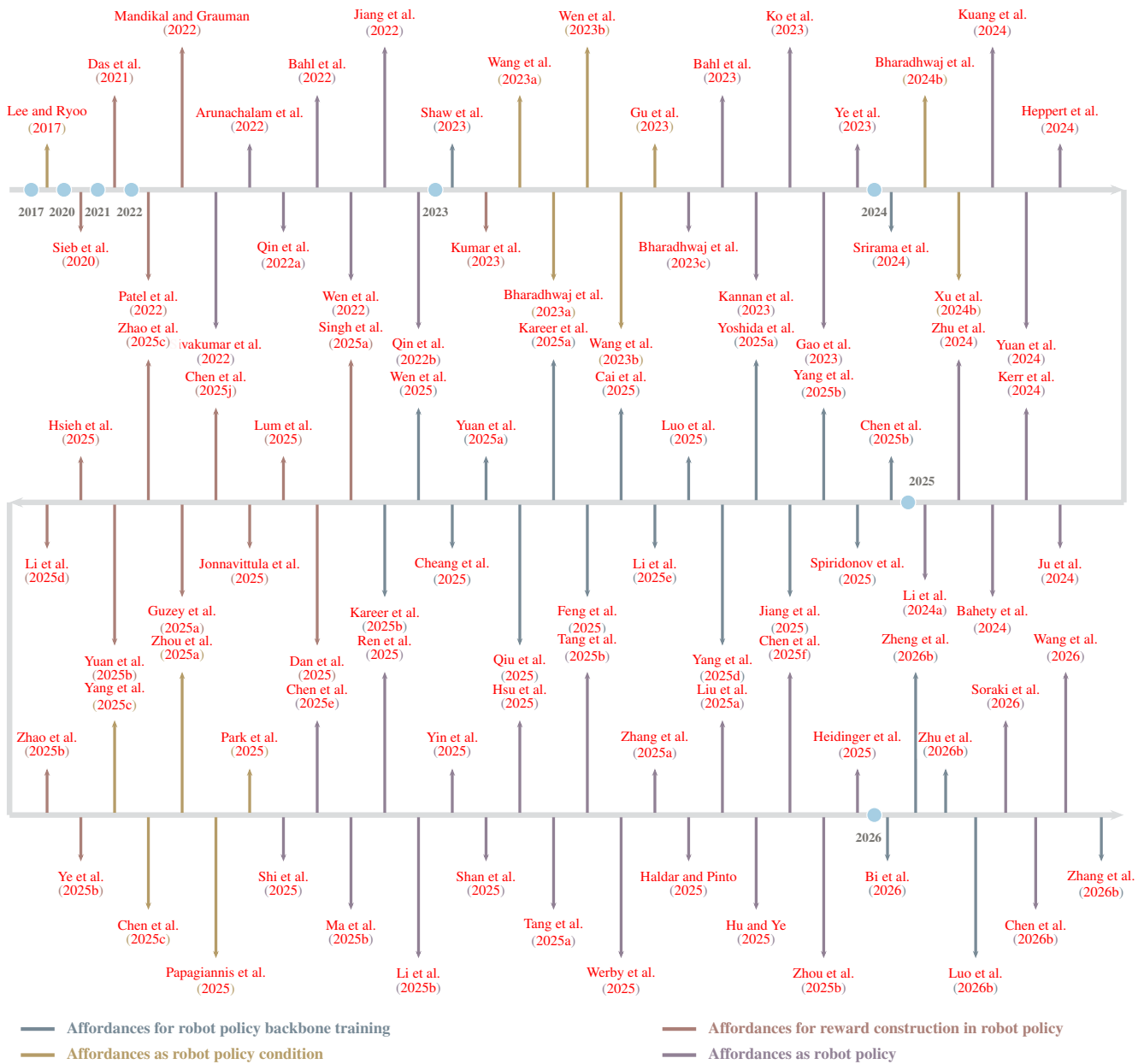
\begin{figure*}[t]
  \centering
  \resizebox{\linewidth}{!}{%
  \begin{tikzpicture}[x=1cm,y=1cm,>=Stealth]
    \definecolor{timelineblue}{RGB}{120,138,150}
    \definecolor{timelinebrick}{RGB}{171,128,120}
    \definecolor{timelinegold}{RGB}{181,156,104}
    \definecolor{timelineplum}{RGB}{145,129,149}
    \definecolor{timelinegray}{RGB}{218,220,221}
    \definecolor{timelineyear}{RGB}{118,116,112}
    \definecolor{timelinebubble}{RGB}{166,206,227}

    \newcommand{\TopEvent}[4]{%
      \pgfmathsetmacro{\starty}{(#3 > 0) ? 0.06 : -0.06}
      \pgfmathsetmacro{\labely}{#3 + ((#3 > 0) ? 0.62 : -0.62)}
      \draw[#2,line width=2.15pt,-{Stealth[length=2.0mm]}] (#1,\starty) -- (#1,#3);
      \node[align=center,font=\fontsize{10.8}{12.0}\selectfont,text=#2,fill=white,fill opacity=0.96,text opacity=1,inner sep=1.0pt,rounded corners=1pt]
      at (#1,\labely) {\timelinecite{#4}};
    }
    \newcommand{\MidEvent}[4]{%
      \pgfmathsetmacro{\starty}{(#3 > -8.40) ? -8.34 : -8.46}
      \pgfmathsetmacro{\labely}{#3 + ((#3 > -8.40) ? 0.62 : -0.62)}
      \draw[#2,line width=2.15pt,-{Stealth[length=2.0mm]}] (#1,\starty) -- (#1,#3);
      \node[align=center,font=\fontsize{10.8}{12.0}\selectfont,text=#2,fill=white,fill opacity=0.96,text opacity=1,inner sep=1.0pt,rounded corners=1pt]
      at (#1,\labely) {\timelinecite{#4}};
    }
    \newcommand{\BotEvent}[4]{%
      \pgfmathsetmacro{\starty}{(#3 > -16.80) ? -16.74 : -16.86}
      \pgfmathsetmacro{\labely}{#3 + ((#3 > -16.80) ? 0.62 : -0.62)}
      \draw[#2,line width=2.15pt,-{Stealth[length=2.0mm]}] (#1,\starty) -- (#1,#3);
      \node[align=center,font=\fontsize{10.8}{12.0}\selectfont,text=#2,fill=white,fill opacity=0.96,text opacity=1,inner sep=1.0pt,rounded corners=1pt]
      at (#1,\labely) {\timelinecite{#4}};
    }

    \draw[timelinegray,line width=4.2pt,-{Stealth[length=3.6mm]}] (-0.10,0) -- (29.60,0);
    \draw[timelinegray,line width=4.2pt] (29.60,0) -- (29.60,-8.40);
    \draw[timelinegray,line width=4.2pt,{Stealth[length=3.6mm]}-] (-0.10,-8.40) -- (29.60,-8.40);
    \draw[timelinegray,line width=4.2pt] (-0.10,-8.40) -- (-0.10,-16.80);
    \draw[timelinegray,line width=4.2pt,-{Stealth[length=3.6mm]}] (-0.10,-16.80) -- (29.60,-16.80);

    \fill[timelinebubble,draw=white,line width=0.8pt] (0.48,0) circle (0.20);
    \node[font=\bfseries\fontsize{9.8}{10.8}\selectfont,text=timelineyear] at (0.48,-0.60) {2017};

    \fill[timelinebubble,draw=white,line width=0.8pt] (1.35,0) circle (0.20);
    \node[font=\bfseries\fontsize{9.8}{10.8}\selectfont,text=timelineyear] at (1.3,-0.60) {2020};

    \fill[timelinebubble,draw=white,line width=0.8pt] (2.25,0) circle (0.20);
    \node[font=\bfseries\fontsize{9.8}{10.8}\selectfont,text=timelineyear] at (2.25,-0.60) {2021};

    \fill[timelinebubble,draw=white,line width=0.8pt] (3.15,0) circle (0.20);
    \node[font=\bfseries\fontsize{9.8}{10.8}\selectfont,text=timelineyear] at (3.1,-0.60) {2022};

    \fill[timelinebubble,draw=white,line width=0.8pt] (11.25,0) circle (0.20);
    \node[font=\bfseries\fontsize{9.8}{10.8}\selectfont,text=timelineyear] at (11.25,-0.60) {2023};

    \fill[timelinebubble,draw=white,line width=0.8pt] (22.95,0) circle (0.20);
    \node[font=\bfseries\fontsize{9.8}{10.8}\selectfont,text=timelineyear] at (22.95,-0.60) {2024};

    \fill[timelinebubble,draw=white,line width=0.8pt] (23.85,-8.40) circle (0.20);
    \node[font=\bfseries\fontsize{9.8}{10.8}\selectfont,text=timelineyear] at (24.22,-7.82) {2025};

    \fill[timelinebubble,draw=white,line width=0.8pt] (22.95,-16.80) circle (0.20);
    \node[font=\bfseries\fontsize{9.8}{10.8}\selectfont,text=timelineyear] at (22.80,-17.40) {2026};

    \TopEvent{0.90}{timelinegold}{1.25}{lee2017learning}
    \TopEvent{1.80}{timelinebrick}{-1.15}{sieb2020graph}
    \TopEvent{2.70}{timelinebrick}{2.55}{das2021model}
    \TopEvent{3.60}{timelinebrick}{-2.45}{patel2022learning}
    \TopEvent{4.50}{timelinebrick}{3.85}{mandikal2022dexvip}
    \TopEvent{5.40}{timelineplum}{-3.75}{sivakumar2022robotic}
    \TopEvent{6.30}{timelineplum}{1.25}{arunachalam2022dexterous}
    \TopEvent{7.20}{timelineplum}{-1.15}{qin2022one}
    \TopEvent{8.10}{timelineplum}{2.55}{bahl2022human}
    \TopEvent{9.00}{timelineplum}{-2.45}{wen2022you}
    \TopEvent{9.90}{timelineplum}{3.85}{jiang2022ditto}
    \TopEvent{10.80}{timelineplum}{-3.75}{qin2022dexmv}
    \TopEvent{11.70}{timelineblue}{1.25}{shaw2023videodex}
    \TopEvent{12.60}{timelinebrick}{-1.15}{kumar2023graph}
    \TopEvent{13.50}{timelinegold}{2.55}{wang2023mimicplay}
    \TopEvent{14.40}{timelinegold}{-2.45}{bharadhwaj2023towards}
    \TopEvent{15.30}{timelinegold}{3.85}{wen2023any}
    \TopEvent{16.20}{timelinegold}{-3.75}{wang2023robot}
    \TopEvent{17.10}{timelinegold}{1.25}{gu2023rt}
    \TopEvent{18.00}{timelineplum}{-1.15}{bharadhwaj2023zero}
    \TopEvent{18.90}{timelineplum}{2.55}{bahl2023affordances}
    \TopEvent{19.80}{timelineplum}{-2.45}{kannan2023deft}
    \TopEvent{20.70}{timelineplum}{3.85}{ko2023learning}
    \TopEvent{21.60}{timelineplum}{-3.75}{gao2023k}
    \TopEvent{22.50}{timelineplum}{1.25}{ye2023learning}
    \TopEvent{23.40}{timelineblue}{-1.15}{srirama2024hrp}
    \TopEvent{24.30}{timelinegold}{2.55}{bharadhwaj2024track2act}
    \TopEvent{25.20}{timelinegold}{-2.45}{xu2024flow}
    \TopEvent{26.10}{timelineplum}{3.85}{kuang2024ram}
    \TopEvent{27.00}{timelineplum}{-3.75}{yuan2024general}
    \TopEvent{27.90}{timelineplum}{1.25}{heppert2024ditto}

    \MidEvent{27.90}{timelineplum}{-9.65}{ju2024robo}
    \MidEvent{27.00}{timelineplum}{-5.85}{kerr2024robot}
    \MidEvent{26.10}{timelineplum}{-10.95}{bahety2024screwmimic}
    \MidEvent{25.20}{timelineplum}{-4.55}{zhu2024vision}
    \MidEvent{24.30}{timelineplum}{-10.5}{li2024okami}
    \MidEvent{23.40}{timelineblue}{-7.15}{chen2025visa}
    \MidEvent{22.50}{timelineblue}{-9.65}{spiridonov2025generalist}
    \MidEvent{21.60}{timelineblue}{-5.85}{yang2025ar}
    \MidEvent{20.70}{timelineblue}{-10.95}{jiang2025rynnvla}
    \MidEvent{19.80}{timelineblue}{-4.55}{yoshida2025developing}
    \MidEvent{18.90}{timelineblue}{-12.25}{yang2025egovla}
    \MidEvent{18.00}{timelineblue}{-7.15}{luo2025being}
    \MidEvent{17.10}{timelineblue}{-9.65}{li2025scalable}
    \MidEvent{16.20}{timelineblue}{-5.85}{cai2025n}
    \MidEvent{15.30}{timelineblue}{-10.95}{feng2025spatial}
    \MidEvent{14.40}{timelineblue}{-4.55}{kareer2025egomimic}
    \MidEvent{13.50}{timelineblue}{-12.25}{qiu2025humanoid}
    \MidEvent{12.60}{timelineblue}{-7.15}{yuan2025motiontrans}
    \MidEvent{11.70}{timelineblue}{-9.65}{cheang2025gr}
    \MidEvent{10.80}{timelineblue}{-5.85}{wen2025gr}
    \MidEvent{9.90}{timelineblue}{-10.95}{kareer2025emergence}
    \MidEvent{9.00}{timelinebrick}{-4.55}{singh2025deep}
    \MidEvent{8.10}{timelinebrick}{-12.25}{dan2025x}
    \MidEvent{7.20}{timelinebrick}{-7.15}{lum2025crossing}
    \MidEvent{6.30}{timelinebrick}{-9.65}{jonnavittula2025view}
    \MidEvent{5.40}{timelinebrick}{-5.85}{chen2025vividex}
    \MidEvent{4.50}{timelinebrick}{-10.95}{guzey2025bridging}
    \MidEvent{3.60}{timelinebrick}{-4.55}{zhao2025dexh2r}
    \MidEvent{2.70}{timelinebrick}{-12.25}{yuan2025hermes}
    \MidEvent{1.80}{timelinebrick}{-7.15}{hsieh2025dexman}
    \MidEvent{0.90}{timelinebrick}{-9.65}{li2025maniptrans}

    \BotEvent{0.90}{timelinebrick}{-15.55}{zhao2025towards}
    \BotEvent{1.80}{timelinebrick}{-18.05}{ye2025video2policy}
    \BotEvent{2.70}{timelinegold}{-14.25}{yang2025tra}
    \BotEvent{3.60}{timelinegold}{-19.35}{chen2025graphmimic}
    \BotEvent{4.50}{timelinegold}{-12.95}{zhou2025human}
    \BotEvent{5.40}{timelinegold}{-20.65}{papagiannis2025r}
    \BotEvent{6.30}{timelinegold}{-15.55}{park2025demodiffusion}
    \BotEvent{7.20}{timelineplum}{-18.05}{shi2025zeromimic}
    \BotEvent{8.10}{timelineplum}{-14.25}{chen2025vidbot}
    \BotEvent{9.00}{timelineplum}{-19.35}{ma2025uni}
    \BotEvent{9.90}{timelineplum}{-12.95}{ren2025motion}
    \BotEvent{10.80}{timelineplum}{-20.65}{li2025novaflow}
    \BotEvent{11.70}{timelineplum}{-15.55}{yin2025object}
    \BotEvent{12.60}{timelineplum}{-18.05}{shan2025slot}
    \BotEvent{13.50}{timelineplum}{-14.25}{hsu2025spot}
    \BotEvent{14.40}{timelineplum}{-19.35}{tang2025functo}
    \BotEvent{15.30}{timelineplum}{-12.95}{tang2025mimicfunc}
    \BotEvent{16.20}{timelineplum}{-20.65}{werby2025articulated}
    \BotEvent{17.10}{timelineplum}{-15.55}{zhang2025actron3d}
    \BotEvent{18.00}{timelineplum}{-18.05}{haldar2025point}
    \BotEvent{18.90}{timelineplum}{-14.25}{liu2025egozero}
    \BotEvent{19.80}{timelineplum}{-19.35}{hu2025learning}
    \BotEvent{20.70}{timelineplum}{-12.95}{chen2025web2grasp}
    \BotEvent{21.60}{timelineplum}{-20.65}{zhou2025you}
    \BotEvent{22.50}{timelineplum}{-15.55}{heidinger20252handedafforder}
    \BotEvent{23.30}{timelineblue}{-18.05}{bi2026h}
    \BotEvent{24.05}{timelineblue}{-12.95}{zheng2026egoscale}
    \BotEvent{24.85}{timelineblue}{-15.55}{zhu2026emma}
    \BotEvent{25.65}{timelineblue}{-20.65}{luo2026being}
    \BotEvent{26.45}{timelineplum}{-14.25}{soraki2026objectforesight}
    \BotEvent{27.25}{timelineplum}{-19.35}{chen2026dexterous}
    \BotEvent{28.05}{timelineplum}{-12.95}{wang2026paws}
    \BotEvent{29.00}{timelineblue}{-18.05}{zhang2026unidex}

    \draw[timelineblue,line width=2.8pt,rounded corners=1pt] (0.20,-22.55) -- (1.10,-22.55);
    \node[anchor=west,font=\bfseries\fontsize{12.2}{11.4}\selectfont,text=timelineblue] at (1.30,-22.55) {Affordances for robot policy backbone training};

    \draw[timelinebrick,line width=2.8pt,rounded corners=1pt] (17.30,-22.55) -- (18.20,-22.55);
    \node[anchor=west,font=\bfseries\fontsize{12.2}{11.4}\selectfont,text=timelinebrick] at (18.40,-22.55) {Affordances for reward construction in robot policy};

    \draw[timelinegold,line width=2.8pt,rounded corners=1pt] (0.20,-23.25) -- (1.10,-23.25);
    \node[anchor=west,font=\bfseries\fontsize{12.2}{11.4}\selectfont,text=timelinegold] at (1.30,-23.25) {Affordances as robot policy condition};

    \draw[timelineplum,line width=2.8pt,rounded corners=1pt] (17.30,-23.25) -- (18.20,-23.25);
    \node[anchor=west,font=\bfseries\fontsize{12.2}{11.4}\selectfont,text=timelineplum] at (18.40,-23.25) {Affordances as robot policy};

  \end{tikzpicture}%
  }
  \caption{Chronological overview of methods under \textit{affordances as a bridge} in Sec.~\ref{sec:affordances}.}
  \label{fig:affordances_timeline}
\end{figure*}

Given the affordances from HOI analysis, as shown in Fig.~\ref{fig:affordances_transfer}, we organize the affordance-based bridging mechanism according to how tightly they are coupled with robot actions, ranging from backbone optimization, reward shaping, policy conditioning, to direct policy construction. The timeline of the collected works is provided in Fig.~\ref{fig:affordances_timeline}.

\begin{table*}[t]
\centering
\scriptsize
\setlength{\tabcolsep}{3.2pt}
\caption{Comparison of representative methods that formulate affordances for robot policy backbone training.}
\label{tab:affordance_backbone}
\begin{tabularx}{\linewidth}{p{3.3cm}p{3.3cm} X X p{3.3cm}}
\toprule
Reference & Hand affordance & Object affordance & Joint affordance & End-effector \\
\midrule
\rowcolor{gray!15}
\multicolumn{5}{l}{\textit{Affordance-based pretraining}} \\
\cite{shaw2023videodex} & 6D wrist pose, MANO & -- & -- & Dexterous hand \\
\cite{srirama2024hrp} & 2D hand position & 2D object bbox & Contact region & Parallel gripper \\
\cite{chen2025visa} & 2D hand mask, 2D action flow & 2D object mask, 2D action flow & -- & Parallel gripper \\
\cite{spiridonov2025generalist} & 3D hand keypoint flow & -- & -- & Parallel gripper \\
\cite{yang2025ar} & 3D hand keypoint & -- & -- & Parallel gripper \\
\cite{jiang2025rynnvla} & 2D hand keypoint & -- & -- & Parallel gripper \\
\cite{yoshida2025developing} & -- & 6D object pose & -- & Parallel gripper \\
\cite{yang2025egovla} & 6D wrist pose, MANO & -- & -- & Dexterous hand \\
\cite{luo2025being} & 6D wrist pose, MANO & -- & -- & Dexterous hand \\
\cite{li2025scalable} & 3D hand trajectory & -- & -- & Dexterous hand \\
\cite{cai2025n} & 6D wrist pose, 3D finger position & -- & -- & Dexterous hand \\
\cite{feng2025spatial} & 3D hand trajectory, MANO & 2D object bbox & -- & Dexterous hand, parallel gripper \\
\cite{bi2026h} & 6D wrist pose, 3D finger position & -- & -- & Parallel gripper \\
\cite{zheng2026egoscale} & 6D wrist pose, hand joint angle & -- & -- & Dexterous hand \\
\cite{zhang2026unidex} & 3D fingertip trajectory & -- & -- & Dexterous hand \\
\midrule
\rowcolor{gray!15}
\multicolumn{5}{l}{\textit{Affordance-based co-training}} \\
\cite{kareer2025egomimic} & 6D hand pose & -- & -- & Parallel gripper \\
\cite{qiu2025humanoid} & 6D wrist pose, 3D finger position & -- & -- & Dexterous hand \\
\cite{yuan2025motiontrans} & 6D wrist pose, 3D hand keypoint & -- & -- & Dexterous hand \\
\cite{cheang2025gr} & 3D hand trajectory & -- & -- & Parallel gripper \\
\cite{wen2025gr} & 6D wrist pose, hand joint angle, 3D fingertip position & -- & -- & Dexterous hand \\
\cite{kareer2025emergence} & 3D hand keypoint, 6D hand pose & -- & -- & Parallel gripper \\
\cite{luo2026being} & 6D wrist pose, MANO & -- & Interaction timing & Dexterous hand, parallel gripper \\
\cite{zhu2026emma} & 6D hand pose & -- & -- & Parallel gripper \\
\bottomrule
\end{tabularx}
\end{table*}

\textit{(b) Affordances for robot policy backbone training:} Affordances encompass rich and explicit hand-object interaction patterns. Therefore, they provide more action-grounded transfer signals between human videos and robot policies, which can be used to pretrain the backbone of robot policies after attaching an affordance-specific decoder. The high-level diagram of this bridging mechanism is illustrated in Fig.~\ref{fig:affordances_transfer}(a). For example, VideoDex~\citep{shaw2023videodex} explicitly extracts human wrist poses and hand pose parameters from Internet videos, and retargets them into robot wrist and finger trajectories. These reconstructed motions are used as an action prior for pretraining a Neural Dynamic Policy (NDP), which is then adapted with a small number of in-domain teleoperated demonstrations. To further enrich the supervision signals from affordances, HRP~\citep{srirama2024hrp} is required to predict future contact points, human hand poses, and the target object bounding boxes given a video frame as input. Thus, it is encouraged to focus on actionable scene regions, target objects, and interaction-relevant hand motion, leading to stronger downstream robotic performance across diverse tasks, robot morphologies, and camera views. Nevertheless, the NDP model of VideoDex~\citep{shaw2023videodex} and the ViT-B backbone of HRP~\citep{srirama2024hrp} are both relatively lightweight, with parameter scales below roughly one hundred million. This may limit the effectiveness of large-scale affordance-based pretraining for the backbone of robot policies. Therefore, more researchers scale affordance supervisions to optimize the VLA backbone with hundreds of millions to billions of parameters (e.g., PaliGemma~\citep{beyer2024paligemma}, InternVL3~\citep{chen2024expanding}, NVILA~\citep{liu2025nvila}). For instance, ViSA-Flow~\citep{chen2025visa} explicitly learns to predict future 2D hand-object interaction masks with a vision decoder. Its large-scale Transformer backbone then has the ability to capture semantic action flow to enhance robot skill learning with only a small dataset of robot demonstrations. In contrast, \cite{spiridonov2025generalist} directly pretrains the VLM of MotoVLA by predicting the 3D flow of human hand keypoints. Similarly, AR-VRM~\citep{yang2025ar} also adopts hand keypoint prediction to pretrain the VLA backbone. It further distinguishes itself by retrieving analogous human video exemplars as additional context for the mapping between hand motions and robot components. RynnVLA-001~\citep{jiang2025rynnvla} instead uses hand keypoint trajectories as auxiliary motion signals for future video prediction. It allows the backbone to anticipate incoming visual evolution along with physical dynamics for subsequent robot action planning. Considering the wide availability of off-the-shelf 6D hand\&object pose annotation tools, some pretraining schemes have switched to optimize the backbone with the task of hand\&object pose prediction. For example, \cite{yoshida2025developing} leverages the EgoScaler object-trajectory generation framework~\citep{yoshida2025generating} to construct annotated 6D object pose trajectories from egocentric videos. These pose labels are then used to pretrain the VLM backbone of a $\pi_0$ VLA model~\citep{black2024pi_0}. However, a larger body of existing works focuses on predicting hand poses and joint parameters for affordance-based pretraining. Compared with object-centric alternatives, capturing hand motion~\citep{pavlakos2024reconstructing,potamias2025wilor,zhang2025hawor} is generally more feasible on diverse Internet-scale human videos, because it does not depend on prior object reconstruction or point tracking. For example, \cite{yang2025egovla} pretrain EgoVLA by directly regressing future human wrist translations, wrist rotations, and hand joint parameters from egocentric videos in a unified action space. This hand-centric objective encourages the backbone to internalize temporal motion dynamics priors for action planning of the dexterous hands of humanoid robots. Instead of sharing the action head, Being-H0~\citep{luo2025being} introduces a dedicated part-level motion tokenizer that separately discretizes wrist and finger motions into hand motion tokens. This work also proposes UniHand-1.0, a large-scale mixed human video dataset with unified hand pose annotations. UniHand-1.0 comprises 130 million frames, substantially exceeding the approximately 500,000 frames used by EgoVLA~\citep{yang2025egovla}. As a concurrent work of Being-H0~\citep{luo2025being}, \cite{li2025scalable} argue that the training recipe with the original human annotations leads to temporal or granularity misalignment between text and actions. This weakens instruction following of the predicted actions. Thus, they convert human videos into robot-style VLA episodes with atomic hand-action segments, frame-aligned 3D hand trajectories, and newly generated imperative action descriptions. They also demonstrate that their proposed VITRA outperforms Being-H0~\citep{luo2025being}, even trained with a smaller human video dataset containing 26 million frames. To better organize the backbone pretraining data, \cite{cai2025n} distinguish human videos into in-the-wild data, which exceed 1,000 hours and are diverse and easy to collect, and on-task data, which exceed 20 hours and are aligned with the target robot tasks. VIPA-VLA~\citep{feng2025spatial} performs explicit visual-physical alignment on human videos by jointly leveraging 3D visual annotations and hand trajectory annotations, enabling the VLA backbone to acquire 2D-3D spatial grounding before robot post-training. A representative extension of hand-centric affordance pretraining to bimanual manipulation is H-RDT~\citep{bi2026h}. It pretrains a diffusion Transformer backbone with 3D bimanual hand pose annotations in a unified 48-dimensional action space. It then transfers these human manipulation priors to different robot embodiments through modular action encoders and decoders during cross-embodiment fine-tuning. JoyAI-RA~\citep{zhang2026joyai} further scales pretraining by integrating web data, large-scale egocentric human videos, simulation-generated trajectories, and real-robot demonstrations in a multi-source multi-level fashion. It recovers hand trajectories from human videos and retargets them to multiple robot embodiments through action-space unification. EgoScale~\citep{zheng2026egoscale} interestingly shows that affordance-based backbone pretraining on egocentric human videos follows a clear log-linear scaling law, where increasing human data scale consistently reduces the validation loss of wrist motion and retargeted
dexterous hand action prediction. More importantly, this validation loss strongly correlates with downstream robot performance, suggesting that scaling affordance-based backbone pretraining provides a predictable route for improving dexterous manipulation. Recently, UniDex~\citep{zhang2026unidex} addresses the embodiment discrepancy for backbone pretraining from both motion and perception aspects. Specifically, it relies on an interactive retargeting process to map human fingertip motions to robot hands while maintaining physically reasonable hand-object contacts. It removes human hands from reconstructed 3D point clouds to further alleviate the kinematic and visual mismatch. The resulting data is used for large-scale backbone pretraining to establish a dexterous hand foundation model that supports cross-hand transfer. Overall, affordance-based pretraining mainly serves to endow the backbone with transferable manipulation priors rather than a directly executable robot policy. In practice, these priors still need robot-specific post-training or fine-tuning to bridge the residual embodiment and action-space gaps.

In contrast to using affordances to pretrain the backbone of robot policies, some researchers turn to a unified co-training scheme. As shown in Fig.~\ref{fig:affordances_transfer}(b), this scheme alleviates the loss of human manipulation priors during subsequent robot fine-tuning, and potential composite failures in the aforementioned two-stage training pipeline. Besides, unlike affordance-based pretraining, co-training often yields policies that are directly usable for downstream robotic manipulation, since its training process already incorporates robot-specific demonstrations. For example, EgoMimic~\citep{kareer2025egomimic} co-trains imitation policies on egocentric human videos with 3D hand tracking and teleoperated robot data, treating both as equally important embodied demonstrations. It witnesses a significant performance jump of the co-training scheme from scaling with human data. Drawing inspiration from ALOHA~\citep{zhao2023learning}, this work designs a novel robot manipulator to mimic human arm motions, and equips the robot with Aria glasses to better match the egocentric observations of human demonstrations. These alignment techniques help narrow the embodiment gap inherent in co-training data. EMMA~\citep{zhu2026emma} further extends EgoMimic~\citep{kareer2025egomimic} to mobile manipulation by co-training with egocentric human full-body motion data with static robot manipulation data. It bridges the navigation kinematic gap by projecting the human head pose onto the ground plane and optimizing velocity commands with human waypoints. However, \cite{qiu2025humanoid} discover that the strict visual sensor alignment and heuristic designs like visual masking in EgoMimic~\citep{kareer2025egomimic} lead to composite failures. Therefore, they instead train a Human Action Transformer in a unified human-robot state-action space. With differentiable retargeting and simple image augmentations, this design avoids strict sensor matching, heuristic masking, and specialized hardware redesign when deploying on a humanoid robot with dexterous hands. Given the potential imbalance between human and robot data in the co-training schemes of \cite{qiu2025humanoid}, \cite{yuan2025motiontrans} exploit a weighted strategy inspired by \cite{wei2025empirical}. It adopts dataset sizes to determine domain weights such that the total contributions of human and robot data are balanced during co-training. This work also validates the effectiveness of multi-task human-robot co-training on the VLA model~\citep{black2024pi_0} for zero-shot manipulation. \cite{cheang2025gr} also incorporate human trajectory data and co-train the VLA model with vision-language data and robot trajectories. Compared to GR-1~\citep{wu2023unleashing} and GR-2~\citep{cheang2024gr}, GR-3 significantly improves few-shot generalization to unseen objects by introducing human hand trajectory supervisions during fine-tuning. GR-Dexter~\citep{wen2025gr} follows the backbone of GR-3, but further substantially enriches the action space from binary discrete gripper actions to the combination of arm joint actions, arm end-effector poses, hand joint actions, and fingertip positions. In contrast to GR-Dexter which uses a composite action vector, Being-H0.5~\citep{luo2026being} proposes a semantically aligned unified action space to co-train human and heterogeneous robot embodiments under a shared physical vocabulary. This work also constructs the UniHand-2.0 dataset, which is composed of human, robot, and visual-language data, and is approximately 200 times larger than the UniHand-1.0 dataset proposed in its prior work~\citep{luo2025being}. Interestingly, \cite{kareer2025emergence} have recently discovered that human-robot transfer is an emergent property of diverse VLA pretraining. A simple human-robot co-training recipe begins to work only after sufficiently diverse robot pretraining across scenes, tasks, and embodiments. Their analysis further suggests that this emergence is driven by increasingly embodiment-agnostic latent representations, which allow human and robot observations to align in feature space without any explicit transfer mechanism. These key findings would further support unified human-robot co-training as a scalable route for leveraging human videos in VLA models, especially when sufficient diversity has already been included in robot pretraining data. 

As can be noted, both affordance-based pretraining and unified human-robot co-training exhibit a clear scaling trend, evolving from lightweight task-specific models toward large VLA backbones trained on increasingly diverse human-video mixtures. This progression is largely enabled by advances in HOI analysis techniques, the rapid growth of human video datasets with richer annotations, and the emergence of large foundation models that make large-scale cross-embodiment learning increasingly practical.

\begin{table*}[t]
\centering
\scriptsize
\setlength{\tabcolsep}{3.2pt}
\caption{Comparison of representative methods that formulate affordances for reward construction in robot policy learning.}
\label{tab:affordance_reward}
\begin{tabularx}{\linewidth}{p{3.3cm}p{3.3cm}p{5.5cm} X X}
\toprule
Reference & Hand affordance & Object affordance & Joint affordance & End-effector \\
\midrule
\rowcolor{gray!15}
\multicolumn{5}{l}{\textit{Single-modality reward construction}} \\
\cite{das2021model} & -- & 2D object keypoint & -- & Parallel gripper \\
\cite{singh2025deep} & 3D fingertip position & -- & -- & Dexterous hand \\
\cite{dan2025x} & -- & 6D object pose & -- & Parallel gripper \\
\cite{ye2025video2policy} & -- & 6D object pose, 2D object mask & -- & Parallel gripper \\
\midrule
\rowcolor{gray!15}
\multicolumn{5}{l}{\textit{Multimodal reward construction}} \\
\cite{sieb2020graph} & 3D finger position & 6D object pose, 3D object keypoint, 3D object bbox & Grasp state & Parallel gripper \\
\cite{patel2022learning} & 6D palm pose, hand joint angle & 6D object pose & -- & Parallel gripper \\
\cite{mandikal2022dexvip} & 6D hand pose & 3D object keypoint & Contact point & Dexterous hand \\
\cite{kumar2023graph} & 2D hand bbox & 2D object bbox, inter-object graph & -- & Parallel gripper \\
\cite{lum2025crossing} & MANO & 6D object pose & -- & Dexterous hand \\
\cite{jonnavittula2025view} & 3D wrist position & 3D object position & Contact state & Parallel gripper \\
\cite{chen2025vividex} & 3D hand joint position & 6D object pose & -- & Dexterous hand \\
\cite{guzey2025bridging} & 2D fingertip position & 2D object keypoint & -- & Dexterous hand \\
\cite{zhao2025dexh2r} & 6D hand pose, MANO & 6D object pose & -- & Dexterous hand \\
\cite{yuan2025hermes} & 3D hand keypoint, 6D palm pose & 6D object pose & Contact prior & Dexterous hand \\
\cite{hsieh2025dexman} & 6D wrist pose, 3D fingertip position, MANO & 6D object pose & Contact prior & Dexterous hand \\
\cite{li2025maniptrans} & 6D wrist pose, MANO & 6D object pose & Contact prior & Dexterous hand \\
\cite{zhao2025towards} & MANO & 3D object keypoint &  Contact region & Dexterous hand \\
\bottomrule
\end{tabularx}
\end{table*}

\textit{(c) Affordances for reward construction in robot policy learning:}
To more tightly incorporate affordances into robot policy learning, some works use affordances to construct a reward function that directly guides robot policy optimization. By evaluating how well robot actions reproduce the affordance patterns observed in human videos, these reward signals help align robot actions with human interaction dynamics. Compared with the aforementioned affordance-based backbone training, this paradigm more directly affects robot policy optimization. It enables robots to refine their behaviors through trial-and-error while remaining human interaction grounding. Fig.~\ref{fig:affordances_transfer}(c) presents constructing rewards using a single modality of interaction. For example, \cite{das2021model} learn a cost function from human videos with a pre-trained self-supervised keypoint detector, and then use it to optimize robot behavior through visual model predictive control. Instead of keypoint prediction, \cite{singh2025deep} train a predictive model of future human fingertip motion from human interaction videos and use its zero-shot predictions on robot trajectories to define a tracking reward. This learned motion-tracking reward is then combined with sparse task rewards to optimize robot sensorimotor policies through RL. Instead of using hand motion as references, \cite{dan2025x} use object motion extracted from human videos to define an object-centric reward in a reconstructed photorealistic simulator. Concretely, the reward is constructed from the discrepancy between the simulated robot-induced object trajectory and the target object trajectory extracted from the video. Thus, the policy optimization is driven to match the object state changes in an embodiment-agnostic manner. More recently, Video2Policy~\citep{ye2025video2policy} scales this direction by training RL policies with in-context object-centric reward functions automatically generated by LLM. Rather than generating rewards from a single demonstrated video, it uses large numbers of everyday human videos to synthesize diverse rewardable simulation tasks, supporting scalable Real2Sim2Real policy learning.

As illustrated in Fig.~\ref{fig:affordances_transfer}(d), more related works combine the motion priors of hands and target objects in human videos to construct reward functions, rather than only using a single motion modality. An early transition toward richer hand-object reward design is from \cite{patel2022learning}. This work jointly optimizes 4D trajectories of both hands and objects from Internet videos, while still training robot policies mainly with object pose imitation in simulation. It can be regarded as a bridge from purely single-modality rewards to later methods that combine hand priors and object-based reward signals. As exemplified by \cite{mandikal2022dexvip}, they construct an object-affordance contact reward that encourages the robot hand to approach human-contacted grasp regions on the object. Simultaneously, they also introduce a pose reward that encourages the robot joints to match the retargeted 6D human grasp pose. To decrease the number of exploration rollouts in prior RL paradigms, \cite{lum2025crossing} harness a pre-manipulation MANO hand pose as the exploration prior. Then, they construct an embodiment-agnostic reward by comparing the robot-induced object motion against the demonstrated object trajectory from the human video. In contrast, VIEW~\citep{jonnavittula2025view} and ViViDex~\citep{chen2025vividex} extract finer-grained hand (fingertip) waypoints from human videos as the prior to further improve exploration efficiency. Afterwards, the human-guided RL paradigm of using hand trajectories as priors while using object trajectories as reward signals of residual learning has become a popular design pattern in this direction~\citep{guzey2025bridging,zhao2025dexh2r,yuan2025hermes,hsieh2025dexman}. \cite{zhao2025dexh2r} further introduce test-time guidance with desired human hand and object trajectories, allowing the robot to adapt the learned policy to novel manipulation scenarios with stronger generalization. HERMES~\citep{yuan2025hermes} extends this RL paradigm from tabletop dexterous manipulation to longer-range mobile manipulation. It combines object-centric distance chain and object-tracking rewards, together with a power penalty, to capture the dynamic spatial relationships during hand-object contact and enhance execution smoothness. A navigation foundation model is further integrated with closed-loop PnP localization for mobile manipulation capability. \cite{hsieh2025dexman} also realize the importance of hand-object contact guidance like HERMES~\citep{yuan2025hermes}, and propose a contact-prior attraction reward to refine retargeted motions. ManipTrans~\citep{li2025maniptrans} further introduces a contact-force reward on top of hand motion imitation and object-tracking rewards. Specifically, when the demonstrated hand is sufficiently close to the object, it rewards nonzero fingertip contact forces in simulation, thus encouraging more stable grasping and improving residual learning for dexterous bimanual manipulation. Toward more semantically grounded contact supervision, \cite{zhao2025towards} introduce additional negative affordance guidance to penalize functionally inappropriate contact regions during residual policy refinement. 

A more structured reward variant is proposed by~\cite{sieb2020graph}. They generate rewards by matching the relative spatial configurations of hand/gripper-object 3D locations between the robot scene and the human demonstration. To address the limitation of same-domain visual trajectory following in this prior work~\citep{sieb2020graph}, \cite{kumar2023graph} further targets cross-domain and cross-embodiment imitation by learning a general alignment function from diverse videos. Instead of relying on manually designed correspondence rules, it represents demonstrations as graphs and learns hand-object interactions directly in graph space, thereby yielding reward functions that transfer more robustly across domains and embodiments.

Unlike affordance-based backbone training, affordance-based reward construction incorporates affordances into robot learning by shaping the optimization objective itself. Related works have evolved from single-modality motion rewards to richer hand-object reward designs that jointly encode grasp priors, object motion, contact geometry, and contact force. This design provides stronger flexibility in LfHV settings since robots can refine their behaviors through trial-and-error under human-grounded objectives.

\begin{table*}[t]
\centering
\scriptsize
\setlength{\tabcolsep}{3.2pt}
\caption{Comparison of representative methods that formulate affordances as robot policy condition.}
\label{tab:affordance_condition}
\begin{tabularx}{\linewidth}{p{2.7cm}X X X p{1.7cm}}
\toprule
Reference & Hand affordance & Object affordance & Joint affordance & End-effector \\
\midrule
\rowcolor{gray!15}
\multicolumn{5}{l}{\textit{Predicted affordances as policy conditions}} \\
\cite{lee2017learning} & 2D hand position & 2D object position & -- & Parallel gripper \\
\cite{wang2023mimicplay} & 3D hand position & -- & -- & Parallel gripper \\
\cite{bharadhwaj2023towards} & 2D hand mask & 2D object mask & -- & Parallel gripper \\
\cite{wen2023any} & -- & -- & Point track & Parallel gripper \\
\cite{bharadhwaj2024track2act} & -- & -- & Point track & Parallel gripper \\
\cite{yang2025tra} & -- & -- & Point track & Parallel gripper \\
\cite{xu2024flow} & -- & 2D object bbox, 2D object keypoint, motion flow & -- & Parallel gripper \\
\cite{chen2025graphmimic} & 2D fingertip position & 2D object mask, 2D object keypoint & Interaction graph & Parallel gripper \\
\cite{zhou2025human} & 3D hand keypoint & -- & -- & Parallel gripper \\
\midrule
\rowcolor{gray!15}
\multicolumn{5}{l}{\textit{Prompt-extracted affordances as policy conditions}} \\
\cite{wang2023robot} & 3D hand trajectory & Object features, insertion geometry & -- & Parallel gripper \\
\cite{gu2023rt} & 2D hand trajectory sketch & -- &  2D interaction marker & Parallel gripper \\
\cite{papagiannis2025r} & 3D hand keypoints, MANO & -- & 3D scene keypoints & Parallel gripper \\
\cite{park2025demodiffusion} & 6D hand pose & -- & -- & Parallel gripper \\
\bottomrule
\end{tabularx}
\end{table*}

 \textit{(d) Affordances as robot policy condition:} 
In addition to serving as supervision for backbone training and reward signals for policy optimization, affordances can also be injected into robot policies more directly as explicit conditions. Most existing works in this direction focus on predicting \textit{future affordances} as conditions, since they can be used as the action-guiding signals (goals) to establish a tighter connection between human video understanding and robot execution. As shown in Fig.~\ref{fig:affordances_transfer}(e), this bridging mechanism also requires pretraining an affordance prediction model like affordance-based robot policy backbone pretraining. However, it further integrates the predicted affordances into downstream robot policies explicitly, rather than using the backbone. \cite{lee2017learning} present an early example of using affordances as explicit policy conditions by forecasting future hand locations from egocentric human videos. These predicted future hand-object interaction states are then fed into a manipulation network as intermediate action-guiding signals. This scheme enables the robot to generate motor commands conditioned on anticipated future interactions rather than only current observations. Instead of explicitly predicting hand locations, MimicPlay~\citep{wang2023mimicplay} conditions robot control on predicted future latent subgoals from human videos. Future 3D human hand trajectories supervise the latent prediction model after attaching a GMM decoder. To further relax the assumption of in-domain videos in MimicPlay~\citep{wang2023mimicplay}, \cite{bharadhwaj2023towards} train a diffusion model with numerous in-the-wild human video clips, to hallucinate future hand and object masks that condition a separate robot manipulation policy. After that, more related works with predicted affordances as conditions attend to forecasting denser point tracks (motion flow) for finer robot control guidance. For example, ATM~\citep{wen2023any} pretrains a language-conditioned track Transformer to predict future trajectories of arbitrary points in the image. It then uses the resulting dense point tracks as explicit conditions for policy learning. These point-based motion cues provide detailed control guidance and preserve object permanence. However, the ATM framework is not readily applicable to web videos, because its policy depends on per-step image observations to predict point tracks. Besides, ATM requires costly real-world robot data to train alongside the in-domain human demonstration data. To address these limitations, as a concurrent work of ATM, Track2Act~\citep{bharadhwaj2024track2act} learns to predict point tracks from only an initial image and a goal only with Internet human videos. It leaves an interface for open-loop execution by converting the predicted tracks into rigid object transforms. The predicted tracks can also serve as conditions for learning a lightweight embodiment-specific residual policy to compensate for execution errors and embodiment mismatch. Tra-MoE~\citep{yang2025tra} also aims to improve ATM by introducing a sparse MoE trajectory predictor for broader multi-domain data when learning trajectory-conditioned policies. It also integrates an adaptive policy conditioning technique, which constructs a mask-modality input with image-aligned learnable embeddings. Action-labeled robot demonstrations are used for subsequent behavior cloning. As can be seen, Track2Act~\citep{bharadhwaj2024track2act} and Tra-MoE~\citep{yang2025tra} both require collecting real-world robot data for learning closed-loop robot policies. In contrast, \cite{xu2024flow} factorize the pipeline into a flow generation network trained on real human videos and a flow-conditioned closed-loop policy trained entirely on simulated robot play data. By conditioning robot actions on object-centric flow rather than embodiment-specific trajectories, this scheme bypasses the need for collecting real-world robot training data. It can still be directly deployed in the real world with a minimal sim-to-real gap. Nonetheless, \cite{chen2025graphmimic} argue that the above flow-based methods directly model representations from pixel space, neglecting internal structures of objects, spatial object-object and object-effector relationships. Therefore, they propose GraphMimic, which abstracts each human video frame into a structured graph with object vertices and visual action vertices. Then, they pretrain a graph-to-graphs generative model to predict future graphs as policy conditions. By modeling object structure and spatial relations in graph space, it provides more structured action guidance than pixel-space flow. By directly predicting the 3D trajectory of the midpoint of the thumb and index finger keypoints with a trajectory expert, Traj2Action~\citep{zhou2025human} produces a high-level motion plan to condition a robot action expert. 

Compared to existing works using predicted affordances as conditions, a few works receive videos as additional prompts, which can directly extract affordances as conditions without an additional prediction stage, as shown in Fig.~\ref{fig:affordances_transfer}(f). For example, \cite{wang2023robot} extract hand trajectories from a single human demonstration video to generate imitated robot approach motions, while simultaneously localizing object features for subsequent visual servoing. RT-Trajectory~\citep{gu2023rt} instead uses human videos to construct coarse hindsight trajectory sketches, which visually encode end-effector motion and interaction markers as explicit task specifications for a robot policy. Representing human demonstrations as rough motion sketches rather than precise geometric trajectories, it provides action-guiding conditions while remaining flexible enough to generalize across novel tasks. R+X~\citep{papagiannis2025r} does not optimize robot policies with explicit affordance prompts. Instead, it retrieves relevant clips from long, unlabelled first-person human videos, extracts their visual keypoints and hand joints, and finally uses them to condition an in-context imitation learning method~\citep{di2024keypoint}. More recently, DemoDiffusion~\citep{park2025demodiffusion} leverages 6D hand poses as an effective initialization for retargeted open-loop robot motions, and then refines this trajectory with a pre-trained generalist diffusion policy to make it both task-consistent and robot-feasible. Thus, it mitigates both the embodiment gap and the lack of closed-loop feedback. 

Compared with affordance-based backbone training and reward shaping, using affordances as policy conditions establishes a tighter connection between human videos and robot execution, by directly injecting interaction-aware geometric cues into downstream control. Existing methods have evolved from predicting sparse future hand locations and latent subgoals to generating denser motion conditions such as masks, point tracks, flow, and structured graphs. A smaller body of work directly extracts affordances from human videos as prompts for policy learning or execution. Although this paradigm provides more explicit action guidance than backbone training and reward construction, it still typically relies on a separate affordance prediction or extraction stage. In many cases, additional robot demonstrations or policy adaptation are required to ground the conditions into embodiment-specific actions.

\begin{table*}[t]
\centering
\scriptsize
\setlength{\tabcolsep}{3.2pt}
\caption{Comparison of representative methods that formulate affordances as robot policy.}
\label{tab:affordance_policy}
\begin{tabularx}{\linewidth}{p{2.9cm}X X p{4.5cm} p{1.7cm}}
\toprule
Reference & Hand affordance & Object affordance & Joint affordance & End-effector \\
\midrule
\rowcolor{gray!15}
\multicolumn{5}{l}{\textit{Hand-centric policy grounding}} \\
\cite{sivakumar2022robotic} & 6D hand pose, MANO & -- & -- & Dexterous hand \\
\cite{arunachalam2022dexterous} & 2.5D hand keypoint & -- & -- & Dexterous hand \\
\cite{qin2022one} & 2D hand bbox, 6D wrist pose, SMPL-X & -- & -- & Dexterous hand \\
\cite{bahl2022human} & 2D hand bbox, 3D wrist rotation & -- & Interaction timing & Parallel gripper \\
\cite{bharadhwaj2023zero} & 6D palm pose & -- & -- & Parallel gripper \\
\cite{bahl2023affordances} & 2D hand trajectory & -- & 2D contact point & Parallel gripper \\
\cite{kannan2023deft} & 6D wrist pose, hand joint angles, MANO & -- & 3D contact point & Dexterous hand \\
\cite{kuang2024ram} & 3D hand trajectory, post-contact direction vector & -- & 3D contact point & Parallel gripper \\
\cite{shi2025zeromimic} & 6D wrist trajectory & -- & 2D contact point & Parallel gripper \\
\cite{ma2025uni} & 3D hand keypoints & -- & Interaction timing & Parallel gripper \\
\midrule
\rowcolor{gray!15}
\multicolumn{5}{l}{\textit{Object-centric policy grounding}} \\
\cite{wen2022you} & -- & 6D object pose & -- & Parallel gripper \\
\cite{ko2023learning} & -- & 2D object mask, 3D object keypoint & Optical flow & Parallel gripper \\
\cite{yuan2024general} & -- & 3D general flow & -- & Parallel gripper \\
\cite{heppert2024ditto} & -- & 6D object pose & -- & Parallel gripper \\
\cite{ju2024robo} & -- & 3D function point & -- & Parallel gripper \\
\cite{kerr2024robot} & -- & 4D differentiable part model & -- & Parallel gripper \\
\cite{li2025novaflow} & -- & 2D object mask, 3D object keypoint & -- & Parallel gripper \\
\cite{yin2025object} & -- & 2D object mask, 3D motion field & -- & Parallel gripper \\
\cite{shan2025slot} & -- & 2D object mask & -- & Parallel gripper \\
\cite{hsu2025spot} & -- & 6D object pose & -- & Parallel gripper \\
\cite{tang2025functo} & -- & 2D objece mask, 3D function point & 3D grasp point & Parallel gripper \\
\cite{zhang2025actron3d} & -- & 3D point flow & 2D contact mask & Parallel gripper \\
\midrule
\rowcolor{gray!15}
\multicolumn{5}{l}{\textit{Interaction-centric policy grounding}} \\
\cite{qin2022dexmv} & 6D hand pose, MANO & 6D object pose & -- & Dexterous hand \\
\cite{gao2023k} & 3D hand keypoint & 3D object keypoint & Geometric keypoint constraint & Dexterous hand \\
\cite{ye2023learning} & 3D hand joint position & Object point cloud & -- & Dexterous hand \\
\cite{bahety2024screwmimic} & 6D wrist pose & Object point cloud & 3D grasp/placement point & Parallel gripper \\
\cite{li2024okami} & SMPL-H trajectory & 2D object keypoint, 2D object keypoint, object point cloud & -- & Dexterous hand \\
\cite{haldar2025point} & 3D hand keypoint & 3D object keypoint & Grasp timing & Parallel gripper \\
\cite{liu2025egozero} & 6D hand pose, 3D hand keypoint & 3D object keypoint & Grasp timing & Parallel gripper \\
\cite{chen2025web2grasp} & MANO & Object point cloud & Point-to-point distance & Dexterous hand \\
\cite{zhou2025you} & 6D hand pose & Object point cloud & Bimanual coordination order & Parallel gripper \\
\cite{heidinger20252handedafforder} & 2D hand mask & 2D object mask & Interaction region & Parallel gripper \\
\cite{chen2026dexterous} & 6D hand pose & 6D object pose, object mesh & Contact map & Dexterous hand \\
\cite{wang2026paws} & 3D hand trajectory, MANO & Articulated object structure & Motion type, motion axis, and motion origin & Parallel gripper \\
\bottomrule
\end{tabularx}
\end{table*}

\textit{(e) Affordances as robot policy:} As illustrated in Fig.~\ref{fig:affordances_transfer}(g), affordances from human videos can also be used to directly construct robot policies themselves. Compared to the above affordance-based bridging mechanisms, affordances here are treated as executable action representations. They can be directly transformed into robot control commands. Rather than learning a separate policy conditioned on affordances, these methods interpret affordances as the policy output space, where predicted hand-object interaction cues (e.g., hand trajectories, contact points, or object transformations) are mapped to robot motions through retargeting, optimization, or embodiment-specific controllers. Therefore, affordances function not only as action guidance but also as the primary representation of robot behavior, enabling a more direct action-oriented transfer from human videos to robot execution. Early works emphasize hand-centric affordances, where human hand motion dominates policy grounding despite accessible object affordance.
For example, Robotic Telekinesis~\citep{sivakumar2022robotic} learns to retarget monocular human hand-arm motion into smooth and safe robot hand-arm trajectories, allowing the demonstrated hand motion itself to directly define the executed robot behavior. \cite{arunachalam2022dexterous} further extend this hand-centric transfer paradigm into an efficient learning framework. The fingertip motions estimated from a single RGB camera by MediaPipe~\citep{zhang2020mediapipe} are converted into robot demonstrations for training dexterous manipulation policies through IL and RL. As a concurrent work of \cite{sivakumar2022robotic} and \cite{arunachalam2022dexterous}, \cite{qin2022one} develop a customized robot hand that better matches the operator's kinematic structure. The customized hand functions as a bridge to stably transfer the collected hand trajectories to multiple actual specified robot hands. This research provides the codebase of the following work~\citep{singh2025hand}, which further introduces Internet human data as priors. \cite{bahl2022human} first extract hand positions of the timestep where the interaction starts and ends in the human video, along with a midpoint. They are used to initialize a robot policy, which is then improved iteratively via an online task-agnostic exploration policy. However, these prior works are limited by the requirements of a video prompt to mimic or per-task online fine-tuning in the real-world. To address these limitations, \cite{bharadhwaj2023zero} learn a scene-conditioned hand trajectory prediction model, and directly deploy it for robot manipulation by mapping predicted trajectories to the robot actions in a zero-shot fashion. Motivated by the similar limitation of directly mimicking a demonstration, \cite{bahl2023affordances} train an affordance model that can predict a contact heatmap together with post-contact wrist waypoints from human-agnostic video frames. This hand-centric VRB formulation explicitly captures the target interaction location and the corresponding manipulation pattern in an agent-agnostic form. Thus, the learned affordances are more versatile for multiple downstream robot learning paradigms. The affordance representation in this work has inspired many subsequent works~\citep{kannan2023deft,kuang2024ram,shi2025zeromimic,chen2025vidbot,ma2025uni}. For instance, DEFT~\citep{kannan2023deft} further extends the VRB representation by how to grasp through wrist rotations and hand joint angles. In contrast to DEFT which predicts hand MANO parameters~\citep{romero2022embodied}, ZeroMimic~\citep{shi2025zeromimic} instead uses AnyGrasp~\citep{fang2023anygrasp} to determine the grasp actions on the point clouds selected by VRB. Following these works, more works like RAM~\citep{kuang2024ram} and VidBot~\citep{chen2025vidbot} consistently extend the VRB representation from 2D space to 3D space. RAM~\citep{kuang2024ram} hierarchically retrieves similar demonstrations from a large multi-source affordance memory. It then lifts the retrieved 2D affordance into in-domain 3D contact points and 3D post-contact directions with a sampling-based strategy, which also uses AnyGrasp~\citep{fang2023anygrasp} to generate grasp proposals. This retrieval-and-transfer design enables more generalizable zero-shot manipulation. Following RAM~\citep{kuang2024ram}, VidBot~\citep{chen2025vidbot} also lifts pixel-level trajectories from VRB to 3D using normal clusters as cues. It further adopts a coarse-to-fine affordance model that first predicts coarse action types and then generates fine-grained 3D interaction trajectories with a diffusion model. This design improves zero-shot manipulation under novel scenes and embodiments. Instead of following the VRB-style affordance representations, Motion Tracks~\citep{ren2025motion} define affordances as short-horizon 2D motion tracks that capture the predicted direction of motion for either human hands or robot end-effectors in image space. The predicted tracks are lifted to executable 6D robot trajectories via multi-view synthesis. This work introduces a learnable retargeting network to map robot keypoints into a human hand structure in image space, thus narrowing the embodiment gap. The grasp timestamps are determined from the proximity between fingertip keypoints and the segmented object mask. More recently, \cite{ma2025uni} further enhance hand-centric affordance construction by explicitly decoupling camera egomotion from hand motion and forecasting future hand trajectories. They replace the spatial contact affordances in prior works with the temporal contact and separation events, which can be automatically localized by EgoLoc~\citep{zhang2025zero,ma2025egoloc}.

The above works predominantly adopt hand-centric affordances, where human hand motion serves as the primary anchor for policy construction. In contrast, some works have shifted toward object-centric affordances, regarding object transformations and functional object states as the primary executable representation of robot policies. We have introduced Track2Act~\citep{bharadhwaj2024track2act} in \textit{affordances as robot policy condition}. By contrast, its open-loop variant falls under \textit{affordances as robot policy}. This is because the predicted object-centric point tracks are directly converted into rigid object transforms and then retargeted into executable robot trajectories. To address the limitations of 2D flow and in-domain finetuning inherent in Track2Act, \cite{yuan2024general} directly train a closed-loop 3D flow prediction model. They boost the zero-shot generalization capability of the resulting closed-loop policy by Hand Mask Augmentation and Query Points Sampling techniques. In contrast to \cite{bharadhwaj2024track2act} and \cite{yuan2024general} explicitly predicting motion flow, \cite{ko2023learning} instead implement video prediction and infer closed-form actions through estimated optical flow between frames. NovaFlow~\citep{li2025novaflow} similarly generates a task video from an initial image and a language instruction. It then extracts actionable 3D object flow via depth estimation and 3D tracking to map it directly to robot actions. Its main extension over \cite{ko2023learning} is support for articulated and deformable object manipulation. Considering the above point-cloud 3D flow is noisy and cannot represent motion accurately, \cite{yin2025object} learns a diffusion model in simulation to predict dense 3D object motion field. This motion field preserves finer object motion details than point-based flow and can be directly translated into executable zero-shot robot actions. Their experiments demonstrate the superiority of depth robustness and distractor generalization against prior works~\citep{bharadhwaj2024track2act,yuan2024general}. In contrast to the above object-flow-based methods, DITTO~\citep{heppert2024ditto} follows an object-centric trajectory transformation pipeline from a single RGB-D human demonstration. It extracts the relative object pose changes offline, re-detects objects in the current scene, and then warps the demonstration trajectory for direct robot execution. A closely related work~\citep{wen2022you} also reprojects the demonstrated object trajectory for direct execution. Compared with DITTO~\citep{heppert2024ditto}, it learns a category-level canonical object representation in simulation, enabling the demonstrated trajectory to be adapted across different object instances. For precise placement tasks, \cite{shan2025slot} further transfers object-centric placement relations from a human video. This work identifies the manipulated object and the target placement slot, then estimates slot-level 6D placement transforms in the robot scene for direct execution. Benefiting from the emergence of FoundationPose~\citep{wen2024foundationpose}, SPOT~\citep{hsu2025spot} and ObjectForesight~\citep{soraki2026objectforesight} can directly train diffusion models with accurate pose annotations, to predict 6D object pose trajectories as robot policies. More recently, ActiveGlasses~\citep{zou2026activeglasses} further extends object-centric affordances to active perception. It collects egocentric demonstrations with smart glasses and extracts object trajectories along with human head motion. An object-centric 3D point-cloud policy is trained to predict manipulation trajectories and active camera movements jointly. Compared to using object flow and poses, Robo-ABC~\citep{ju2024robo} represents object-centric affordances as functional contact points. By retrieving similar objects and transferring their contact points to unseen objects via semantic correspondence, it enables zero-shot cross-category grasping without additional training, which inspires the following work~\citep{kuang2024ram}. Functo~\citep{tang2025functo} and MimicFunc~\citep{tang2025mimicfunc} further extend object-centric policy construction with VLM functional reasoning inspired by ReKep~\citep{huang2024rekep}. They abstract complex tools into a function-centric spatial skeleton defined by three keypoints, namely the grasp point, function point, and center point. Then, they use semantic correspondence to transfer these functional anchors across novel heterogeneous tools for zero-shot manipulation. For articulated objects, object affordances must further capture kinematic structure and part-state evolution, rather than only static contact anchors. Ditto~\citep{jiang2022ditto} provides an early foundation in this direction by reconstructing part-level geometry together with the articulation model from observations before and after interaction. \citep{kerr2024robot} instead recovers 4D part motion from a single monocular human demonstration and a static object scan, then plans robot motions to reproduce the demonstrated part trajectories. Pushing articulated affordance extraction toward less controlled settings, \cite{werby2025articulated} estimate articulated part trajectories and joint axes directly from egocentric human videos under camera motion and partial observability. They make such object-centric representations more accessible in the wild. More recently, Actron3D~\citep{zhang2025actron3d} distills geometry, appearance, and affordance cues from a few uncalibrated monocular videos into a compact Neural Affordance Function, where transferable 6D manipulation policies can be retrieved and optimized. 

Some other works attend to interaction-centric affordances, where hand and object cues jointly dominate robot policy generation. An early example is DexMV~\citep{qin2022dexmv}, which extracts 6D hand and object poses from human videos and retargets their joint trajectories into robot demonstrations for imitation learning. In contrast, K-VIL~\citep{gao2023k} represents interaction-centric affordances as a sparse set of hand-object keypoint constraints extracted from human videos. In this work, the affordances reflect the structured geometric relation induced by hand-object interaction, which is then reproduced on the robot through a keypoint-based admittance controller. Subsequently, more works focus on extracting hand-object keypoint motion as interaction-centric affordances~\citep{haldar2025point,liu2025egozero,hu2025learning}. For example, Point Policy~\citep{haldar2025point} jointly encodes translated human hand keypoints and semantically meaningful object keypoints. A Transformer-based policy predicts future 3D point tracks, and robot actions are further recovered through rigid-body geometry constraints. Inspired by this work, EgoZero~\citep{liu2025egozero} also overcomes the morphology gap between human videos and robot execution by representing states and actions as compact sets of points. Moreover, it removes the reliance of Point Policy~\citep{haldar2025point} on a multi-camera calibration setup by extracting the same point-based state-action representation directly from smart-glasses recordings. In contrast to using hand-object keypoints as interaction affordances, some related works take human hand pose and object point clouds as joint inputs to train directly executable robot policies with human videos. \cite{ye2023learning} propose a Continuous Grasping Function to generate smooth, continuous dexterous grasping trajectories with paired human hand poses and object point clouds as input. They first retarget large-scale human hand-object interaction trajectories into robot demonstrations and then learn an implicit generative policy over time. Similar to \cite{ye2023learning}, \cite{chen2025web2grasp} also learn directly executable dexterous grasping policies from joint hand-object geometry. However, the policy outputs the desired pairwise distances between the robot hand and object point clouds in the target grasp pose. YOTO~\citep{zhou2025you} also treats bimanual hand poses together with manipulated object point clouds as the interaction-centric affordances for robot policy construction. It adopts a data proliferation strategy, which efficiently augments a single demonstration across varied objects and placements. \cite{chen2026dexterous} instead use DemoGen~\citep{xue2025demogen} to synthesize diverse training trajectories from a single reconstructed interaction while preserving hand-object contact structure. To solve the challenge of learning screw motion from human videos, ScrewMimic~\citep{bahety2024screwmimic} extracts bimanual wrist poses together with grasp contact points from the human video, and interprets the relative motion between the two hands as a screw action. It then ground hand motion on the manipulated object through its 3D point cloud, enabling a PointNet model to predict executable screw actions for robot bimanual manipulation. PAWS~\citep{wang2026paws} further extends interaction-centric affordance extraction to articulated object manipulation. It extracts hand trajectories and object joint structure as scalable articulation supervision. To achieve open-world imitation from observation, \cite{zhu2024vision} extract an open-world object graph from a single human video, grounding hands and manipulated objects along with their attributes and relations into an object-centric manipulation plan. A robot policy is then conditioned on this graph-structured plan, enabling generalization to novel open-world settings. Considering this work can only be deployed on single-arm manipulators, OKAMI~\citep{li2024okami} uses vision foundation models to identify task-relevant objects and retargets upper-body motion from a human video. The rollout trajectories from object-aware warping are used to train closed-loop visuomotor policies via behavioral cloning. This work facilitates open-world imitation from observation~\citep{zhu2024vision} more practical for humanoid manipulation. In contrast, \cite{heidinger20252handedafforder} argue that the specific object regions are more important in the context of human-object interactions. Thus, they propose a paradigm for precisely extracting actionable affordance regions from human-object interaction videos. A VLM-based model is further optimized to predict pixel-wise affordance segments in images.

\textit{(f) Conclusion for affordances:} \textit{Affordances as a bridge} provide the most explicit action-level interface between human videos and robot execution by exposing where and how interactions should occur. Related works have evolved from extracting relatively simple hand-centric and object-centric cues to interaction-centric representations that additionally capture coupled hand-object dynamics. Correspondingly, affordances have been used with increasingly tighter coupling to robot control, ranging from backbone pretraining and reward construction to policy conditioning and direct policy generation. This progression reflects a clear trend toward more structured, executable, and generalizable representations for cross-embodiment transfer. Considering that affordance-based bridging mechanisms constitute the largest body of work with diverse extracted human motion data in the LfHV literature, we summarize the representative papers reviewed in this section in Tab.~\ref{tab:affordance_backbone}, Tab.~\ref{tab:affordance_reward}, Tab.~\ref{tab:affordance_condition}, and Tab.~\ref{tab:affordance_policy}. We present their adopted hand affordance format, object affordance format, hand-object joint affordance format, and end-effector type. Detailed implementations underlying individual entries can be found in the corresponding original papers. Here are some key findings across these statistics:

\begin{itemize}[leftmargin=1em]
\setlength{\parskip}{0pt}
\item Affordances for backbone training are mostly used as scalable supervision signals. Thus, they are often hand-centric and in relatively simple formats. In contrast, affordances as robot policy are the most execution-oriented, covering the richest spectrum from hand-centric to object-centric and interaction-centric formulations. This trend suggests that relational HOI structure becomes increasingly important as affordances move closer to directly shaping robot behavior.

\item Reward construction with affordances relies much more heavily on multimodal hand-object coupling. Compared with backbone training, reward-oriented methods more frequently combine hand poses, object poses, and explicit interaction cues such as grasp states and contact regions. This pattern indicates that rewards must evaluate whether an interaction is successful, rather than merely whether a representation is informative.

\item Affordances used as policy conditions are strikingly homogeneous in the end-effector type. Concretely, all representative works in this category target parallel grippers. This is an interesting contrast to the other three categories, where dexterous hands appear regularly. This suggests that current conditional affordance interfaces are mostly compact and low-dimensional, which are more naturally compatible with simpler gripper control pipelines than with the fine-grained control required by dexterous hands.

\item Conversely, the end-effector type is strongly correlated with affordance granularity. Related works using dexterous hands are much more likely to use high-dimensional hand representations such as MANO parameters, joint angles, and fingertip positions, while parallel-gripper methods more often rely on sparse hand trajectories, motion flow, contact points, and hand-object masks. This reflects a trade-off between embodiment complexity and affordance compactness.
\end{itemize}

Despite the rapid development of affordance-based transfer, its effectiveness still depends heavily on reliable HOI analysis, accurate spatial grounding, and robust retargeting. This is particularly challenged in complex manipulation tasks, especially in open-world settings with severe occlusion, large viewpoint variation, and diverse object and effector categories.

\begin{figure*}[t]
  \centering
  \resizebox{\linewidth}{!}{%
  \begin{tikzpicture}[x=1cm,y=1cm,>=Stealth]
    \definecolor{timelineblue}{RGB}{120,138,150}
    \definecolor{timelinebrick}{RGB}{171,128,120}
    \definecolor{timelinegray}{RGB}{218,220,221}
    \definecolor{timelineyear}{RGB}{118,116,112}
    \definecolor{timelinebubble}{RGB}{166,206,227}

    \draw[timelinegray,line width=4.2pt,-{Stealth[length=3.6mm]}] (-0.10,0) -- (21.70,0);

    \foreach \x/\year in {0.45/2024,2.65/2025,11.45/2026} {
      \fill[timelinebubble,draw=white,line width=0.7pt] (\x,0) circle (0.17);
      \node[font=\bfseries\scriptsize,text=timelineyear] at (\x,-0.36) {\year};
    }

    \draw[timelineblue,line width=2.5pt,-{Stealth[length=2.2mm]}] (1.00,0.05) -- (1.00,0.95);
    \node[align=center,font=\fontsize{7.0}{7.8}\selectfont,text=timelineblue] at (1.00,1.24) {\timelinecite{ye2024latent}};

    \draw[timelineblue,line width=2.5pt,-{Stealth[length=2.2mm]}] (2.10,-0.05) -- (2.10,-0.95);
    \node[align=center,font=\fontsize{7.0}{7.8}\selectfont,text=timelineblue] at (2.10,-1.24) {\timelinecite{chen2024igor}};

    \draw[timelineblue,line width=2.5pt,-{Stealth[length=2.2mm]}] (3.20,0.05) -- (3.20,1.30);
    \node[align=center,font=\fontsize{7.0}{7.8}\selectfont,text=timelineblue] at (3.20,1.60) {\timelinecite{chen2025moto}};

    \draw[timelineblue,line width=2.5pt,-{Stealth[length=2.2mm]}] (4.30,-0.05) -- (4.30,-1.30);
    \node[align=center,font=\fontsize{7.0}{7.8}\selectfont,text=timelineblue] at (4.30,-1.60) {\timelinecite{bu2025agibot}};

    \draw[timelineblue,line width=2.5pt,-{Stealth[length=2.2mm]}] (5.40,0.05) -- (5.40,0.95);
    \node[align=center,font=\fontsize{7.0}{7.8}\selectfont,text=timelineblue] at (5.40,1.24) {\timelinecite{bu2025univla}};

    \draw[timelineblue,line width=2.5pt,-{Stealth[length=2.2mm]}] (6.50,-0.05) -- (6.50,-0.95);
    \node[align=center,font=\fontsize{7.0}{7.8}\selectfont,text=timelineblue] at (6.50,-1.24) {\timelinecite{kim2025uniskill}};

    \draw[timelineblue,line width=2.5pt,-{Stealth[length=2.2mm]}] (12.00,0.05) -- (12.00,0.95);
    \node[align=center,font=\fontsize{7.0}{7.8}\selectfont,text=timelineblue] at (12.00,1.24) {\timelinecite{garrido2026learning}};

    \draw[timelineblue,line width=2.5pt,-{Stealth[length=2.2mm]}] (13.10,-0.05) -- (13.10,-1.30);
    \node[align=center,font=\fontsize{7.0}{7.8}\selectfont,text=timelineblue] at (13.10,-1.60) {\timelinecite{lee2026mvp}};

    \draw[timelineblue,line width=2.5pt,-{Stealth[length=2.2mm]}] (14.20,0.05) -- (14.20,1.30);
    \node[align=center,font=\fontsize{7.0}{7.8}\selectfont,text=timelineblue] at (14.20,1.60) {\timelinecite{gao2026dreamdojo}};

    \draw[timelineblue,line width=2.5pt,-{Stealth[length=2.2mm]}] (15.30,-0.05) -- (15.30,-1.30);
    \node[align=center,font=\fontsize{7.0}{7.8}\selectfont,text=timelineblue] at (15.30,-1.60) {\timelinecite{govind2026unilact}};

    \draw[timelineblue,line width=2.5pt,-{Stealth[length=2.2mm]}] (16.40,0.05) -- (16.40,0.95);
    \node[align=center,font=\fontsize{7.0}{7.8}\selectfont,text=timelineblue] at (16.40,1.24) {\timelinecite{sun2026vla}};

    \draw[timelinebrick,line width=2.5pt,-{Stealth[length=2.2mm]}] (7.60,0.05) -- (7.60,1.30);
    \node[align=center,font=\fontsize{7.0}{7.8}\selectfont,text=timelinebrick] at (7.60,1.60) {\timelinecite{chen2025villa}};

    \draw[timelinebrick,line width=2.5pt,-{Stealth[length=2.2mm]}] (8.70,-0.05) -- (8.70,-1.30);
    \node[align=center,font=\fontsize{7.0}{7.8}\selectfont,text=timelinebrick] at (8.70,-1.60) {\timelinecite{routray2025vipra}};

    \draw[timelinebrick,line width=2.5pt,-{Stealth[length=2.2mm]}] (9.80,0.05) -- (9.80,0.95);
    \node[align=center,font=\fontsize{7.0}{7.8}\selectfont,text=timelinebrick] at (9.80,1.24) {\timelinecite{bi2025motus}};

    \draw[timelinebrick,line width=2.5pt,-{Stealth[length=2.2mm]}] (10.90,-0.05) -- (10.90,-0.95);
    \node[align=center,font=\fontsize{7.0}{7.8}\selectfont,text=timelinebrick] at (10.90,-1.24) {\timelinecite{li2025latbot}};

    \draw[timelinebrick,line width=2.5pt,-{Stealth[length=2.2mm]}] (17.50,-0.05) -- (17.50,-0.95);
    \node[align=center,font=\fontsize{7.0}{7.8}\selectfont,text=timelinebrick] at (17.50,-1.24) {\timelinecite{shi2026care}};

    \draw[timelinebrick,line width=2.5pt,-{Stealth[length=2.2mm]}] (18.60,0.05) -- (18.60,1.30);
    \node[align=center,font=\fontsize{7.0}{7.8}\selectfont,text=timelinebrick] at (18.60,1.60) {\timelinecite{luo2026joint}};

    \draw[timelinebrick,line width=2.5pt,-{Stealth[length=2.2mm]}] (19.70,-0.05) -- (19.70,-1.30);
    \node[align=center,font=\fontsize{7.0}{7.8}\selectfont,text=timelinebrick] at (19.70,-1.60) {\timelinecite{zhang2026clap}};

    \draw[timelinebrick,line width=2.5pt,-{Stealth[length=2.2mm]}] (20.80,0.05) -- (20.80,0.95);
    \node[align=center,font=\fontsize{7.0}{7.8}\selectfont,text=timelinebrick] at (20.80,1.24) {\timelinecite{dai2026conla}};

    \draw[timelineblue,line width=2.6pt,rounded corners=1pt] (0.10,-2.05) -- (0.78,-2.05);
    \node[anchor=west,font=\bfseries\footnotesize,text=timelineblue] at (0.92,-2.05) {(a) Latent actions from visual reconstruction};

    \draw[timelinebrick,line width=2.6pt,rounded corners=1pt] (0.10,-2.42) -- (0.78,-2.42);
    \node[anchor=west,font=\bfseries\footnotesize,text=timelinebrick] at (0.92,-2.42) {(b) Latent actions from visual-action co-reconstruction};

  \end{tikzpicture}%
  }
  \caption{Chronological overview of methods under \textit{latent actions as a bridge} in Sec.~\ref{sec:latent_action_transfer}.}
  \label{fig:latent_actions_timeline}
\end{figure*}
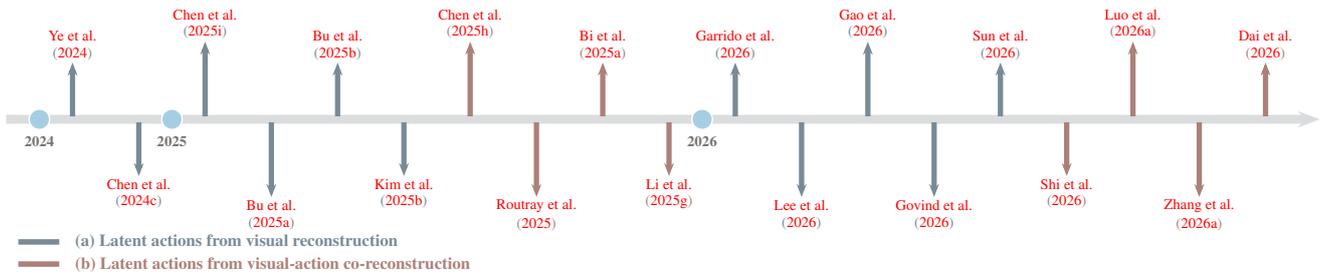

\begin{figure}[t]
  \centering
  \includegraphics[width=1\linewidth]{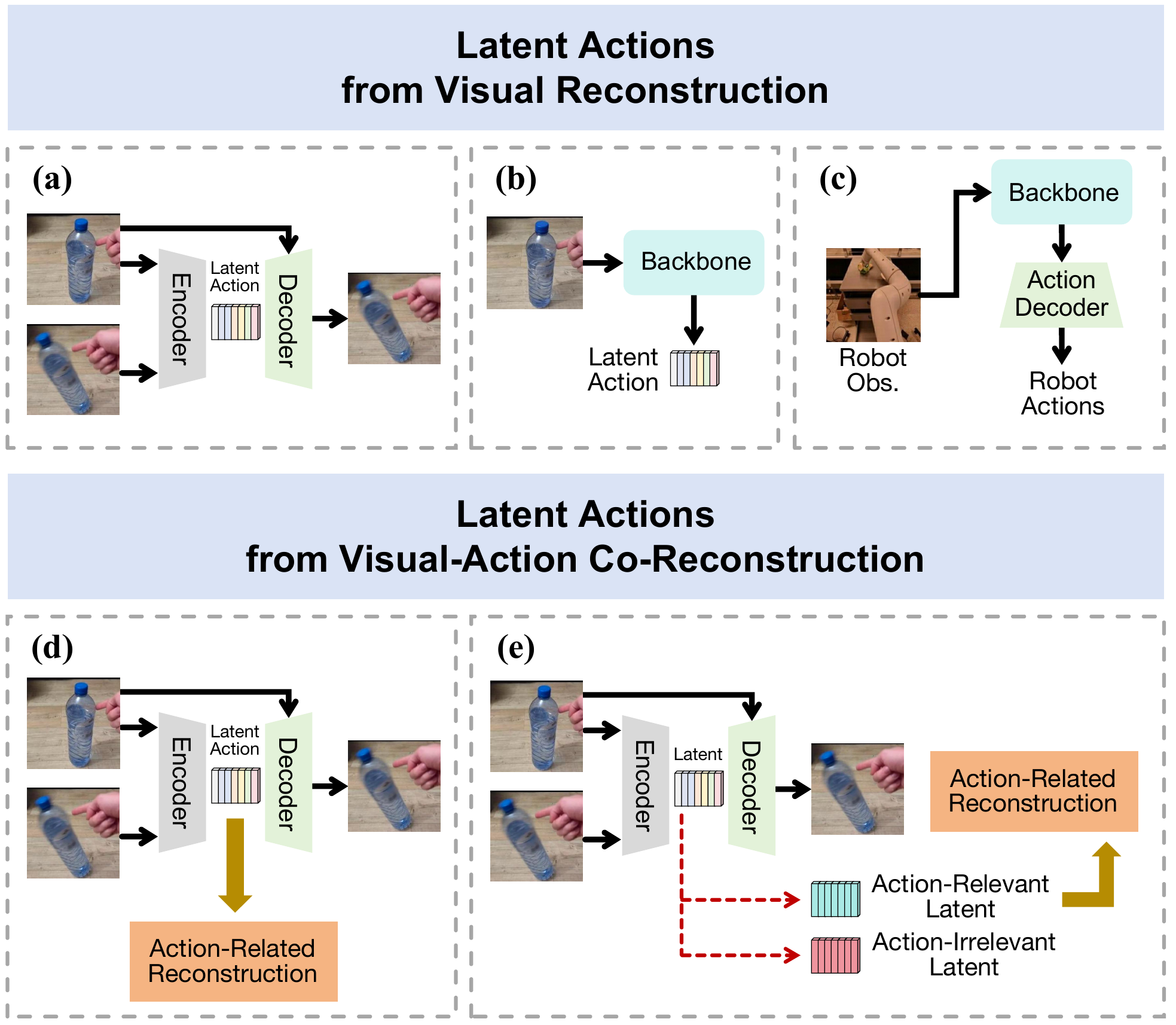}
  \caption{High-level diagram of \textit{latent actions as a bridge}. Some elements of the illustrations are adapted from \cite{ye2024latent}.}
  \label{fig:latent_action_transfer}
\end{figure}

\subsubsection{Latent Actions} \label{sec:latent_action_transfer}

Compared with affordances that explicitly encode geometric interaction cues, \textit{latent actions as a bridge} aim to learn compact and transferable action priors directly from unlabeled human videos. The typical paradigm is to use a latent action model (LAM), in which an inverse dynamics model (IDM) infers a latent action from adjacent observations, while a forward dynamics model (FDM) reconstructs or predicts the future observation conditioned on the current observation and the inferred latent action. Different FDM reconstruction objectives further bias the learned latent actions toward different aspects of motion, such as coarse visual change or more task-relevant action semantics. Therefore, we divide this trendy latent-action-based bridging into two categories: \textit{latent actions from visual reconstruction}, which learns latent actions solely by reconstructing future visual observations, and \textit{latent actions from visual-action co-reconstruction}, which further constrains latent actions to jointly recover visual transitions and action-related outputs. The timeline of related work is showcased in Fig.~\ref{fig:latent_actions_timeline}.

\textit{(a) Latent actions from visual reconstruction:} LAPA~\citep{ye2024latent} provides a pioneering example of learning latent actions from visual reconstruction. It extracts discrete latent actions from large-scale unlabeled human videos through an encoder trained with a VQ-VAE-based visual reconstruction objective (see Fig.~\ref{fig:latent_action_transfer}(a)). These latent actions are then used as pseudo-action labels to pretrain a latent VLA model (see Fig.~\ref{fig:latent_action_transfer}(b)). This enables robot policies to absorb generic motion priors from Internet human videos when fine-tuning with robot demonstrations (see Fig.~\ref{fig:latent_action_transfer}(c)). LAPA can be seamlessly integrated into large heterogeneous pretraining pipelines such as GR00T N1~\citep{bjorck2025gr00t} and SurgWorld~\citep{he2025surgworld}. As a concurrent work, IGOR~\citep{chen2024igor} learns a unified latent action space by compressing the visual change between an initial image and its goal state. It further emphasizes cross-embodiment semantic consistency, enabling the latent actions inferred from human videos to jointly support foundation policy and world model training across both humans and robots.
Moto~\citep{chen2025moto} instead introduces an end-to-end model to autoregressively predict a trajectory of latent motion tokens for future video clips. It uses an M-Former with query tokens to extract latent motion tokens from videos. During robot policy learning, action query tokens are appended to predict robot actions, co-finetuning motion token prediction and robot action planning. GO-1~\citep{bu2025agibot} similarly introduces sequential latent action tokens as an intermediate planning interface. However, it casts them more explicitly as a hierarchical planning interface in the Vision-Language-Latent-Action (ViLLA) framework for large-scale robot learning. This work also incorporates failure recovery data during training data collection, and further adopts a human-in-the-loop approach to assess and refine data quality.

Considering the prior works implicitly incorporate task-irrelevant dynamics such as the disturbance of camera shake, visual noise, and other agents, \cite{bu2025univla} focus on task-centric latent actions to decouple task-relevant dynamics from irrelevant visual changes. Specifically, they first learn generic latent actions from unlabeled videos and then refine them with language instructions in the DINO feature space, finally leading to faster convergence and more robust performance. Also motivated by the fact that latent actions are susceptible to distractors, \cite{garrido2026learning} follows the standard IDM-FDM paradigm of latent action modeling, but redesigns the latent action world model with continuous constrained actions and explicit information regularization. To further suppress viewpoint distractions in latent action learning, MVP-LAM~\citep{lee2026mvp} learns discrete action-centric latent actions from time-synchronized multi-view videos through a cross-viewpoint reconstruction objective. By requiring a latent action inferred from one view to explain future observations from another view, it reduces reliance on viewpoint-specific cues and yields pseudo-action labels that are more informative about the underlying actions. UniSkill~\citep{kim2025uniskill} instead treats latent actions as explicit skill representations and directly trains a skill-conditioned policy on the learned representations. This framework requires an additional human video prompt to extract a sequence of embodiment-agnostic skills, which are then executed by a skill-conditioned robot policy. DreamDojo~\citep{gao2026dreamdojo} further extends this direction to action-conditioned world modeling by introducing continuous latent actions as unified proxy actions for large-scale human videos. Unlike prior works which mainly use latent actions for policy pretraining or intermediate goal abstraction, this work uses them to condition future video prediction in a generalist robot world model. More recently, to further address the lack of explicit 3D geometry in RGB-based latent actions, \cite{govind2026unilact} introduce UniLACT, which incorporates geometric structure through depth-aware latent pretraining. This model learns shared RGB-depth latent actions with explicit cross-modal interactions, such that the resulting pseudo-action labels encode stronger spatial priors for downstream robotic manipulation. To further mitigate the issue of information leakage and pixel-level bias in latent action learning, \cite{sun2026vla} propose VLA-JEPA, a JEPA-style~\citep{assran2025v} latent world modeling framework that learns action-relevant state transitions via latent-space prediction. Instead of reconstructing future frames or directly conditioning on them, future observations are only used as supervision targets, preventing latent actions from collapsing into shortcuts. By predicting future latent states conditioned on current latent action and past latent states, this method encourages more robust and action-centric representations, while simplifying the training pipeline compared to prior multi-stage approaches.

\begin{table*}[t]
\centering
\scriptsize
\setlength{\tabcolsep}{3.2pt}
\caption{Comparison of representative methods that formulate latent actions as a bridge.}
\label{tab:latent_action}
\begin{tabularx}{\linewidth}{p{2.8cm}p{3.3cm}p{1.8cm}X X}
\toprule
Reference & Supervision signal & Disentanglement & Model for IDM & Model for FDM \\
\midrule
\rowcolor{gray!15}
\multicolumn{5}{l}{\textit{Latent actions from visual reconstruction}} \\
\cite{ye2024latent} & Pixel & No & C-C-ViViT tokenizer & LWM-Chat-1M \\
\cite{chen2024igor} & Pixel & No & ViT (encoder + decoder) + ST-Transformer (encoder) & Open-Sora \\
\cite{bjorck2025gr00t} & Pixel & No & C-ViViT tokenizer + DiT & Eagle-2 VLM \\
\cite{chen2025moto} & Pixel & No & M-Former + ViT (encoder + decoder) & Moto-GPT \\
\cite{bu2025agibot} & Pixel & No & C-ViViT tokenizer & InternVL2.5-2B \\
\cite{bu2025univla} & Future DINOv2 feature & Yes & DINOv2 + C-ViViT tokenizer & Prismatic-7B \\
\cite{kim2025uniskill} & Pixel + Depth & No & Depth encoder + ST Transformer encoder + diffusion decoder & Diffusion policy \\
\cite{garrido2026learning} & Pixel + latent bottleneck regularization & No & V-JEPA2 encoder + MLP/Transformer & ViT + RoPE + AdaLN-zero world model \\
\cite{lyu2026lda} & Future DINOv3 feature & No & Qwen3-VL + DINOv3 + MM-DiT & Multi-modal diffusion Transformer \\
\cite{lee2026mvp} & Self-view and cross-view future DINO feature & No & DINOv2 + ST-Transformer encoder + spatial Transformer decoder + VQ-tokenizer & Prismatic-7B \\
\cite{gao2026dreamdojo} & Pixel & No & Spatiotemporal Transformer VAE & Cosmos-Predict2.5 \\
\cite{govind2026unilact} & Pixel + depth & No & Diffusion policy & GPT-2 \\
\cite{sun2026vla} & Future V-JEPA2 feature & No & Qwen3-VL & Autoregressive Transformer-based world model \\
\midrule
\rowcolor{gray!15}
\multicolumn{5}{l}{\textit{Latent actions from visual-action co-reconstruction}} \\
\cite{chen2025villa} & Pixel + proprioception & No & ST-Transformer (encoder) + ViT (visual decoder) + MLP & PaliGemma-3B \\
\cite{routray2025vipra} & Pixel + perception + optical flow  & No & DINOv2 + ST-Transformer (encoder + decoder) & LWM-Chat-1M \\
\cite{bi2025motus} & Optical flow & No & DC-AE (encoder + decoder) & Qwen2.5-VL-2B + Wan 2.1 5B + Transformer \\
\cite{li2025latbot} & Pixel + action & Yes & InternVL3.5-2B encoder + SANA-1.6B decoder & PaliGemma-3B \\
\cite{shi2026care} & Future visual feature + keypoint trajectory & No & Prismatic-7B & Prismatic-7B \\
\cite{luo2026joint} & Motion token + hidden state & No & Query cross-sttention encoder + self-attention + MLP + VQ-tokenizer & InternVL3-2B \\
\cite{zhang2026clap} & Future DINOv3 feature + action (contrastive) & Yes & DINOv3 + ST-Transformer encoder + spatial Transformer decoder + VQ-tokenizer & Qwen3VL-4B \\
\cite{dai2026conla} & Pixel + action (contrastive) + vision (contrastive) & Yes & ST Transformer encoder + VQ-tokenizer + spatial Transformer decoder & Large World Model-7B \\
\bottomrule
\end{tabularx}
\end{table*}

\textit{(b) Latent actions from visual-action co-reconstruction:} Some related works reconstruct additional action-related information to encode more explicit motion dynamics into latent actions (see Fig.~\ref{fig:latent_action_transfer}(d)), rather than merely compressing frame differences. villa-X~\citep{chen2025villa} advances the aforementioned ViLLA paradigm by an additional proprioceptive forward dynamics branch. It factorizes policy learning into a latent action expert and a robot action expert, where latent actions condition robot action prediction. Meanwhile, the learned latent actions are constrained not only by visual reconstruction but also by future robot states and actions. \cite{routray2025vipra} instead incorporate the optical flow consistency loss into latent action model training. This loss for motion flow reconstruction encourages predicted frames to capture underlying action patterns, supporting temporally coherent dynamics. \cite{bi2025motus} also implement optical-flow-based motion reconstruction for latent action represention. During training, they mix 90\% unlabeled data for self-supervised reconstruction with 10\% labeled trajectories for weak action supervision. In contrast to optical flow, CARE~\citep{shi2026care} learns continuous latent actions through a multi-task objective, using keypoint trajectory prediction from CoTracker as an auxiliary signal for latent action learning. JALA~\citep{luo2026joint} learns jointly-aligned latent actions by matching predictive motion embeddings with both inverse dynamics and real actions. This design directly biases latent actions toward transition-aware behavior modeling, making them more scalable for VLA pretraining on heterogeneous in-the-wild human data. 

Moreover, some researchers explicitly disentangle action-relevant and action-irrelevant components within latent representations (see Fig.~\ref{fig:latent_action_transfer}(e)), thereby making the learned latent actions more transferable and physically grounded. For example, LatBot~\citep{li2025latbot} decomposes latent actions into learnable motion tokens and scene tokens to separate robot-induced motion from passive environmental changes. These tokens are jointly used to guide future frame reconstruction and inter-frame action reconstruction. CLAP~\citep{zhang2026clap} instead tackles the visual entanglement problem by contrastively aligning the action-relevant latent factors inferred from video transitions with a physical action space derived from robot trajectories. This cross-modal alignment forces the latent space to preserve manipulation-relevant dynamics while filtering out background shifts and other irrelevant visual factors. More recently, ConLA~\citep{dai2026conla} further addresses shortcut learning in latent action extraction by introducing a contrastive disentanglement module. It leverages action category and temporal priors from human videos. Instead of relying solely on reconstruction losses, it encourages semantically similar motions to cluster compactly across different environments while separating action dynamics from irrelevant visual content. Therefore, this work produces more motion-centric latent actions for downstream robot policy learning.

\textit{(c) Conclusion for latent actions:} \textit{Latent actions as a bridge} provide a compact alternative to explicit affordance extraction by learning transferable motion priors directly from unlabeled human videos. Existing works have evolved from purely visual reconstruction of single-step latent actions to richer formulations. These newer formulations incorporate task-centric filtering, multi-view consistency, depth cues, world modeling, autoregressive latent planning, and joint reconstruction of action-related signals such as proprioception, optical flow, keypoint trajectories, and executable robot actions. Latent action learning is gradually moving away from merely compressing inter-frame appearance changes toward capturing more structured, temporally coherent, and physically grounded action semantics. Compared with explicit affordances, this paradigm is particularly attractive for cross-embodiment scalability, since it does not rely on costly geometric annotation or precise HOI parsing and can naturally exploit large-scale in-the-wild human videos. For easier reference, we summarize the current works reviewed in this section in Tab.~\ref{tab:latent_action}, including the supervision signal, whether latent-action disentanglement is explicitly modeled, and the corresponding model designs for IDM and FDM. Precise implementation details underlying individual entries can be found in the corresponding papers. It is expected that this summary will provide a useful reference for future research on the design of latent action models.

Nonetheless, a major challenge in this category still exists in disentangling action-relevant dynamics from nuisance visual factors such as camera egomotion and embodiment-irrelevant background changes. In addition, although latent actions can capture transferable motion structure at scale, grounding them into embodiment-specific control still requires extra robot data supervision. These issues become more pronounced in contact-rich interactions because physically meaningful latent action abstraction is harder to learn from raw videos alone.

\subsubsection{Summary for Action-Oriented Transfer}

Compared with task-oriented and observation-oriented transfer, action-oriented transfer provides the most direct bridge from human videos to robot execution. Its central goal is no longer to understand \textit{which task should be done} or \textit{how the scene should be perceived}, but to extract representations that can more tightly constrain \textit{how the robot should act}. In this category, \textit{affordances as a bridge} and \textit{latent actions as a bridge} represent two complementary philosophies for action transfer. Affordance-based methods emphasize explicit action grounding. They expose geometric interaction cues such as hand trajectories, object motion flow, and hand-object relations. Thus, they provide interpretable interfaces for retargeting, reward shaping, policy conditioning, and direct policy construction. In contrast, latent-action-based methods compress behavior into implicit action abstractions learned from large-scale videos. They sacrifice geometric explicitness in exchange for better scalability and broader coverage of in-the-wild human data. Their difference essentially reflects a tradeoff between \textit{physical interpretability} and \textit{data scalability}: affordances offer stronger grounding and controllability, whereas latent actions provide a more economical route for absorbing large-scale behavioral priors.

A clear trend in this category is that action-oriented transfer is moving toward increasingly structured and executable representations. Affordance-based methods have evolved from simple hand-centric cues to richer object-centric and interaction-centric formulations. Latent-action methods have progressed from appearance-driven reconstruction to more temporally coherent, task-centric, and physically informed motion abstractions. These developments suggest that successful action transfer requires capturing visible motion patterns while preserving the interaction structure that makes those motions executable across embodiments. Nonetheless, both directions still share a common bottleneck. Action representations learned from human videos must ultimately be grounded in robot-compatible action spaces, preserving physical validity and cross-domain generality. Therefore, future progress may depend on combining the strengths of the two paradigms. For example, explicit affordances could impose stronger physical constraints on latent actions, while latent action learning could improve the scalability and robustness of affordance-based transfer. Such integration may be essential for making action-oriented transfer a more reliable interface for scalable robot learning from human videos.

\subsection{Bridging Mechanisms Across Data Configurations and Learning Paradigms}

The above taxonomy is organized by \textit{what} intermediate information is transferred from human videos to robots. An equally important cross-family question is \textit{which data configurations} and \textit{which learning paradigms} each bridge most naturally supports. For data configurations, we attend to the different viewpoint preferences and the extent of dependence on robot data. For learning paradigms, imitation learning, reinforcement learning, and other relevant policy learning schemes are discussed. Here, we analyze these two dimensions using the dominant design choice in reviewed works. We hope this can inspire future works on the selection of data sources and learning paradigms in the LfHV literature.

\begin{table*}[t]
\centering
\scriptsize
\setlength{\tabcolsep}{4pt}
\renewcommand{\arraystretch}{1.15}
\begin{tabularx}{\textwidth}{
>{\raggedright\arraybackslash}p{2.6cm}
>{\raggedright\arraybackslash}p{1.6cm}
X
>{\centering\arraybackslash}p{1.6cm}
}
\toprule
Oriented transfer & View type & Reference & Percentage \\
\midrule

\multirow{5}{*}{Task-oriented}
& Ego
& \cite{pertsch2022cross,lin2025physbrain,li2026act}
& 13\% \\

& Exo
& \cite{yang2015robot,sermanet2016unsupervised,nguyen2018translating,yu2018one,yu2018one_hil,sharma2019third,yang2022learning,xu2023xskill,ding2024knowledge,jain2024vid2robot,chen2024vlmimic,clark2025action,chen2025fmimic,hori2025interactive,hori2025robot}
& 65\% \\

& Ego+Exo
& \cite{jang2022bc,wake2024gpt,wang2024vlm,ma2025egoloc,ye2025watch}
& 22\% \\

\midrule

\multirow{10}{*}{Observation-oriented}
& Ego
& \cite{ma2022vip,nair2022r3m,bhateja2023robotic,chang2023look,duan2023ar2,ma2023liv,majumdar2023we,mendonca2023structured,wu2023unleashing,li2024ag2manip,liu2024masked,zeng2024learning,goswami2025world,jiang2025rynnvla,lepert2025masquerade,li2025h2r,li2025mimicdreamer,punamiya2025egobridge,song2025mitty,sun2025vtao,xiong2025ag2x2,freeman2026warped,ye2026visual}
& 49\% \\

& Exo
& \cite{sermanet2017time,liu2018imitation,schmeckpeper2020reinforcement,smith2020avid,xiong2021learning,bahl2022human,sun2022learning,zakka2022xirl,jain2024vid2robot,qian2024contrast,zhu2024vision,ci2025h2r,kedia2025one,lepert2025phantom,liu2025immimic,shah2025mimicdroid,tang2025trajectory,zhang2025generative,zhu2025learning}
& 40\% \\

& Ego+Exo
& \cite{xiao2022masked,xiong2022robotube,dasari2023unbiased,radosavovic2023real,cheang2024gr}
& 11\% \\

\midrule

\multirow{18}{*}{Action-oriented}
& Ego
& \cite{lee2017learning,bahl2023affordances,bharadhwaj2023towards,bharadhwaj2023zero,kannan2023deft,shaw2023videodex,ju2024robo,kuang2024ram,srirama2024hrp,yuan2024general,ye2024latent,cai2025n,cheang2025gr,chen2025vidbot,feng2025spatial,heidinger20252handedafforder,hsu2025spot,jiang2025rynnvla,kareer2025egomimic,kareer2025emergence,li2025scalable,liu2025egozero,luo2025being,ma2025egoloc,ma2025uni,papagiannis2025r,qiu2025humanoid,shi2025zeromimic,wen2025gr,werby2025articulated,yang2025ar,yang2025egovla,yoshida2025developing,yuan2025motiontrans,zhang2025actron3d,zhang2025zero,bi2025motus,bu2025univla,bi2026h,chen2026dexterous,gao2026dreamdojo,li2025latbot,luo2026being,luo2026joint,lyu2026lda,soraki2026objectforesight,wang2026paws,zhang2026clap,zhang2026unidex,zheng2026egoscale,zhu2026emma}
& 52\% \\

& Exo
& \cite{sieb2020graph,das2021model,arunachalam2022dexterous,bahl2022human,qin2022dexmv,qin2022one,wen2022you,gao2023k,gu2023rt,ko2023learning,kumar2023graph,wang2023mimicplay,wang2023robot,wen2023any,ye2023learning,bahety2024screwmimic,heppert2024ditto,li2024okami,xu2024flow,chen2025graphmimic,chen2025vividex,dan2025x,garrido2026learning,haldar2025point,hsieh2025dexman,jonnavittula2025view,li2025novaflow,lum2025crossing,park2025demodiffusion,ren2025motion,shan2025slot,singh2025deep,singh2025hand,spiridonov2025generalist,tang2025functo,tang2025mimicfunc,yang2025tra,yin2025object,zhao2025dexh2r,zhou2025human,zhou2025you}
& 41\% \\

& Ego+Exo
& \cite{mandikal2022dexvip,sivakumar2022robotic,bharadhwaj2024track2act,bjorck2025gr00t,chen2024igor,chen2025villa,kim2025uniskill,lee2026mvp}
& 7\% \\

\bottomrule
\end{tabularx}
\caption{LfHV studies across egocentric, exocentric, and egocentric+exocentric video configurations.}
\label{tab:oriented_ego_exo_breakdown}
\end{table*}

\begin{figure*}[t]
  \centering
  \includegraphics[width=1\linewidth]{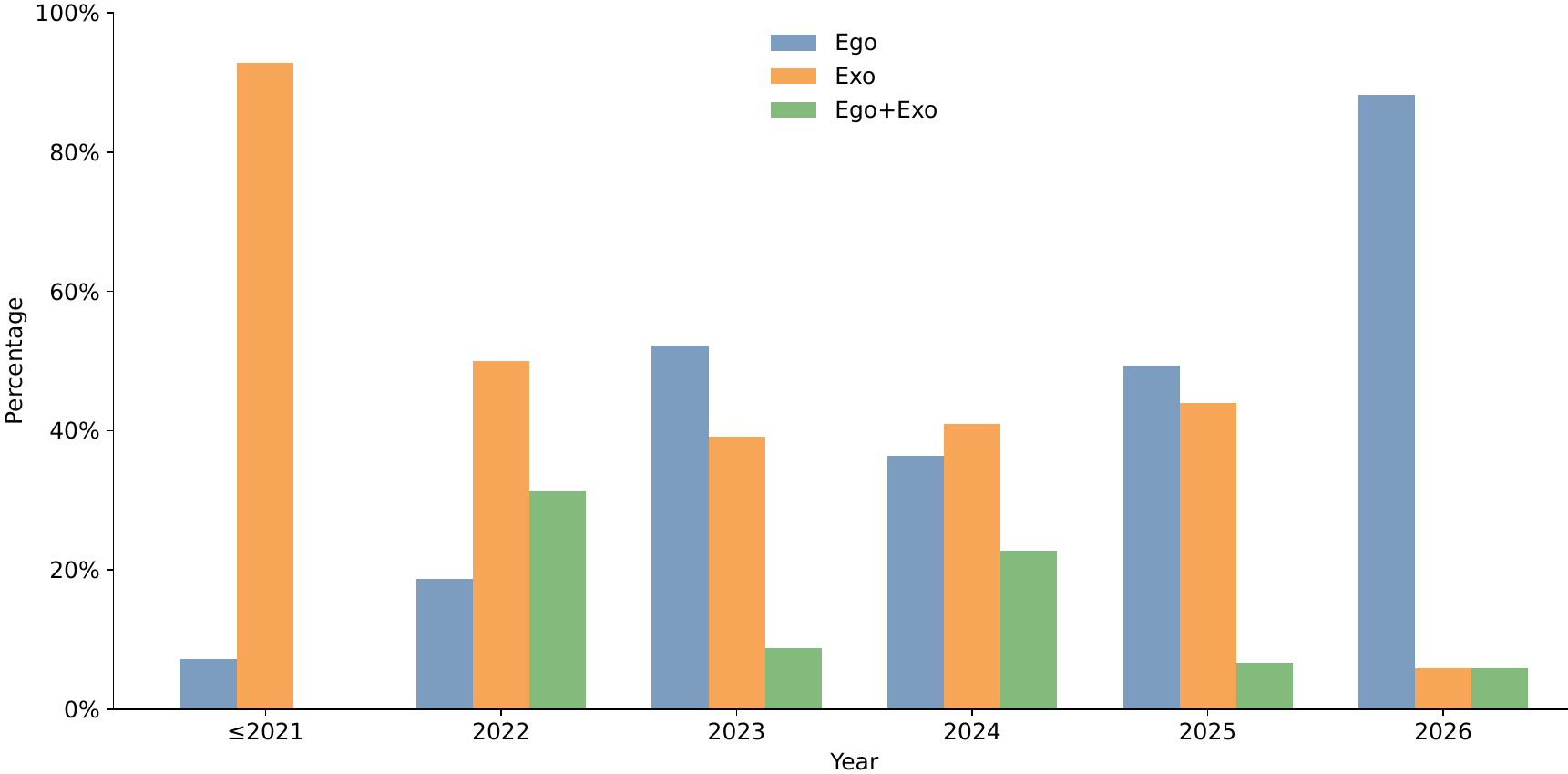}
  \caption{Popularity trends of egocentric, exocentric, and egocentric+exocentric video configurations.}
  \label{fig:yearly_grouped_bar_all}
  \vspace{-0.3cm}
\end{figure*}

\subsubsection{Across Data Configurations}

In terms of data configurations, the first central distinction is between exocentric (third-person) and egocentric (first-person) human videos. Exocentric videos usually provide a cleaner global scene layout and multi-stage task context. Thus, they are attractive when the bridge operates at coarse task and scene granularity. In contrast, egocentric videos expose hand-object contact, manipulation order, and first-person manipulation cues more directly. They are particularly valuable when the bridge must preserve fine-grained interaction geometry or precise action timings. In addition, the other distinction lies in the extent of reliance on robot data. Some LfHV methods are designed to learn only from human videos, while others still require real-world robot demonstrations or interactions with environments to ground the transferred knowledge into executable control. In this section, we summarize the detailed taxonomy of data configurations, including viewpoints (i.e., purely exocentric videos, purely egocentric videos, and mixed exocentric-egocentric settings), and dependence on real-world robot data (i.e., human videos + real-world robot demonstration, human videos + real-world interaction, human videos only). We only list works with strong evidence for these categories.

\textit{(a) Viewpoint preference across transfer families and over time:} The three transfer families, i.e., task-, observation-, and action-oriented transfer, exhibit clearly different viewpoint affinities. As shown in Tab.~\ref{tab:oriented_ego_exo_breakdown}, task-oriented transfer is dominated by exocentric videos (65\%), with only a small fraction relying purely on egocentric data (13\%). This bias is structurally reasonable. Task-oriented methods aim to recover global procedural structure, subtask boundaries, and potential intents. These signals are easier to infer from a stable external viewpoint that preserves the full scene-level arrangement and reduces camera egomotion. In contrast, first-person views often emphasize local interaction details at the expense of holistic scene context. This makes long-horizon task decomposition less direct. The mixed-view portion (22\%) further suggests that recent studies increasingly attend to task-level generalizability across both egocentric and exocentric videos.

Compared with task-oriented transfer, observation-oriented and action-oriented transfer are egocentric-dominated. Observation-oriented methods have a clear preference for egocentric sources over exocentric counterparts (49\% vs. 40\%). This trend indicates that once the bridge moves from symbolic task understanding to narrowing vision gaps, first-person data become substantially more valuable. This is because robots typically manipulate objects within the field of view of their own cameras, and egocentric videos thus provide a better viewpoint match for capturing fine-grained hand motion patterns and local object appearance. Nonetheless, exocentric videos remain highly competitive in this category because their stable viewpoint simplifies cross-scene transformation and progress estimation. The egocentric tendency becomes even more pronounced in action-oriented transfer, where egocentric and exocentric sources account for 52\% and 41\% of the reviewed works, respectively. This stronger preference is structurally expected. Once the bridge mechanism approaches executable action, it is more important to preserve interaction geometry and temporally precise manipulation cues beyond only aligning visual semantics. Egocentric videos naturally outperform exocentric videos in this category, because they provide closer observations of hand-object interactions and manipulation timings. These factors are essential for learning affordances, latent actions, and transferable action priors. This is particularly evident in dexterous manipulation, bimanual coordination, and contact-rich tasks, where subtle changes in grasp pose, wrist motion, or object pose can directly determine task success. In addition, the very small fraction of mixed-view action transfer indicates that jointly resolving the viewpoint gap and the embodiment gap becomes substantially harder in action-oriented transfer. Therefore, most methods still attend to one dominant viewpoint. Overall, the stronger egocentric dominance in action-oriented transfer suggests that as the bridge mechanism moves closer to robot control, viewpoint alignment becomes increasingly tied to physical interaction fidelity rather than only semantic awareness.

The popularity trend in Fig.~\ref{fig:yearly_grouped_bar_all} further clarifies how the viewpoint preferences evolved along the time axis. Before 2022, the reviewed literature was overwhelmingly exocentric. This early trend reflects data availability and methodological convenience. Third-person videos were easier to collect on the Internet and were already common in action recognition benchmarks. They also better matched systems that depended on static cameras. However, starting from 2022, egocentric sources expand rapidly and soon become a major driver of growth. In particular, egocentric usage rises sharply after 2024. This trend may benefit from the emergence of large-scale egocentric datasets~\citep{grauman2022ego4d,liu2022hoi4d,wang2023holoassist,banerjee2024hot3d,hoque2025egodex} and wearable devices (e.g., Vision Pro, Aria), along with rapid improvements in egocentric HOI analysis techniques~\citep{labbe2022megapose,wen2023bundlesdf,wen2024foundationpose,pavlakos2024reconstructing,karaev2024cotracker,karaev2025cotracker3}. Besides, dexterous manipulation~\citep{shaw2023videodex,wen2025gr,zhang2026unidex} and VLA pretraining~\citep{yuan2025motiontrans,kareer2025emergence,cheang2025gr} both benefit from first-person interaction evidence, further facilitating wider adoption of large-scale egocentric human videos. By contrast, mixed egocentric+exocentric settings appear mainly as a transitional design choice. Their limited long-term prevalence suggests that although multi-view fusion is conceptually appealing, its low-efficiency collection pipeline and practical cost in synchronization and representation alignment still restrict scalability. It is worth noting that egocentric usage significantly dominates the recent LfHV studies in 2026. This highlights the rapid development of egocentric data ecosystems and the increasing innovation in algorithms adapted to egocentric data.

\begin{table*}[t]
\centering
\scriptsize
\setlength{\tabcolsep}{4pt}
\renewcommand{\arraystretch}{1.15}
\begin{tabularx}{\textwidth}{
>{\raggedright\arraybackslash}p{2.8cm}
>{\raggedright\arraybackslash}p{2.6cm}
X
>{\centering\arraybackslash}p{1.6cm}
}
\toprule
Oriented transfer & Robot data requirement & Reference & Percentage \\
\midrule

\multirow{3}{*}{Task-oriented}
& Human Videos + Real-World Robot Demonstration
& \cite{yu2018one,yu2018one_hil,sharma2019third,jang2022bc,xu2023xskill,jain2024vid2robot,clark2025action,lin2025physbrain,li2026act}
& 43\% \\

& Human Videos + Real-World Interaction
& \cite{sermanet2016unsupervised,yang2022learning}
& 10\% \\

& Human Videos Only
& \cite{yang2015robot,nguyen2018translating,pertsch2022cross,chen2024vlmimic,ding2024knowledge,wake2024gpt,wang2024vlm,chen2025fmimic,ma2025egoloc,ye2025watch}
& 48\% \\

\midrule

\multirow{3}{*}{Observation-oriented}
& Human Videos + Real-World Robot Demonstration
& \cite{ma2022vip,nair2022r3m,bhateja2023robotic,dasari2023unbiased,duan2023ar2,ma2023liv,majumdar2023we,radosavovic2023real,wu2023unleashing,cheang2024gr,jain2024vid2robot,li2024ag2manip,zeng2024learning,ci2025h2r,jiang2025rynnvla,kedia2025one,lepert2025masquerade,li2025h2r,li2025mimicdreamer,liu2025immimic,punamiya2025egobridge,song2025mitty,zhang2025generative,zhu2025learning}
& 51\% \\

& Human Videos + Real-World Interaction
& \cite{sermanet2017time,liu2018imitation,schmeckpeper2020reinforcement,smith2020avid,bahl2022human,chang2023look,mendonca2023structured,qian2024contrast,ye2026visual}
& 19\% \\

& Human Videos Only
& \cite{xiong2021learning,sun2022learning,xiao2022masked,xiong2022robotube,zakka2022xirl,liu2024masked,zhu2024vision,goswami2025world,lepert2025phantom,shah2025mimicdroid,sun2025vtao,tang2025trajectory,xiong2025ag2x2,freeman2026warped}
& 30\% \\

\midrule

\multirow{3}{*}{Action-oriented}
& Human Videos + Real-World Robot Demonstration
& \cite{lee2017learning,das2021model,arunachalam2022dexterous,qin2022one,sivakumar2022robotic,bharadhwaj2023towards,gu2023rt,shaw2023videodex,wang2023mimicplay,wen2023any,chen2024igor,srirama2024hrp,ye2024latent,bi2025motus,bjorck2025gr00t,bu2025univla,cai2025n,chen2025graphmimic,chen2025villa,feng2025spatial,jiang2025rynnvla,kareer2025egomimic,kareer2025emergence,kim2025uniskill,li2025latbot,li2025scalable,luo2025being,qiu2025humanoid,ren2025motion,singh2025hand,spiridonov2025generalist,wen2025gr,yang2025ar,yang2025egovla,yang2025tra,yoshida2025developing,yuan2025motiontrans,zhou2025human,bi2026h,gao2026dreamdojo,garrido2026learning,lee2026mvp,luo2026being,luo2026joint,lyu2026lda,zhang2026clap,zhang2026unidex,zheng2026egoscale,zhu2026emma}
& 49\% \\

& Human Videos + Real-World Interaction
& \cite{sieb2020graph,bahl2022human,kannan2023deft,bahety2024screwmimic,chen2025vividex,jonnavittula2025view}
& 6\% \\

& Human Videos Only
& \cite{mandikal2022dexvip,qin2022dexmv,wen2022you,bahl2023affordances,bharadhwaj2023zero,gao2023k,ko2023learning,kumar2023graph,wang2023robot,ye2023learning,bharadhwaj2024track2act,heppert2024ditto,ju2024robo,kuang2024ram,li2024okami,xu2024flow,yuan2024general,chen2025vidbot,dan2025x,haldar2025point,heidinger20252handedafforder,hsieh2025dexman,hsu2025spot,li2025novaflow,liu2025egozero,lum2025crossing,ma2025egoloc,ma2025uni,papagiannis2025r,park2025demodiffusion,shan2025slot,shi2025zeromimic,singh2025deep,tang2025functo,tang2025mimicfunc,werby2025articulated,yin2025object,zhang2025actron3d,zhang2025zero,zhao2025dexh2r,zhou2025you,chen2026dexterous,soraki2026objectforesight,wang2026paws}
& 44\% \\

\bottomrule
\end{tabularx}
\caption{Statistics of dependence on real-world robot data in LfHV studies.}
\label{tab:oriented_robot_data_requirement}
\vspace{-0.2cm}
\end{table*}

\textit{(b) Dependence on robot data:} 
Considering human video sources are fundamentally used to mitigate the data collection cost of real-world robot data, we further present how completely each transfer mechanism resolves the embodiment gap before the robot starts learning or executing. Here, we restrict the statistics to works that have been deployed on real-world robot experiments in their original papers. Specifically, Tab.~\ref{tab:oriented_robot_data_requirement} presents the extent of dependence on real-world robot data across different transfer families. As can be noted, task-oriented transfer exhibits the weakest dependence on real-world robot data, where Human Videos Only accounts for 48\%. As the task-level abstractions are relatively far from low-level execution, many task-oriented methods remain effective without additional real-robot demonstrations, especially when the transferred output is consumed by symbolic planning, VLM-based reasoning, or program generation modules rather than a learned visuomotor controller~\citep{ding2024knowledge,chen2024vlmimic,chen2025fmimic,wake2024gpt,wang2024vlm,ye2025watch}. This trend also indicates that task-oriented transfer methods often stop at the level of \emph{what to do}, not \emph{how a specific robot should physically do it}. Accordingly, task-oriented transfer can reduce the need for robot data most effectively when the main bottleneck lies in semantic task specification rather than embodiment-specific control grounding. When it comes to fine-grained manipulations in complex environments, real-world robot data are still required to learn a reasonable visuomotor policy.

In contrast, observation-oriented and action-oriented transfer both exhibit stronger dependence on real-world robot data (robot demonstrations and interactions). For observation-oriented methods, human videos mainly help narrow the visual domain gap by providing translated observations or transferable visual representations. However, they usually do not by themselves determine how those observations should be mapped into robot actions. Consequently, many methods still require in-domain robot demonstrations to ground the learned observation space into executable control~\citep{li2024ag2manip,li2025h2r,li2025mimicdreamer,lepert2025masquerade,jiang2025rynnvla}. Action-oriented transfer is closer to execution, but most relevant works still remain robot-data-dependent when human action patterns captured by affordance-based pretraining/co-training~\citep{srirama2024hrp,luo2025being,kareer2025egomimic,zhu2026emma} and latent actions~\citep{ye2024latent,chen2024igor,chen2025villa,li2025latbot}, must still be grounded to robot kinematics by robot action supervision or interactions with real-world environments. Nonetheless, action-oriented transfer contains the largest number of LfHV works that enable real-world robot deployment using only human video data. This is because action-oriented methods are more likely to produce relatively executable intermediate interfaces, such as trajectories and contact regions, which can be directly retargeted or integrated with generic off-the-shelf policies (e.g., AnyGrasp~\citep{fang2023anygrasp}, KAT~\citep{di2024keypoint}) for robot control without additional real-world robot demonstrations~\citep{bahl2023affordances,bharadhwaj2024track2act,kuang2024ram,papagiannis2025r,hsu2025spot,zhang2025actron3d}.

\begin{figure}[t]
  \centering
  \includegraphics[width=1\linewidth]{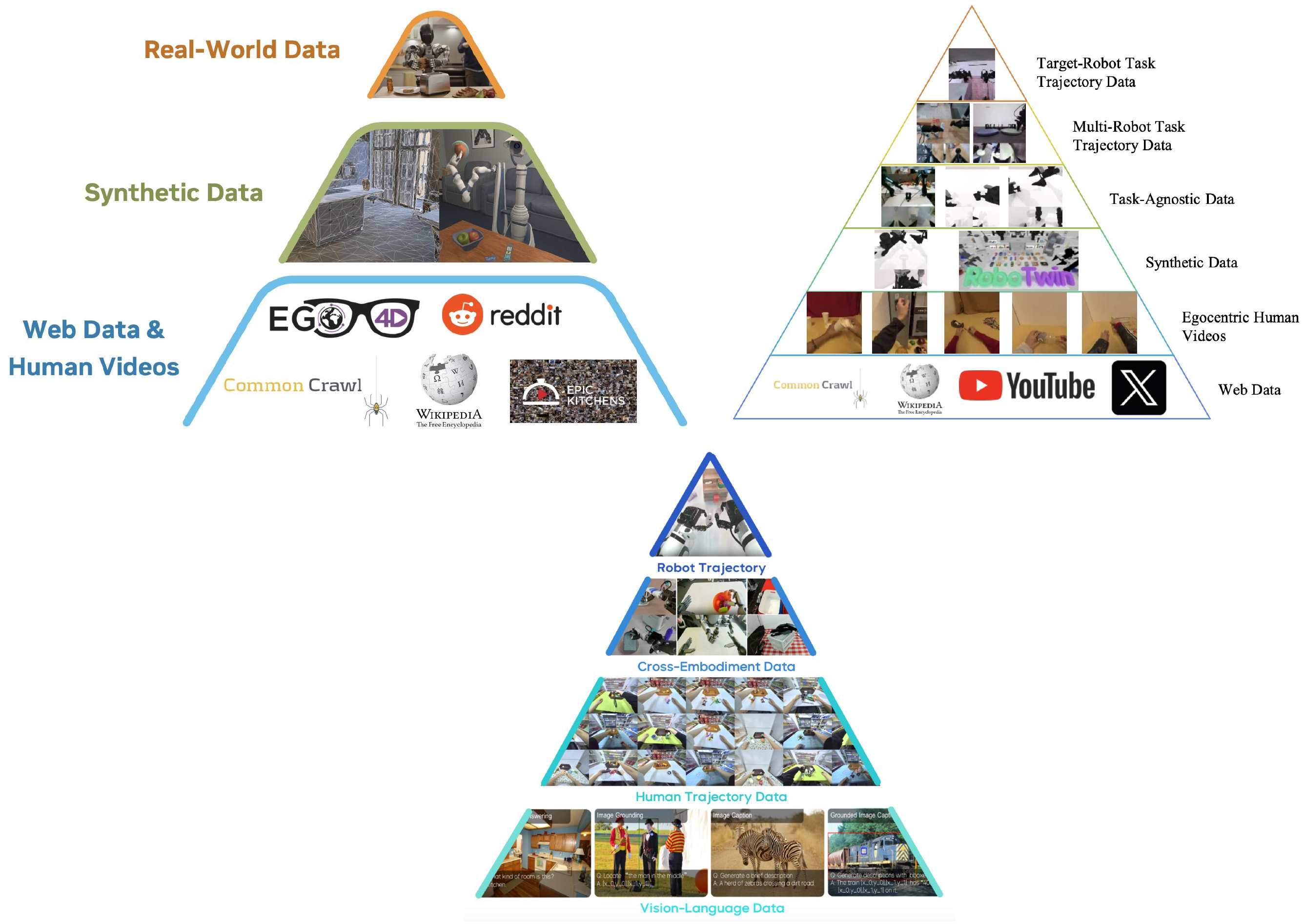}
  \caption{Data pyramids defined by~\cite{bjorck2025gr00t,bi2025motus,wen2025gr}, respectively.}
  \label{fig:data_pyramid}
  \vspace{-0.3cm}
\end{figure}

As can be noted in recent studies, the LfHV community has not fully eliminated the need for real-world robot data. Although a meaningful subset of works support real-world deployment using only human videos, their robustness in complex target scenarios, especially under out-of-distribution conditions, still requires further improvement. In most existing works, real-robot demonstrations or interactions are still needed to generate robot-specific policies with feasible kinematics and closed-loop correction. Thus, the present evidence suggests that human videos can substantially serve as one layer of the embodied data pyramid (see Fig.~\ref{fig:data_pyramid}), but cannot yet completely replace real robot data in general for the development of generalist agents.

\subsubsection{Across Learning Paradigms}

In addition to data configurations, we further explore the differences in learning paradigms of these transfer families. The key distinction between learning paradigms is whether human videos provide a signal that can be consumed directly by a policy, or a signal that shapes robot learning indirectly through reward or exploration priors. The former naturally aligns with imitation learning (IL) and VLA-style supervised post-training, whereas the latter more aligns with reinforcement learning (RL) or exploration policies. Beyond these two dominant paradigms, a substantial subset of works convert human videos into programs, affordances, or other structured intermediates that are deployed through retargeting, optimization, or analytic controllers rather than learned end-to-end policies like IL and RL.

We observe that task-oriented transfer is primarily dominated by IL-style formulations. This is because task structures and task intents usually serve as high-level prompts, plans, or subgoal specifications that a downstream policy follows, rather than dense objectives that can be optimized directly. Representative examples include BC-Z~\citep{jang2022bc} and Vid2Robot~\citep{jain2024vid2robot}. In these works, human videos condition robot policies through task embeddings or prompt videos, while the final behavior is still learned by behavior cloning. A relatively smaller subset falls outside standard IL or RL, because the transferred output is an executable semantic program or task plan rather than a policy supervision signal itself~\citep{ding2024knowledge,wake2024gpt,wang2024vlm,ye2025watch}. RL appears only occasionally in this category, typically when task progress is explicitly converted into reward signals~\citep{sermanet2016unsupervised}.

Compared to task-oriented transfer, observation-oriented transfer is more paradigm-flexible. Its central role is to reduce the human-robot visual observation gap. Therefore, the same bridge can naturally support different downstream paradigms depending on how the visual representation is exploited. When transformed videos or learned visual features are treated as robot-aligned policy inputs, the resulting pipeline is typically IL-centric~\citep{li2024ag2manip,lepert2025masquerade,li2025h2r,li2025mimicdreamer}. In contrast, when the observation bridge is used to define goal similarity, progress measures, or success scores, it becomes more naturally compatible with RL paradigms~\citep{smith2020avid,zakka2022xirl}. Moreover, observation-oriented transfer also exhibits a notable direction centered on exploration policies, where human videos mainly provide priors for exploratory decision making~\citep{chang2023look,mendonca2023structured,goswami2025world}. 

\begin{table*}[t]
\centering
\scriptsize
\setlength{\tabcolsep}{3pt}
\renewcommand{\arraystretch}{1.15}
\begin{tabularx}{\textwidth}{
>{\raggedright\arraybackslash}p{2.3cm}
X
X
X
}
\toprule
Oriented transfer & Why it works & When it tends to fail & Prefer this route when \\
\midrule
Task-oriented transfer &
It transfers information at the most embodiment-agnostic level. Task structures, intents, and programs can bypass the low-level human-robot action gap and guide existing robot skills or planners~\citep{ding2024knowledge,wake2024gpt,wang2024vlm,ye2025watch}. &
It fails when high-level plans are underspecified for contact-rich execution, when VLM-generated steps hallucinate or miss physical constraints, or when the robot lacks the primitive skills needed to ground the plan. &
The task is long-horizon or semantically complex, the target robot already has a usable skill library or controller, and the main problem is deciding \textit{what to do next} rather than learning low-level control. \\
\midrule
Observation-oriented transfer &
It reduces the perceptual gap between human videos and robot observations by editing embodiment appearance, predicting robot-like views, or learning shared visual representations~\citep{nair2022r3m,li2024ag2manip,lepert2025masquerade,li2025h2r}. &
It fails when visual alignment is only appearance-level: generated videos may contain artifacts, shared embeddings may ignore contact-relevant dynamics, and visually similar states may still require different robot actions. &
The downstream policy or controller is already available or can be trained with limited robot data, and the main problem is improving visual generalization, data augmentation, or goal/reward matching. \\
\midrule
Affordance-based action transfer &
It exposes explicit geometric cues such as contact regions, hand trajectories, and object poses. These cues are inspectable and can be used for reward shaping, policy conditioning, retargeting, or direct execution~\citep{bahl2023affordances,bharadhwaj2024track2act,kuang2024ram,chen2025vidbot,zhang2025actron3d}. &
It fails when HOI parsing is unreliable because of occlusion, camera motion, object reconstruction errors, or severe morphology mismatch. It also struggles when force, compliance, or contact stability cannot be inferred from vision alone. &
The task requires spatially precise manipulation, one-shot or few-shot transfer, object-centric motion, contact localization, or a human-video-only deployment pipeline with analytic retargeting or a generic controller. \\
\midrule
Latent-action transfer &
It learns compact action-related abstractions from large-scale videos without requiring explicit geometric annotation. This makes it scalable for VLA pretraining and heterogeneous in-the-wild data~\citep{ye2024latent,chen2024igor,chen2025villa,luo2025being,yang2025egovla}. &
It fails when the latent code captures camera motion, background change, or other nuisance dynamics instead of controllable actions. Even well-learned latent actions still need robot grounding before they become executable. &
The goal is large-scale policy pretraining, cross-dataset absorption of human behavior, or building a generalist VLA backbone where robot demonstrations are available for downstream grounding. \\
\bottomrule
\end{tabularx}
\caption{Practical guidelines for selecting LfHV transfer routes.}
\label{tab:route_selection_guidelines}
\end{table*}

Action-oriented transfer spans the widest range of paradigms. A large portion of this family still follows imitation learning paradigms. This is because once human videos provide sufficiently action-relevant cues, they can be more directly used as action supervision for policy learning than rewards for interaction-based optimization. This tendency is particularly evident in affordance-based backbone pretraining and co-training pipelines~\citep{wang2023mimicplay,kareer2025egomimic,yang2025egovla,luo2025being}. For latent-action-based methods, human videos are converted into compact action abstractions, and thus additional IL-style post-training with robot demonstrations is always required~\citep{ye2024latent,chen2024igor,chen2025villa}. In contrast, RL-based and exploration-centric action-oriented methods are mainly used in settings where affordances directly specify the optimization objective, or where large embodiment mismatch and complex interaction dynamics make direct imitation unreliable~\citep{das2021model,sieb2020graph,kumar2023graph,lum2025crossing}. Beyond IL and RL paradigms, action-oriented transfer also contains the largest number of direct execution methods, such as zero-shot retargeting, trajectory optimization, and affordance-guided deployment~\citep{bahl2023affordances,kuang2024ram,chen2025vidbot,zhang2025actron3d,shi2025zeromimic}. This is because affordances as policy produce physically grounded intermediate representations that can be directly transformed into robot actions without passing through a full policy learning stage.

As can be seen, with the improvement of observation alignment, affordance extraction, and latent action modeling, more recent methods can convert human videos into robot-aligned observations, pseudo-actions, and executable intermediates. This makes supervised policy learning and VLA-style post-training substantially more practical at scale. Nonetheless, RL remains indispensable when the embodiment gap is large, interaction dynamics are difficult to model offline, and online refinement is required for physical feasibility. Therefore, we argue that an important future direction for LfHV lies in combining IL-style pretraining from large-scale human videos with RL-based online improvement in target task environments. This direction unifies scalable prior acquisition with task-specific physical adaptation.

\subsection{Route Selection: Effectiveness, Failure Modes, and Practical Guidelines}
\label{sec:route_selection}

The preceding sections achieve intra-family analysis and cross-family comparison of LfHV methods. A practical question remains: \textit{which bridge should be selected for a new robot learning problem?} The answer depends less on the nominal model architecture than on the bottleneck that the human video is expected to resolve. If the main bottleneck is task specification, a high-level semantic bridge is usually sufficient. If the bottleneck is a visual domain mismatch, observation-level alignment becomes more suitable. If the bottleneck is executable motion, the bridge must move closer to actions through affordances and latent actions. Tab.~\ref{tab:route_selection_guidelines} summarizes this decision logic.

The effectiveness of a transfer route is therefore governed by its distance from robot actions. Task-oriented transfer is effective because semantic task knowledge is highly transferable across embodiments, but its distance from robot control makes it heavily dependent on downstream grounding. Observation-oriented transfer is effective when the policy failure is caused by visual domain shift, but it cannot by itself solve the action-missing problem. Affordance-based transfer becomes effective when the extracted geometric cues are close enough to executable motion and remain reliable under the target viewpoint and interaction geometry. Latent-action transfer is effective at scale because it replaces expensive explicit annotation with self-supervised motion abstraction. However, its implicit nature makes controllability and physical grounding harder to verify.

This perspective also clarifies the major failure modes in the literature. Methods far from actions usually fail through \textit{under-grounding}: the robot understands the task but cannot execute it precisely. Methods close to actions usually fail through \textit{mis-grounding}: the extracted trajectory, contact cue, or latent action appears plausible in the human video but violates the target robot's kinematics, control frequency, collision constraints, or contact dynamics. Observation-oriented methods occupy an intermediate position and often fail through \textit{false visual equivalence}, where aligned images or embeddings do not imply aligned actions. 

\textbf{For practical method design, a conservative strategy is to select the highest-level bridge that still resolves the limiting bottleneck.} If a robot already has reliable low-level manipulation primitives, task-oriented transfer can provide a data-efficient route for new long-horizon tasks. If the target task is clear and the controller exists but the visual domain gap is large, observation-oriented transfer is preferable. If the target behavior depends on where and how interactions occur, explicit affordances should be prioritized because they expose checkable and explicit intermediate states. If the goal is to train a scalable generalist policy from diverse videos, latent actions or affordance-supervised VLA pretraining become more suitable. Notably, they should be paired with robot demonstrations or interaction-based refinement for embodiment grounding. In contact-rich or high-risk settings, the most robust pipeline can be conducted in a hybrid manner. That is, task-level plans provide temporal structure, observation-level modules improve perception, action-level affordances or latent actions propose motion priors, and RL or closed-loop control corrects residual physical mismatch.

\section{Data Foundations} \label{sec:data_foundations}

The acquisition of human video data is a prerequisite for LfHV research. In this section, we first introduce the sources of existing human video datasets, and then review current generative techniques for synthesizing human videos from scratch.

\subsection{Open-Source Datasets}\label{sec:data_sources}

Although a large volume of human videos can be obtained directly through web crawling, researchers prefer organizing online videos in a targeted manner or recording scripted videos in controlled laboratory and everyday environments. These efforts have yielded a rich ecosystem of open-source human video datasets that facilitate the advancement of LfHV techniques. 
In this work, we systematically compile a diverse collection of these datasets. Compared to previous surveys and dataset papers~\citep{liu2022hoi4d,grauman2022ego4d,mccarthy2025towards,eze2025learning,hoque2025egodex,banerjee2024hot3d,feng2026human}, our dataset statistics present the following key features:
\begin{itemize}[leftmargin=1em]
    \setlength{\parskip}{0pt}
    \item \textbf{Broad yet well-justified dataset coverage:} We provide an extensive review of 50 human video datasets, encompassing those widely adopted or referenced in LfHV research from 2014 to 2026, as well as emerging ones with significant potential for future applications. To the best of our knowledge, this survey compares the largest number of human video datasets in the literature.
    \item \textbf{Comprehensive report of dataset attributes:} We present a comprehensive analysis of dataset attributes, such as frame and participant counts, geographic coverage, recording modalities, collection types, and annotation features. These statistics can help future LfHV studies select datasets that are most suitable for their specific technical requirements.
    \item \textbf{Temporal view of dataset development trends:} We provide a chronological analysis of dataset evolution over the past years and discuss development trends and future directions. This can help researchers better understand how dataset design choices have evolved over time and identify promising directions for constructing their own datasets.
    \item \textbf{Additional analysis of dataset prevalence:} Finally, we additionally analyze dataset usage frequency by summarizing which LfHV studies have used each dataset, thus revealing the popularity of these datasets across different LfHV categories. This may help future dataset builders identify which types of datasets are most suitable for their research goals and most likely to be broadly adopted in the LfHV literature.
\end{itemize}

In Tab.~\ref{tab:human_video_datasets}, we summarize the following attributes of 50 human video datasets:
\begin{itemize}[leftmargin=1em]
    \setlength{\parskip}{0pt}
    \renewcommand\labelitemi{\(\circ\)}
    \item \textbf{Year:} the publication year of the dataset paper. We sort the datasets according to the time when they were first publicly released.
    \item \textbf{Frames:} the total number of image frames contained in the dataset videos.
    \item \textbf{Sequences:} the number of untrimmed videos, which are typically raw recorded footage before further processing, with an average length of more than \textit{one minute}.
    \item \textbf{Clips:} the number of segmented video clips or action/trajectory segments, typically with an average length of less than \textit{one minute}.
    \item \textbf{Hours:} the total duration of all videos in the dataset.
    \item \textbf{Participants:} the number of subjects involved in collecting the dataset videos.
    \item \textbf{Geographic coverage:} the geographic scope or category coverage of the dataset. Since these datasets use different terms to describe collection sites like \textit{locations}, \textit{scenarios}, \textit{environments}, \textit{cities}, and \textit{categories}, we directly report the original terms used in each dataset paper.
    \item \textbf{View:} the types of viewpoints included in the dataset. \textit{Ego}: egocentric, first-person. \textit{Exo}: exocentric, third-person.
    \item \textbf{Camera:} the camera devices used for video recording, excluding functional cameras used for motion capture.
    \item \textbf{Hand type:} whether the video recording attends to dual hands or only a single hand.
    \item \textbf{2D hand det.:} whether the dataset provides 2D hand bounding box or segmentation mask annotations.
    \item \textbf{Hand pose:} whether the dataset provides 6-DOF hand pose annotations.
    \item \textbf{Hand joint:} whether the dataset provides 3D hand joint position labels.
    \item \textbf{Pose annot.:} the annotation protocol used for hand pose labeling. \textit{Mocap}: directly captured by motion capture systems. \textit{Device tracking}: directly recorded by built-in tracking algorithms of the headsets. \textit{RGB}: estimated from RGB images only. \textit{RGB(-D) + opt.}: estimated from RGB or RGB-D images and refined by optimization.
    \item \textbf{Depth:} whether raw depth observations are provided. Monocular depth estimation is not counted.
    \item \textbf{Gaze:} whether gaze information is collected.
    \item \textbf{Audio:} whether audio signals are collected.
    \item \textbf{Language desc.:} whether language descriptions or narrations are provided. Short action labels in a \textit{verb + noun} form are not counted.
    \item \textbf{Source:} the data sources corresponding to the collection protocols. \textit{Curated}: videos are collected under an organized acquisition setup with predefined tasks, devices, or recording procedures. \textit{In-the-wild}: videos are gathered from naturally occurring or unconstrained settings, such as web videos or unscripted daily recordings, without strict collection control. \textit{Mixed}: videos are compiled from multiple open-source datasets, including both curated and in-the-wild sources.
\end{itemize}

Some entries in Tab.~\ref{tab:human_video_datasets} are estimated from the available statistics. For example, the total duration in hours may be computed by multiplying the number of clips by the average clip length. The symbol ``-'' indicates that the corresponding information is unavailable. We emphasize the acquisition and annotation of human hand motion because effective hand-object interaction is the key factor that enables human video data to benefit robot learning. By examining the relationship between the properties of human video datasets and their publication time, we observe the following temporal trends:
\begin{itemize}[leftmargin=1em]
    \setlength{\parskip}{0pt}
    \item \textbf{Consistently large video duration.} The total recording time of human video datasets is generally very large, often reaching hundreds or even thousands of hours. This trend highlights the convenience of collecting human video data and further demonstrates its scalability compared with conventional robot demonstration data. In addition, in-the-wild datasets such as HowTo100M~\citep{miech2019howto100m}, EPIC-KITCHENS-100~\citep{damen2020rescaling}, Panda-70M~\citep{chen2024panda}, and Action100M~\citep{chen2026action100m} basically have a larger scale than curated counterparts, benefiting from more convenient collection protocols such as web crawling or unscripted activity recording.
    \item \textbf{A decline in purely in-the-wild datasets.} Although in-the-wild datasets are easier to collect, they have appeared less frequently in recent years. This suggests that the community is increasingly prioritizing data quality, annotation reliability, and controllability, despite sacrificing diversity in unconstrained real-world scenarios. For example, while EgoDex~\citep{hoque2025egodex}, OakInk2~\citep{zhan2024oakink2}, TACO~\citep{liu2024taco}, and HOT3D~\citep{banerjee2024hot3d} all include $<5$ collection scenarios, many researchers still prefer to harness them for extracting high-quality human motion that can be transferred to robotic manipulation.
    \item \textbf{From standalone datasets to mixed compositions.} Earlier datasets were typically introduced as independent resources with relatively fixed acquisition protocols. However, there has been a clear trend toward mixed compositions recently, where multiple datasets are combined to form more comprehensive training resources, such as UniHand series~\citep{luo2025being,luo2026joint,luo2026being}. This reduces the cost of human video collection while significantly improving data diversity in different dimensions like viewpoints, modalities, and scenarios.
    \item \textbf{Increasing emphasis on fine-grained hand annotation.} Recent datasets~\citep{luo2025being,hoque2025egodex,qiu2025humanoid,luo2026being} pay more attention to detailed hand pose and joint annotations. This trend reflects the growing recognition that accurate hand motion modeling is critical for transferring HOI knowledge to robot manipulation. Besides, with the development of smart glasses (e.g., Vision Pro and Aria), hand poses can be efficiently captured with built-in tracking algorithms, without requiring additional annotation pipelines~\citep{hoque2025egodex,chavan2025indego,qiu2025humanoid}.
    \item \textbf{More task-related language descriptions.} Text annotation has also become increasingly complete over time. Compared with earlier datasets~\citep{kuehne2014language,caba2015activitynet,goyal2017something} that only provide coarse action labels in a \textit{verb + noun} form, newer datasets~\citep{chavan2025indego,zhao2025taste,chen2026action100m} are more likely to include richer language descriptions and narrations for human videos, which are more suitable for language-driven robot learning.
\end{itemize}

\clearpage
\begin{sidewaystable}[t]
\vspace{9cm}
\centering
\tiny
\setlength{\tabcolsep}{2.2pt}
\renewcommand{\arraystretch}{1.2}
\caption{Statistics of open-source human video datasets.}
\begin{tabular}{cccccccccccccccccccc}
\toprule
Dataset & Year & Frames & Sequences & Clips & Hours & Participants & Geographic coverage & View & Camera & Hand type & 2D hand det. & Hand pose & Hand joint & Pose annot. & Depth & Gaze & Audio & Language desc. & Source \\
\midrule
Breakfast~\citep{kuehne2014language} & 2014 & 4M & 1,712 & 11,267 & 77 & 52 & 18 kitchens & Exo & \makecell[c]{Prosilica GE680C,\\ Bumblebee} & Dual & \xmark & \xmark & \xmark & - & \xmark & \xmark & \xmark & \xmark & In-the-wild \\
ActivityNet~\citep{caba2015activitynet} & 2015 & - & 27,801 & - & 849 & - & - & Exo & - & Dual & \xmark & \xmark & \xmark & - & \xmark & \xmark & \xmark & \xmark & In-the-wild \\
EgoHands~\citep{bambach2015lending} & 2015 & 130,000 & 48 & - & 1.2 & 4 & 3 locations & Ego & Google Glass & Dual & \cmark & \xmark & \xmark & - & \xmark & \xmark & \xmark & \xmark & Curated \\
Charades~\citep{sigurdsson2016hollywood} & 2016 & 8.6M & - & 9,848 & 82.3 & 267 & 15 categories & Exo & - & Dual & \xmark & \xmark & \xmark & - & \xmark & \xmark & \xmark & \cmark & Curated \\
FPHA~\citep{garcia2017first} & 2017 & 105,459 & - & 1,175 & - & 6 & 3 scenarios & Ego & Intel RealSense SR300 & Single & \xmark & \cmark & \cmark & Mocap & \cmark & \xmark & \xmark & \xmark & Curated \\
Something-Something~\citep{goyal2017something} & 2017 & - & - & 108,499 & 121.5 & 1,133 & - & Exo & - & Dual & \xmark & \xmark & \xmark & - & \xmark & \xmark & \xmark & \xmark & Curated \\
YouCook2~\citep{zhou2017towards} & 2017 & - & 2,000 & 15.4K & 176 & - & - & Exo & - & Dual & \xmark & \xmark & \xmark & - & \xmark & \xmark & \xmark & \cmark & In-the-wild \\
VLOG~\citep{fouhey2017lifestyle} & 2017 & 37.2M & - & 114K & 344 & 10.7K & 6 categories & Exo & - & Dual & \cmark & \xmark & \xmark & - & \xmark & \xmark & \xmark & \xmark & In-the-wild \\
EPIC-KITCHENS~\citep{damen2018scaling} & 2018 & 11.5M & 432 & 39,596 & 55 & 32 & 32 envs & Ego & GoPro & Dual & \xmark & \xmark & \xmark & - & \xmark & \xmark & \cmark & \cmark & In-the-wild \\
EGTEA Gaze+~\citep{li2018eye} & 2018 & 2.5M & 86 & 10,325 & 28 & 32 & 1 scenario & Ego & SMI & Dual & \cmark & \xmark & \xmark & - & \xmark & \cmark & \cmark & \xmark & Curated \\
HowTo100M~\citep{miech2019howto100m} & 2019 & - & 1.22M & 136M & 134,472 & - & - & Ego + Exo & - & Dual & \xmark & \xmark & \xmark & - & \xmark & \xmark & \cmark & \cmark & In-the-wild \\
FreiHAND~\citep{zimmermann2019freihand} & 2019 & 37K & - & - & - & 32 & 1 scenario & Exo & \makecell[c]{Basler acA800-510uc,\\ Basler acA1300-200uc} & Single & \cmark & \cmark & \cmark & RGB + opt. & \cmark & \xmark & \xmark & \xmark & Curated \\
100DOH~\citep{shan2020understanding} & 2020 & 100K & 27.3K & - & - & - & - & Ego + Exo & - & Dual & \cmark & \xmark & \xmark & - & \xmark & \xmark & \xmark & \xmark & In-the-wild \\
EPIC-KITCHENS-100~\citep{damen2020rescaling} & 2020 & 20M & 700 & 89,977 & 100 & 37 & 45 envs & Ego & GoPro & Dual & \xmark & \xmark & \xmark & - & \xmark & \xmark & \cmark & \cmark & In-the-wild \\
Kinetics-700~\citep{smaira2020short} & 2020 & - & - & 650,317 & 1,806 & - & 6 continents & Exo & - & Dual & \xmark & \xmark & \xmark & - & \xmark & \xmark & \xmark & \xmark & In-the-wild \\
MOW~\citep{cao2020reconstructing} & 2020 & 500 & - & - & - & - & - & Exo & - & Single & \cmark & \cmark & \cmark & RGB + mocap + opt. & \xmark & \xmark & \xmark & \xmark & In-the-wild \\
DexYCB~\citep{chao2021dexycb} & 2021 & 582K & - & 1,000 & - & 10 & 1 scenario & Exo & Intel RealSense D415 & Single & \xmark & \cmark & \cmark & RGB-D + opt. & \cmark & \xmark & \xmark & \xmark & Curated \\
H2O~\citep{kwon2021h2o} & 2021 & 571,645 & - & - & - & 4 & 3 envs & Ego & Azure Kinect & Dual & \xmark & \cmark & \cmark & RGB-D + opt. & \cmark & \xmark & \xmark & \xmark & Curated \\
Ego4D~\citep{grauman2022ego4d} & 2021 & 19.2M & 83,647 & - & 3,670 & 931 & \makecell[c]{74 locations /\\ 74 cities} & Ego & \makecell[c]{GoPro, Vuzix Blade,\\ Pupil Labs, ZShades,\\ ORDRO EP6, iVue Rincon \\ 1080, and Weeview} & Dual & \cmark & \xmark & \xmark & - & \xmark & \cmark & \cmark & \cmark & In-the-wild \\
Assembly101~\citep{sener2022assembly101} & 2022 & 111M & 4,321 & 82K & 513 & 53 & 1 scenario & Ego + Exo & - & Dual & \xmark & \cmark & \cmark & RGB + opt. & \xmark & \xmark & \xmark & \xmark & Curated \\
EgoPAT3D~\citep{li2022egocentric} & 2022 & 1M & 150 & 15,000 & 10 & 2 & 15 envs & Ego & Azure Kinect DK & Single & \xmark & \xmark & \xmark & - & \cmark & \xmark & \cmark & \xmark & Curated \\
AGD20K~\citep{luo2022grounded} & 2022 & 26,117 & - & - & - & - & - & Ego + Exo & - & Dual & \xmark & \xmark & \xmark & - & \xmark & \xmark & \xmark & \xmark & Mixed \\
HOI4D~\citep{liu2022hoi4d} & 2022 & 2.4M & - & 4,000 & 7.6 & 4 & 610 rooms & Ego & \makecell[c]{Intel RealSense D455,\\ Kinect v2} & Single & \cmark & \cmark & \cmark & RGB-D + opt. & \cmark & \xmark & \xmark & \xmark & Curated \\
OakInk~\citep{yang2022oakink} & 2022 & 230,064 & - & - & - & 12 & 1 scenario & Exo & Intel RealSense D435 & Single & \xmark & \cmark & \cmark & RGB-D + opt. & \cmark & \xmark & \xmark & \xmark & Curated \\
EgoHOS~\citep{zhang2022fine} & 2022 & 11,243 & 1,000 & - & - & - & - & Ego & GoPro & Dual & \cmark & \xmark & \xmark & - & \xmark & \xmark & \xmark & \xmark & Mixed \\
ARCTIC~\citep{fan2023arctic} & 2023 & 2.1M & 339 & - & 2.3 & 10 & 1 scenario & Ego + Exo & - & Dual & \xmark & \cmark & \cmark & Mocap & \xmark & \xmark & \xmark & \xmark & Curated \\
RH20T-Human~\citep{fang2023rh20t} & 2023 & 10M & 110K & - & 100 & - & - & Ego + Exo & - & Dual & \xmark & \xmark & \xmark & - & \cmark & \xmark & \cmark & \cmark & Curated \\
HoloAssist~\citep{wang2023holoassist} & 2023 & 17.1M & 2,221 & - & 166 & 222 & - & Ego & HoloLens 2 & Dual & \xmark & \cmark & \cmark & Device tracking & \cmark & \cmark & \cmark & \cmark & Curated \\
Ego-Exo4D~\citep{grauman2023ego} & 2023 & - & 5,035 & - & 1,286 & 740 & \makecell[c]{123 scenes /\\ 13 cities} & Ego + Exo & \makecell[c]{Aria,\\ GoPro} & Dual & \cmark & \cmark & \cmark & RGB + opt. & \cmark & \cmark & \cmark & \cmark & Curated \\
CaptainCook4D~\citep{peddi2024captaincook4d} & 2023 & - & 384 & - & 94.5 & 8 & 10 kitchens & Ego & GoPro, Hololens 2 & Dual & \xmark & \xmark & \xmark & - & \cmark & \xmark & \cmark & \cmark & Curated \\
TACO~\citep{liu2024taco} & 2024 & 5.2M & 2.5K & - & 3.2 & 14 & 1 scenario & Ego + Exo & FLIR, Realsense L515 & Dual & \cmark & \cmark & \cmark & RGB + mocap + opt. & \cmark & \xmark & \xmark & \xmark & Curated \\
Panda-70M~\citep{chen2024panda} & 2024 & - & 3.8M & 70.8M & 166.8K & - & - & Ego + Exo & - & Dual & \xmark & \xmark & \xmark & - & \xmark & \xmark & \xmark & \cmark & In-the-wild \\
OakInk2~\citep{zhan2024oakink2} & 2024 & 4.01M & 627 & - & 6.5 & 9 & 4 scenarios & Ego + Exo & - & Dual & \xmark & \cmark & \cmark & Mocap & \xmark & \xmark & \xmark & \xmark & Curated \\
HO-Cap~\citep{wang2024ho} & 2024 & 656K & - & 64 & - & 9 & 1 scenario & Ego + Exo & \makecell[c]{HoloLens, Intel RealSense \\D455, Azure Kinect} & Dual & \cmark & \cmark & \cmark & RGB-D + opt. & \cmark & \xmark & \xmark & \xmark & Curated \\
HOT3D~\citep{banerjee2024hot3d} & 2024 & 3.7M & 425 & 3,832 & 13.9 & 19 & 4 scenarios & Ego & Aria, Quest3 & Dual & \xmark & \cmark & \cmark & Mocap & \xmark & \cmark & \xmark & \xmark & Curated \\
Nymeria~\citep{ma2024nymeria} & 2024 & 201.2M & 1,200 & - & 300 & 264 & \makecell[c]{20 scenarios /\\ 50 locations} & Ego + Exo & Aria & Dual & \xmark & \cmark & \cmark & Mocap & \xmark & \cmark & \cmark & \cmark & In-the-wild \\
EgoVid-5M~\citep{wang2024egovid} & 2024 & 600M & - & 5M & - & - & 5 categories & Ego & - & Dual & \xmark & \xmark & \xmark & - & \xmark & \xmark & \xmark & \cmark & In-the-wild \\
Egocentric-10k~\citep{buildaiegocentric10k2025} & 2025 & 1.08B & 192,900 & - & 10,000 & 2,138 & 85 factories & Ego & Build AI Gen 1 & Dual & \xmark & \xmark & \xmark & - & \xmark & \xmark & \xmark & \xmark & Curated \\
Egocentric-100k~\citep{buildaiegocentric100k2025} & 2025 & 10.8B & 2,010,759 & - & 100,405 & 14,228 & 238 factories & Ego & Build AI Gen 1 & Dual & \xmark & \xmark & \xmark & - & \xmark & \xmark & \xmark & \xmark & Curated \\
HD-EPIC~\citep{perrett2025hd} & 2025 & 4.46M & 156 & 59,454 & 41.3 & 9 & 9 kitchens & Ego & Aria & Dual & \cmark & \xmark & \xmark & - & \xmark & \cmark & \cmark & \cmark & In-the-wild \\
PH$^2$D~\citep{qiu2025humanoid} & 2025 & 3.02M & - & 26,824 & - & - & - & Ego & \makecell[c]{Vision Pro, Quest 3,\\ ZED Mini Stereo} & Dual & \xmark & \cmark & \cmark & Device tracking & \cmark & \xmark & \xmark & \cmark & Curated \\
TASTE-Rob~\citep{zhao2025taste} & 2025 & 9M & - & 100,856 & 130 & - & 6 scenarios & Ego & - & Dual & \xmark & \cmark & \cmark & RGB & \xmark & \xmark & \xmark & \cmark & Curated \\
EgoDex~\citep{hoque2025egodex} & 2025 & 90M & - & 338K & 829 & - & 1 scenario & Ego & Vision Pro & Dual & \xmark & \cmark & \cmark & Device tracking & \xmark & \xmark & \xmark & \cmark & Curated \\
UniHand-1.0~\citep{luo2025being} & 2025 & 130M & - & 444.1K & 1,155 & - & - & Ego & - & Dual & \xmark & \cmark & \cmark & - & \xmark & \xmark & \xmark & \cmark & Mixed \\
IndEgo~\citep{chavan2025indego} & 2025 & 17.6M & 4,552 & - & 294 & 20 & 5 categories & Ego + Exo & \makecell[c]{Aria, Sony A6400 APSC,\\ Samsung Galaxy A51,\\ iPhone 16} & Dual & \xmark & \cmark & \xmark & Device tracking & \xmark & \cmark & \cmark & \cmark & Curated \\
LVP-1M~\citep{chen2025large} & 2025 & - & - & 1.4M & 1,167 & - & - & Ego + Exo & - & Dual & \xmark & \xmark & \xmark & - & \xmark & \xmark & \xmark & \cmark & Mixed \\
Action100M~\citep{chen2026action100m} & 2026 & - & 1.2M & 147M & 127,896 & - & - & Ego + Exo & - & Dual & \xmark & \xmark & \xmark & - & \xmark & \xmark & \xmark & \cmark & In-the-wild \\
UniHand-2.0~\citep{luo2026being} & 2026 & 400M & - & - & 35,000 & - & - & Ego & - & Dual & \xmark & \cmark & \cmark & - & \xmark & \xmark & \xmark & \cmark & Mixed \\
DreamDojo-HV~\citep{gao2026dreamdojo} & 2026 & - & 1,135K & - & 43,827 & - & 9,869 scenes & Ego & - & Dual & \xmark & \cmark & \cmark & - & \xmark & \xmark & \xmark & \cmark & Curated \\
UniHand-Mix~\citep{luo2026joint} & 2026 & - & - & 7.5M & 2,123 & - & - & Ego & - & Dual & \xmark & \cmark & \cmark & - & \xmark & \xmark & \xmark & \cmark & Mixed \\
\bottomrule
\end{tabular}
\label{tab:human_video_datasets}
\end{sidewaystable}
\clearpage

\clearpage
\begin{table*}[t]
\centering
\tiny
\setlength{\tabcolsep}{4pt}
\renewcommand{\arraystretch}{1.5}
\caption{LfHV studies grouped by the open-source human video datasets they use.}
\begin{tabular}{p{0.22\textwidth} p{0.72\textwidth}}
\toprule
Dataset & Used by \\
\midrule
Breakfast~\citep{kuehne2014language} & \citep{nguyen2018translating} \\
ActivityNet~\citep{caba2015activitynet} & \citep{rothfuss2018deep} \\
EgoHands~\citep{bambach2015lending} & \citep{lee2017learning} \\
Charades~\citep{sigurdsson2016hollywood} & \citep{qian2024contrast} \\
FPHA~\citep{garcia2017first} & \citep{ding2024knowledge,luo2025being,feng2025spatial} \\
Something-Something~\citep{goyal2017something} & \citep{rothfuss2018deep,chen2021learning,xiao2022masked,chane2023learning,majumdar2023we,bharadhwaj2023visual,radosavovic2023real,karamcheti2023language,burns2023makes,bharadhwaj2023towards,ye2024latent,bharadhwaj2024track2act,cheang2024gr,chen2024igor,ye2025video2policy,kim2025uniskill,li2025h2r,chen2025large,chen2025visa,jiang2025rynnvla,chen2025moto,chen2025villa,li2025scalable,routray2025vipra,shi2026care,dai2026conla,lyu2026lda,sun2026vla,nie2026lary,zhang2026disentangled} \\
YouCook2~\citep{zhou2017towards}    & \citep{hori2025interactive,hori2025robot}  \\
VLOG~\citep{fouhey2017lifestyle} & \citep{qian2024contrast} \\
EPIC-KITCHENS~\citep{damen2018scaling} & \citep{xiao2022masked,sivakumar2022robotic,majumdar2023we,kannan2023deft,bharadhwaj2023zero,mendonca2023structured,burns2023makes,bharadhwaj2024track2act,li2024ag2manip,cheang2024gr,chen2024igor,xiong2025ag2x2,lepert2025masquerade,song2025mitty,shi2025zeromimic,li2025scalable,chen2025villa,bjorck2025gr00t,jiang2025rynnvla,luo2026being,soraki2026objectforesight,nie2026lary} \\
EGTEA Gaze+~\citep{li2018eye} & \citep{chen2024igor,chen2025villa,li2026gazevla} \\
HowTo100M~\citep{miech2019howto100m} & \citep{mandikal2022dexvip,cheang2024gr,jiang2025rynnvla} \\
FreiHAND~\citep{zimmermann2019freihand} & \citep{sivakumar2022robotic} \\
100DOH~\citep{shan2020understanding} & \citep{xiao2022masked,patel2022learning,sivakumar2022robotic,bahl2023affordances,majumdar2023we,radosavovic2023real,dasari2023unbiased,burns2023makes,srirama2024hrp,singh2025hand,jonnavittula2025view} \\
EPIC-KITCHENS-100~\citep{damen2020rescaling} & \citep{xiao2022masked,pertsch2022cross,bahl2023affordances,ma2023liv,chang2023look,radosavovic2023real,bharadhwaj2023towards,ju2024robo,chen2025large,chen2025vidbot,heidinger20252handedafforder,jiang2025rynnvla,lyu2026lda} \\
Kinetics-700~\citep{smaira2020short} & \citep{dasari2023unbiased,cheang2024gr,zhou2025mitigating} \\
MOW~\citep{cao2020reconstructing}  & \citep{patel2022learning}  \\
DexYCB~\citep{chao2021dexycb} & \citep{sivakumar2022robotic,qin2022dexmv,ye2023learning,qian2024contrast,ci2025h2r,singh2025deep,luo2025being,feng2025spatial,singh2025hand,zhao2025dexh2r,chen2025vividex,luo2026joint} \\
H2O~\citep{kwon2021h2o} & \citep{kim2025uniskill,ma2025uni,luo2025being,feng2025spatial,luo2026joint,zhang2026unidex,li2026gazevla} \\
Ego4D~\citep{grauman2022ego4d} & \citep{nair2022r3m,ma2022vip,majumdar2023we,bhateja2023robotic,radosavovic2023real,chang2023look,wu2023unleashing,kannan2023deft,dasari2023unbiased,burns2023makes,bharadhwaj2023towards,zeng2024learning,cheang2024gr,srirama2024hrp,chen2024igor,chen2025large,zhou2025mitigating,bu2025univla,jiang2025rynnvla,bjorck2025gr00t,li2025h2r,heidinger20252handedafforder,yoshida2025developing,lin2025physbrain,chen2025villa,yang2025ar,li2025scalable,bu2025agibot,luo2026being,luo2026joint,zhang2026clap,lyu2026lda,nie2026lary,zhang2026disentangled,li2026gazevla} \\
Assembly101~\citep{sener2022assembly101} & \citep{bjorck2025gr00t} \\
EgoPAT3D~\citep{li2022egocentric} & \citep{zhang2025zero,ma2025egoloc,ma2025uni,chen2025villa} \\
AGD20K~\citep{luo2022grounded}  & \citep{ju2024robo}  \\
HOI4D~\citep{liu2022hoi4d} & \citep{kannan2023deft,kuang2024ram,yuan2024general,bjorck2025gr00t,wang2025gat,zhu2025learning,feng2025spatial,chen2025villa,zhang2025actron3d,luo2025being,yang2025egovla,luo2026joint,lyu2026lda,zhang2026unidex,li2026gazevla} \\
OakInk~\citep{yang2022oakink} & \citep{chen2025web2grasp} \\
EgoHOS~\citep{zhang2022fine} & \citep{kannan2023learning,soraki2026objectforesight}  \\
ARCTIC~\citep{fan2023arctic} & \citep{li2025maniptrans,luo2025being,feng2025spatial,luo2026joint,lyu2026lda} \\
RH20T-Human~\citep{fang2023rh20t} & \citep{bjorck2025gr00t,zhou2025mitigating,zhu2025learning,spiridonov2025generalist,chen2025villa,lyu2026lda} \\
HoloAssist~\citep{wang2023holoassist} & \citep{bjorck2025gr00t,yang2025egovla,chen2025villa,lyu2026lda,nie2026lary,li2026gazevla} \\
Ego-Exo4D~\citep{grauman2023ego} & \citep{bjorck2025gr00t,yoshida2025developing,li2025scalable,li2026act,lyu2026lda,lee2026mvp,li2026gazevla} \\
CaptainCook4D~\citep{peddi2024captaincook4d} & \citep{li2026act} \\
TACO~\citep{liu2024taco} & \citep{hsieh2025dexman,yang2025egovla,luo2025being,feng2025spatial,luo2026joint,lyu2026lda,zhang2026unidex,nie2026lary,li2026gazevla} \\
Panda-70M~\citep{chen2024panda} & \citep{chen2025large} \\
OakInk2~\citep{zhan2024oakink2} & \citep{yuan2025hermes,li2025maniptrans,zhao2025towards,luo2025being,feng2025spatial,hsieh2025dexman,luo2026joint,lyu2026lda,li2026gazevla} \\
HO-Cap~\citep{wang2024ho} & \citep{chen2025villa} \\
HOT3D~\citep{banerjee2024hot3d} & \citep{luo2025being,ma2025uni,yang2025egovla,soraki2026objectforesight,lyu2026lda,zhang2026unidex,li2026gazevla} \\
Nymeria~\citep{ma2024nymeria} & \citep{yoshida2025developing,gao2026dreamdojo,li2026gazevla} \\
EgoVid-5M~\citep{wang2024egovid} & \citep{jiang2025rynnvla} \\
HD-EPIC~\citep{perrett2025hd} & \citep{yoshida2025developing} \\
PH$^2$D~\citep{qiu2025humanoid} & \citep{cai2025n} \\
TASTE-Rob~\citep{zhao2025taste} & \citep{kim2025dexterous,luo2025being,feng2025spatial,lyu2026lda} \\
EgoDex~\citep{hoque2025egodex} & \citep{bi2025h,lin2025physbrain,goswami2025world,li2025mimicdreamer,li2025latbot,jiang2025rynnvla,luo2025being,feng2025spatial,cai2025n,bi2025motus,luo2026joint,gao2026dreamdojo,zheng2026egoscale,lyu2026lda,nie2026lary,li2026gazevla} \\
UniHand-1.0~\citep{luo2025being} & \citep{luo2026joint,luo2026being} \\
Egocentric-10k~\citep{buildaiegocentric10k2025} & \citep{lin2025physbrain,luo2026being,lyu2026lda} \\
Egocentric-100k~\citep{buildaiegocentric100k2025} & \\
IndEgo~\citep{chavan2025indego} & - \\
LVP-1M~\citep{chen2025large} & - \\
Action100M~\citep{chen2026action100m} & - \\
UniHand-2.0~\citep{luo2026being} & - \\
DreamDojo-HV~\citep{gao2026dreamdojo} & - \\
UniHand-Mix~\citep{luo2026joint} & - \\
\bottomrule
\end{tabular}
\label{tab:human_video_datasets_usedby}
\end{table*}
\clearpage

In addition, Tab.~\ref{tab:human_video_datasets_usedby} summarizes which LfHV works use each of these datasets. We find that Ego4D~\citep{grauman2022ego4d} and the EPIC-KITCHENS series (EPIC-KITCHENS~\citep{damen2018scaling} + EPIC-KITCHENS-100~\citep{damen2020rescaling}) are the most frequently used human video sources in the LfHV literature. Notably, all of them are collected in the wild, indicating a clear preference for large-scale natural human interaction data despite their weaker controllability and noisier visual conditions. This tendency is particularly evident in observation-oriented visual pretraining methods (e.g., R3M~\citep{nair2022r3m}, MVP~\citep{xiao2022masked}) and human video transformation or generation methods (e.g., H2R~\citep{ci2025h2r}, LVP~\citep{chen2025large}), which are strongly biased toward large-scale, weakly annotated, and temporally diverse datasets. These patterns suggest that, for generalizable representation learning and generative modeling, data scale and interaction diversity are often more valuable than precise geometric annotations within limited scenarios. Consistent with this observation, Something-Something~\citep{goyal2017something} also becomes very popular thanks to its large-scale data volume along with fine-grained action taxonomy, although its participants provide curated videos where they act out the templates.

In contrast, action-oriented transfer with affordances like Uni-Hand~\citep{ma2025uni} and Web2Grasp~\citep{chen2025web2grasp}, and H-RDT~\citep{bi2025h}, relies more on annotation-rich datasets such as DexYCB~\citep{chao2021dexycb}, H2O~\citep{kwon2021h2o}, OakInk2~\citep{zhan2024oakink2}, EgoDex~\citep{hoque2025egodex}, and HOT3D~\citep{banerjee2024hot3d}. Compared with large in-the-wild corpora, these datasets provide substantially richer supervision on fine-grained interaction states. This indicates that when the transfer target moves closer to executable manipulation, geometric annotations like explicit hand poses in these datasets become increasingly important. They serve as a more direct bridge from human videos to embodiment-specific robot action planning.

Moreover, recent VLA schemes (e.g., EgoVLA~\citep{yang2025egovla}, Being-H0~\citep{luo2025being}) and latent-action-based pipelines (e.g., GR00T N1~\citep{bjorck2025gr00t}, LDA-1B~\citep{lyu2026lda}) increasingly favor mixed dataset compositions rather than any single source. The emergence of resources such as UniHand-1.0~\citep{luo2025being}, UniHand-2.0~\citep{luo2026being}, and UniHand-Mix~\citep{luo2026joint} indicates that a single dataset cannot provide sufficient coverage of viewpoints, embodiments, hand annotations, and task diversity for scalable cross-embodiment transfer. By combining existing datasets with different biases and strengths, such mixtures can improve both data diversity and cross-domain robustness.

\subsection{Human Video Generation}\label{sec:human_video_gen}

\begin{figure}[t]
  \centering
  \includegraphics[width=1\linewidth]{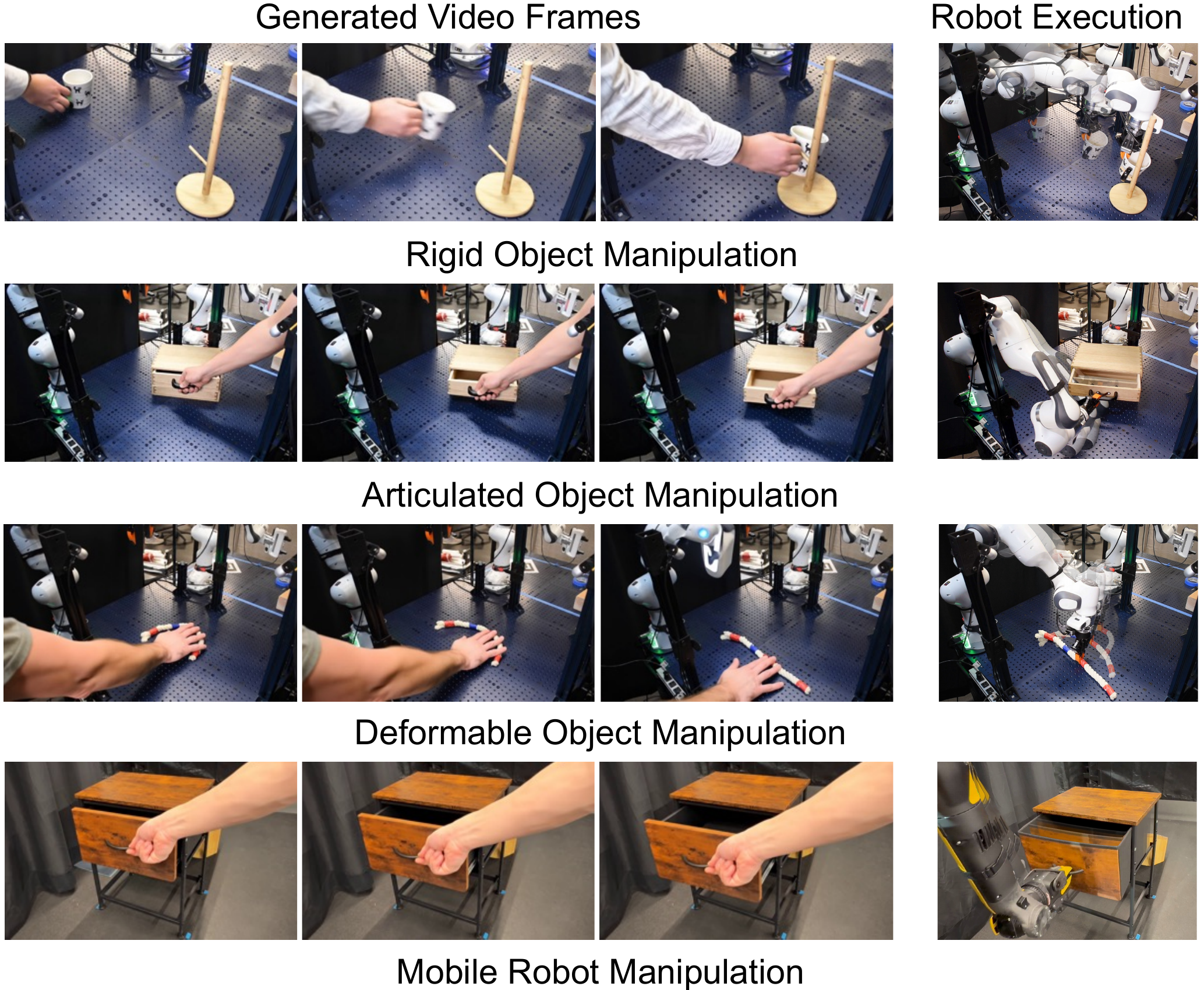}
  \caption{Illustration of human video generation for robot execution, which is originally shown in~\cite{li2025novaflow}.}
  \label{fig:nova_flow_teaser}
\end{figure}

Developing LfHV techniques only on open-source human video datasets remains restrictive. These sources are typically bounded by finite tasks, environments, viewpoints, and embodiments that have already been recorded. Manually collecting additional videos with desired properties increases human labor. This limitation has recently motivated emerging works that move beyond using pre-existing human videos toward generating human videos from scratch (see Fig.~\ref{fig:nova_flow_teaser}). By synthesizing human demonstrations under controllable task settings, viewpoints, and scene configurations, these methods offer a complementary route to expand the diversity, coverage, and adaptability of human-centric data for LfHV. Next, we detail how they generate human videos from scratch and how such videos are further leveraged to derive robot policies.

An early work by~\cite{bonardi2020learning} explores this direction by replacing real human demonstrations during training with domain-randomized simulated human-arm videos. Built on task-embedded control networks, it learns task embeddings from these synthetic human demonstrations and enables a robot to perform one-shot imitation from a single real human video at test time. With the rapid development of generative models, more works attend to directly synthesizing task-relevant human or robot interaction videos conditioned on language descriptions. For example, UniPi~\citep{du2023learning} formulates policy learning as a text-conditioned video generation problem. A video diffusion model~\citep{ho2022imagen} first generates future visual plans from task descriptions and current observations, and an inverse dynamics model then extracts executable actions from the synthesized videos. In contrast to UniPi~\citep{du2023learning}, which uses human videos to optimize the generative planning model, Gen2Act~\citep{bharadhwaj2024gen2act} further incorporates generated human videos into downstream policy optimization. Specifically, it uses an off-the-shelf language-conditioned video generator~\citep{kondratyuk2023videopoet} to synthesize human videos in novel scenes, and then trains a closed-loop robot policy conditioned on these generated videos, together with an auxiliary point-track prediction objective. 

To further alleviate the dependence on extensive robot-specific data, some works explicitly extract object poses from generated human videos and retarget them to the robot end effector. For example, Dreamitate~\citep{liang2024dreamitate} fine-tunes a video diffusion model~\citep{van2024generative} on stereo videos of humans using trackable tools. During inference, it generates human manipulation videos and recovers the 6D pose trajectories of the tools with known CAD models and stereo tracking. The trajectories are directly transferred into robot end-effector actions. Instead of tools, \cite{patel2025robotic} track 6D object poses from human videos generated by Sora~\citep{brooks2024video} and Kling~\citep{KlingAI2024}, enabling a significantly broader range of manipulation tasks. This work additionally introduces GPT-4o~\citep{achiam2023gpt} to automatically filter out unsuccessful video generations with high accuracy. To relax the rigid-body assumption of this work, NovaFlow~\citep{li2025novaflow} replaces 6D pose retargeting with actionable 3D object flow distilled from videos generated by Wan~\citep{wan2025wan} and Veo~\cite{wiedemer2025video}. This method is inherently model-free and therefore applicable to rigid, articulated, and deformable objects. Dream2Flow~\citep{dharmarajan2025dream2flow} also uses object-centric flow as an interface for downstream control. However, it further integrates human video generation~\citep{wan2025wan} with RL, where the reward functions encourage matching the extracted 3D object flow.

Instead of generating human videos for object motion information, \cite{chen2025large} attend to extracting hand movements. They train a large video foundation model as a generative planner on Internet-scale human and robot videos. This method shows especially outstanding performance on dexterous manipulation. \cite{kim2025dexterous} instead feed explicit hand motion information into human video generation by conditioning the model on egocentric hand-mesh renderings together with rendered static 3D scenes. It explicitly separates static scene structure from action-induced changes, using the rendered static scene as a spatially consistent input and training the model to synthesize only the residual dynamics caused by human manipulation. This pipeline thus encourages focusing on human-induced changes rather than regenerating the entire scene.

Human video generation is emerging as a scalable complement to static human video corpora for LfHV. Existing works have progressed from early simulation-based synthesis to language-conditioned video generation, and further to more structured pipelines that couple generated videos with pose retargeting and flow extraction. This direction is particularly promising because it expands task and scene coverage beyond what has been physically recorded. Although the current zero-shot video generation models generally produce higher-quality human videos than robot videos~\citep{bharadhwaj2024gen2act}, their effectiveness heavily depends on the fidelity of synthesized interaction dynamics, the reliability of extracting physically grounded motion signals, and the robustness of transferring those signals across embodiments.

\section{Discussion} \label{sec:discussion}

Despite the rapid progress reviewed above, robot learning from human videos is still far from a mature and standardized paradigm. This field has expanded quickly along multiple dimensions, including task parsing, visual alignment, affordance extraction, latent action modeling, and data scaling. However, several fundamental challenges remain unresolved, especially regarding how to better model HOI dynamics, exploit multimodal human signals, utilize noisy Internet data, extend from single-agent imitation to collaborative settings, and evaluate methods under unified protocols. In this section, we summarize several key challenges and discuss promising research directions that have emerged in response to them.

\subsection{Physically Grounded World Models}
Most existing LfHV methods use human videos to provide supervision, priors, or intermediate guidance for downstream robot policies. Although effective in many target cases, these formulations excessively emphasize local information like task plans, visual representations, affordances, or latent actions without explicitly modeling how the world globally evolves under embodied interaction. Therefore, they often struggle to capture long-horizon causal dependencies and maintain physical consistency across sequential interaction stages, limiting their robustness in open-world manipulation. 
Although some works~\citep{wu2023unleashing,mendonca2023structured,cheang2024gr,zhu2025learning} have shown promising results in human video prediction, they place excessive emphasis on visual similarity supervision while overlooking the structural changes induced by physical interactions. Even more recent DexWM~\citep{goswami2025world} and DWM~\citep{kim2025dexterous} still neglect to impose strict physical constraints on the modeled dynamics, despite recognizing the need to capture physical dynamics by introducing human motion information. The lack of physical constraints may lead to unrealistic robot action planning with physically invalid future rollouts, especially in contact-rich and long-horizon manipulation scenarios.

Therefore, a promising next step is to extend LfHV toward \textit{physically grounded world models}, where robots do not merely imitate observed behavior, but also learn to anticipate future physical interaction states and long-horizon consequences from large-scale human videos. Such a direction could unify currently separate paradigms like visual prediction, affordance forecasting, latent action modeling, and reward construction under a single predictive framework. More importantly, world models may provide a natural foundation for counterfactual reasoning, failure recovery, and closed-loop execution, which are all crucial for generalizable manipulation in open-world environments. However, building world models that remain physically grounded across embodiments rather than only producing visually plausible rollouts is challenging. The inherently limited nature of vision-dominant human videos further exacerbates this problem. Thus, incorporating additional VLM-based human video understanding~\citep{cheng2024egothink,plizzari2025omnia,vinod2025egovlm,su2025annexe,lin2025physbrain} may offer a practical way to inject stronger physics reasoning and long-range temporal constraints into such world models. By improving the policy backbone's ability to parse physical regularities from human videos, we may indirectly impose stronger physical constraints during world modeling without introducing additional input modalities. 

\subsection{Physics-Aware Functional Affordance Extraction}

Current affordance-based LfHV methods mainly extract visually observable cues, such as contact regions, hand trajectories, object poses, and motion flow. These representations provide effective action-level interfaces, but they often remain limited to the HOI geometric trace. For more general manipulation, future affordance extraction should move toward jointly encoding what physical function an object or object part supports, how this function is realized through motion, and which kinematic or dynamic constraints must be respected during robot execution. Recent tool manipulation works provide early evidence for this direction. FUNCTO~\citep{tang2025functo} and MimicFunc~\citep{tang2025mimicfunc} show that tool use can be better transferred by identifying function-centric structures such as function keypoints, rather than relying only on visual or geometric similarity. This suggests that affordances should describe not only where and how interactions occur, but also why a particular contact is functionally meaningful and physically feasible for achieving a manipulation effect. Similarly, articulated object manipulation further highlights the need to incorporate physical constraints into affordance representations. DITTO~\citep{jiang2022ditto} reconstructs articulated object geometry and kinematic structure from interaction, while \cite{kerr2024robot} recover 4D part-centric motion from a monocular human demonstration and plans robot motions to reproduce the demonstrated object part trajectories. More recent works~\citep{werby2025articulated,wang2026paws} further exploit in-the-wild egocentric hand-object interactions to infer articulation axes, part trajectories, and scene-level kinematic structures. These studies suggest that object affordances should not be treated as static regions or free-form trajectories alone. Instead, they should be constrained by physical properties along with manipulation semantics.

Therefore, a promising future direction is to develop \textit{physics-aware functional affordance extraction} methods that integrate visual interaction evidence, semantic functionality, and physical structure into a unified representation. Such affordances could specify not only the contact region or motion trajectory, but also the intended object function, the manipulated part, the feasible interaction mode, and the physical constraints governing execution. This would make affordance-based transfer more robust for tools, articulated objects, and contact-rich manipulation, where successful robot execution depends on respecting object functionality and physical feasibility rather than merely reproducing observed human motion form videos.

\subsection{Continual Learning with Human Video Data}

Recent robot foundation models and VLA systems increasingly rely on large-scale heterogeneous data followed by task- or embodiment-specific post-training~\citep{black2024pi_0,bjorck2025gr00t,luo2025being,yang2025egovla,lyu2026lda}. In this paradigm, human videos usually serve as a fixed offline corpus. However, most existing LfHV pipelines still lack a systematic mechanism for incorporating newly available human videos after the initial training stage. This static formulation is increasingly insufficient for open-ended robot learning, where new objects and environments continuously emerge. However, compared with real robot demonstrations, human videos provide much cheaper and more scalable updates about novel tasks, which deserve greater attention for continual learning.

A promising future direction is therefore to develop \textit{continual learning paradigms} that can continually absorb human video data as a growing source of embodied experience. Egocentric videos may continually enrich first-person manipulation priors for VLA models. Latent action models may convert newly collected videos into pseudo-action supervision for scalable policy post-training. Affordance-based pipelines benefit from updated interaction affordances without requiring dense robot demonstrations for every new task. Realizing this direction will require quality-aware filtering, memory-based sample selection, efficient adaptation, and continual evaluation, potentially drawing inspiration from lifelong robot learning benchmarks~\citep{liu2023libero}.

\subsection{Multi-Agent Interaction}

Although some researchers focus on learning bimanual manipulation policies from human videos~\citep{bahety2024screwmimic,li2025maniptrans,zhou2025you,bi2026h}, almost all existing LfHV methods are developed under a single-agent assumption. That is, one human demonstrates a manipulation behavior, and one robot is expected to reproduce it. Although this formulation covers numerous canonical manipulation settings, it does not fully reflect the collaborative nature of real-world embodied tasks (i.e., from human-human coordination to robot-robot or human-robot collaboration). In practice, many tasks involve coordinated interactions between agents, such as handovers, joint object transport, and collaborative assembly. In particular, interactions between one human and another human contain rich coordination patterns encompassing role assignment, temporal synchronization, and shared constraints on object motion. However, current LfHV pipelines remain limited in their ability to support multi-robot systems that must act jointly in shared environments.

Therefore, we anticipate increased exploration of extending LfHV from single-agent imitation toward modeling \textit{multi-agent interaction}. Possible directions include role-aware latent action modeling for coordinated behavior, and hierarchical planning frameworks that decompose collaborative tasks into agent-specific subgoals with explicit synchronization. However, enabling such directions is nontrivial. Compared to single-agent settings, multi-agent interaction introduces several additional challenges. Firstly, the framework must resolve agent correspondence across time. Besides, it must disentangle overlapping motions from multiple agents. Moreover, it needs to track contact events that occur across agents and shared objects. Finally, it must model how the action of one agent changes the feasible action space of another. The embodiment mismatch between humans and robots significantly hinders the aforementioned progress. More suitable datasets for learning multi-agent collaborative manipulation from human videos are also necessary. Advancing in this direction would substantially broaden the scope of LfHV from individual skill transfer to coordinated multi-agent behavior.

\subsection{Multimodal Signals Accompanying Human Videos}

Human sensorimotor experience is intrinsically multimodal, whereas human videos provide a vision-centric source of supervision. Thus, robot learning from only human visual observations often struggles when crucial interaction evidence is weakly observable from pixels alone. Although many LfHV works have incorporated depth information from RGB-D cameras~\citep{zhu2024vision,bharadhwaj2024track2act,ma2025uni} or metric depth estimation~\citep{chen2025vidbot,li2025novaflow,govind2026unilact}, the observed spatial structures are limited in the vision modality, which are still vulnerable to occlusions and ambiguous interaction states. As task scenarios become more diverse and complex, this limitation of overreliance on the vision modality will become more pronounced in material-dependent interactions and active perception.

Therefore, broadening LfHV from vision-centric learning with richer \textit{multimodal signals accompanying human videos} is a promising direction. \textit{Audio} is arguably the easiest to incorporate into existing LfHV frameworks, as it can be acquired alongside video at minimal additional cost. Internet human videos typically contain synchronized audio, and portable camera devices (e.g., smart phones, glasses) always provide built-in audio recording capability. The audio modality can expose interaction evidence that is difficult to observe visually, such as audio-triggered actions, occluded contact events, material-dependent responses, and failure signals. Associating audio cues with their corresponding visual sources~\citep{seth2026egoavu,zhu2026egosound} may further improve the understanding of human behavior for robot policy generation. \textit{Gaze} provides an explicit indicator of human attention and intention, which may help robots localize task-relevant objects and resolve ambiguous interaction targets. In long-horizon manipulation and active perception settings, harnessing the gaze modality may help robots anticipate phase transitions and determine where to attend next. GazeVLA~\citep{li2026gazevla} provides a pioneering example, showing that gaze can be explicitly modeled as an intermediate intention signal in the VLA framework. As summarized in Tab.~\ref{tab:human_video_datasets}, some open-source human video datasets already provide audio and gaze annotations. In particular, EGTEA Gaze+~\citep{li2018eye}, Ego4D~\citep{grauman2022ego4d}, HoloAssist~\citep{wang2023holoassist}, Ego-Exo4D~\citep{grauman2023ego}, Nymeria~\citep{ma2024nymeria}, HD-EPIC~\citep{perrett2025hd}, and IndEgo~\citep{chavan2025indego} include both modalities. These resources provide valuable data foundations for future LfHV research considering audio and gaze inputs.

With the emergence of egocentric tactile resources such as EgoTouch~\citep{zhou2026touchanything}, \textit{tactile information} could further complement vision by revealing contact force, slip, compliance, and subtle surface interactions that are difficult to recover from RGB observations alone. Given that most existing human video datasets do not provide synchronized tactile sensing, harnessing foundation tactile prediction models such as TouchAnything~\citep{zhou2026touchanything} to generate pseudo tactile annotations is a relatively practical solution.

Effectively integrating these modalities into LfHV remains challenging, as they differ substantially in temporal resolution, noise characteristics, and embodiment dependence. Nevertheless, multimodal formulations provide a promising route beyond appearance-driven transfer toward a more comprehensive understanding of human-object interaction. This may ultimately improve contact-rich manipulation policy robustness and enable more faithful skill transfer from humans to robots.

\subsection{Utilization of Low-Quality Human Videos}

Existing LfHV methods typically benefit from the scale and diversity of Internet human video data. However, a large fraction of such data is inherently low-quality, exhibiting low resolution, motion blur, severe occlusion, camera shake, weak temporal coherence, and even incomplete or failed task execution. Besides, some large egocentric video datasets such as Ego4D~\citep{grauman2022ego4d} contain many passive observation data, while videos of active manipulation are more directly beneficial for robot policy learning. To avoid these issues, the current pipelines either rely on relatively curated datasets or aggressively filter noisy web data before training~\citep{chen2024igor,cheang2024gr,cheang2025gr,zhao2025dexh2r,bjorck2025gr00t,luo2026joint}. Although this improves data cleanliness, it also reduces the scalability advantage of human videos and discards many long-tail interaction patterns that may be valuable for generalizable robot learning.

Therefore, it is important to develop LfHV frameworks that can \textit{make better use of low-quality human videos} instead of simply discarding them. Achieving this goal may require quality-aware representation learning and uncertainty-aware supervision. It may also benefit from robust pseudo-labeling and cross-video consistency modeling, such that robots can still extract reliable task semantics, interaction dynamics, and action priors from noisy observations. More importantly, low-quality human data often reflects the visual ambiguity, scene clutter, and viewpoint variation that robots must handle in open-world environments. Improving the ability of LfHV models to learn from imperfect human video data is likely to be a crucial step toward scalable and generalizable robot skill acquisition.

\subsection{Standardized Benchmarking}

Despite the growing number of LfHV works, benchmarking in this field remains highly fragmented. Existing works are often evaluated on different robots, different task setups, and different human video sources. Therefore, it remains unclear which technical choices for bridging human videos and robot execution are truly responsible for better transfer performance.

A promising direction for future research is to develop more standardized benchmarking protocols tailored to the characteristics of LfHV. A particularly important challenge is that most LfHV cannot be easily benchmarked in simulation alone such as CALVIN~\citep{mees2022calvin}, LIBERO~\citep{liu2023libero}, and RoboTwin~\citep{mu2025robotwin}. Simulation is attractive in conventional robot learning and VLA literature thanks to controllable environments and reproducible evaluation. However, many LfHV works fundamentally depend on in-domain real human videos whose appearance, camera motion, interaction dynamics, and embodiment characteristics are difficult to recreate faithfully in simulators. Therefore, new evaluation protocols will likely require constructing diverse high-fidelity simulated environments that closely correspond to the original video capture scenes. This can be potentially achieved through advanced scene reconstruction techniques such as 3D Gaussian Splatting~\citep{kerbl20233d}. Besides task success rates, future progress may also attend to the policy backbone capability in supporting temporal video understanding quality~\citep{lin2025physbrain,zhang2025zero} and human motion forecasting~\citep{ma2025novel,chen2025flowing}.

\subsection{Egocentric Data Ecosystems}

Ultimately, the LfHV field presents the increasing significance of \textit{egocentric data ecosystems}. Compared with third-person (exocentric) human videos, egocentric observations are more closely aligned with the perceptual setting of embodied agents, especially for manipulation, active perception, and intention inference. Although the popularity of EPIC-KITCHENS series~\citep{damen2018scaling,damen2020rescaling}, Ego4D~\citep{grauman2022ego4d}, EgoDex~\citep{hoque2025egodex}, HOT3D~\citep{banerjee2024hot3d}, and other emerging larger-scale egocentric resources~\citep{li2026egolive} demonstrates the effectiveness of egocentric data in LfHV, recent developments such as EgoVerse~\citep{punamiya2026egoverse} further highlight that the key challenge is not merely dataset scale, but how to support sustained contributions from individual researchers, academic labs, and industry partners under a unified framework.

A mature egocentric data ecosystem for LfHV should go beyond one-off static releases. Instead, it should support standardized collection protocols, shared task semantics, unified storage and access interfaces, and manipulation-relevant annotations that are directly useful for downstream robot learning. Such an ecosystem may integrate egocentric human videos with 3D hand and head poses, language descriptions, and richer multimodal data like aforementioned audio, gaze, and tactile signals. It should also lower the barrier to contribution through lightweight collection and crowdsourced annotation schemes, while maintaining reproducibility through standardized processing and shared evaluation protocols across labs and robot embodiments. Building this kind of ecosystem would substantially reduce duplicated efforts and provide a stronger foundation for scalable, multimodal, and reproducible LfHV research.

\subsection{Summary for Challenges and Future Directions}

The above discussion suggests several tightly connected directions for the next stage of LfHV. In terms of \textit{new modeling paradigms}, future methods may need to move beyond local transfer cues toward physically grounded world models, physics-aware functional affordances, continual learning paradigms, and multi-agent interaction modeling. They will facilitate robots to better reason about long-horizon physical dynamics and collaboration constraints. In terms of \textit{richer data modalities}, progress will likely rely on both incorporating multimodal human signals such as audio, gaze, and tactile information, and making better use of low-quality yet highly scalable Internet human videos. In terms of \textit{more standardized benchmarks}, the field still needs evaluation protocols that fairly connect human video understanding, cross-embodiment transfer, and downstream robot execution, especially because many LfHV methods cannot be adequately benchmarked in simulation alone. In terms of \textit{more collaborative ecosystems}, continuously growing egocentric data infrastructures provide the practical foundation for scalable collection, richer annotation, and reproducible comparison across labs and embodiments.

These possible future directions indicate that the long-term development of LfHV is not simply an avenue of collecting more human videos or designing stronger policy backbones. Instead, it requires a shared and integrated infrastructure in which modeling, data, evaluation, and ecosystem development evolve together. Stronger models must be supported by richer and more structured human data. Larger datasets must also be accompanied by better benchmarks and more reproducible evaluation protocols. Meanwhile, open and collaborative data ecosystems will be essential for sustaining continuous progress across labs, tasks, and robot embodiments. These joint advances can promote LfHV research toward a more scalable and generalizable foundation for embodied intelligence.

\section{Conclusion} \label{sec:conclusion}

Robot learning from human videos has emerged as a promising route toward alleviating the data bottleneck in robotic manipulation. In this survey, we reviewed this field from the dual perspectives of \emph{human-robot skill transfer} and \emph{data foundations}. We first revisited how human videos interface with robot policy learning, and then organized existing methods into a hierarchical taxonomy spanning task-, observation-, and action-oriented transfer. Under this view, human videos can support robot learning through multiple forms of information flow, including task structures, task intents, transformed videos, visual embeddings, affordances, and latent actions. Beyond taxonomy construction, we further discussed how these transfer families differ in their couplings with viewpoint choices, dependence on robot data, and learning paradigms. This highlights the characteristic tradeoffs underlying current LfHV designs. Based on the taxonomy and cross-family comparisons, we also provide practical guidelines for selecting suitable routes to develop LfHV models.

In terms of human video sources, we reviewed open-source datasets along with recent advances in human video generation. We provided a broad statistical view of dataset attributes, development trends, and prevalence in the LfHV literature. The evidence reviewed in this survey suggests that progress in LfHV is shaped not only by stronger policy architectures, but also by the scale and usability of human video data. At the same time, we argue that this field remains far from mature. Key open problems include modeling long-horizon physical interaction dynamics more faithfully, exploiting richer multimodal human signals, fully exploiting noisy Internet videos, establishing more standardized benchmarks, and supporting scalable data ecosystems. We hope this survey can serve as a reference for future LfHV works, and help support the development of more general and scalable robot learning systems driven by human videos.

\bibliographystyle{SageH}
\bibliography{main.bib}

\end{document}